\documentclass[acmtog,nonacm]{acmart}

\setcopyright{none}
\makeatletter
\let\@authorsaddresses\@empty
\makeatother

\usepackage[ruled]{algorithm2e} %

\SetAlFnt{\small}
\SetAlCapFnt{\small}
\SetAlCapNameFnt{\small}
\SetAlCapHSkip{0pt}

\usepackage{multirow}
\usepackage{pifont}
\usepackage{xspace}

\usepackage[most]{tcolorbox}
\usepackage{cleveref}
\usepackage{makecell}
\usepackage{placeins}

\usepackage{lineno}
\usepackage{natbib}
\usepackage{lipsum}
\usepackage[dvipsnames]{xcolor}

\makeatletter
\DeclareRobustCommand\onedot{\futurelet\@let@token\@onedot}
\def\@onedot{\ifx\@let@token.\else.\null\fi\xspace}

\makeatother

\newcommand{\imgbase}{figs/generation_resaults_comparison/images}
\newcommand{\gridimg}{0.09\linewidth}

\newcommand{\rowsep}{\\[2pt] \hline \\[-4pt]}

\newcommand{\methodgrid}[2]{%
    \begin{tabular}{@{}c@{\hspace{1pt}}c@{}}
        \includegraphics[width=\gridimg]{\imgbase/#1/#2_seed_001.jpeg} &
        \includegraphics[width=\gridimg]{\imgbase/#1/#2_seed_002.jpeg} \\[-1pt]
        \includegraphics[width=\gridimg]{\imgbase/#1/#2_seed_003.jpeg} &
        \includegraphics[width=\gridimg]{\imgbase/#1/#2_seed_004.jpeg}
    \end{tabular}%
}

\newcommand{\inputstack}[3]{%
    \begin{tabular}{@{}c@{}}
        \includegraphics[width=\gridimg,height=\gridimg,keepaspectratio=false]{\imgbase/#1/input_1_#2.jpeg} \\[1pt]
        \includegraphics[width=\gridimg,height=\gridimg,keepaspectratio=false]{\imgbase/#1/input_2_#3.jpeg}
    \end{tabular}%
}

\newcommand{\comprow}[3]{%
    \inputstack{#1}{#2}{#3} &
    \methodgrid{#1}{kontext} &
    \methodgrid{#1}{qwenimage} &
    \methodgrid{#1}{nanobanana} &
    \methodgrid{#1}{ours}%
}

\begin{document}
\title{Inspiration Seeds: Learning Non-Literal Visual Combinations for Generative Exploration}

\author{Kfir Goldberg}
\affiliation{\institution{BRIA AI}\country{Israel}}

\author{Elad Richardson}
\affiliation{\institution{Runway}\country{USA}}

\author{Yael Vinker}
\affiliation{\institution{MIT}\country{USA}}

\begin{teaserfigure}
    \centering
    \includegraphics[width=1\linewidth]{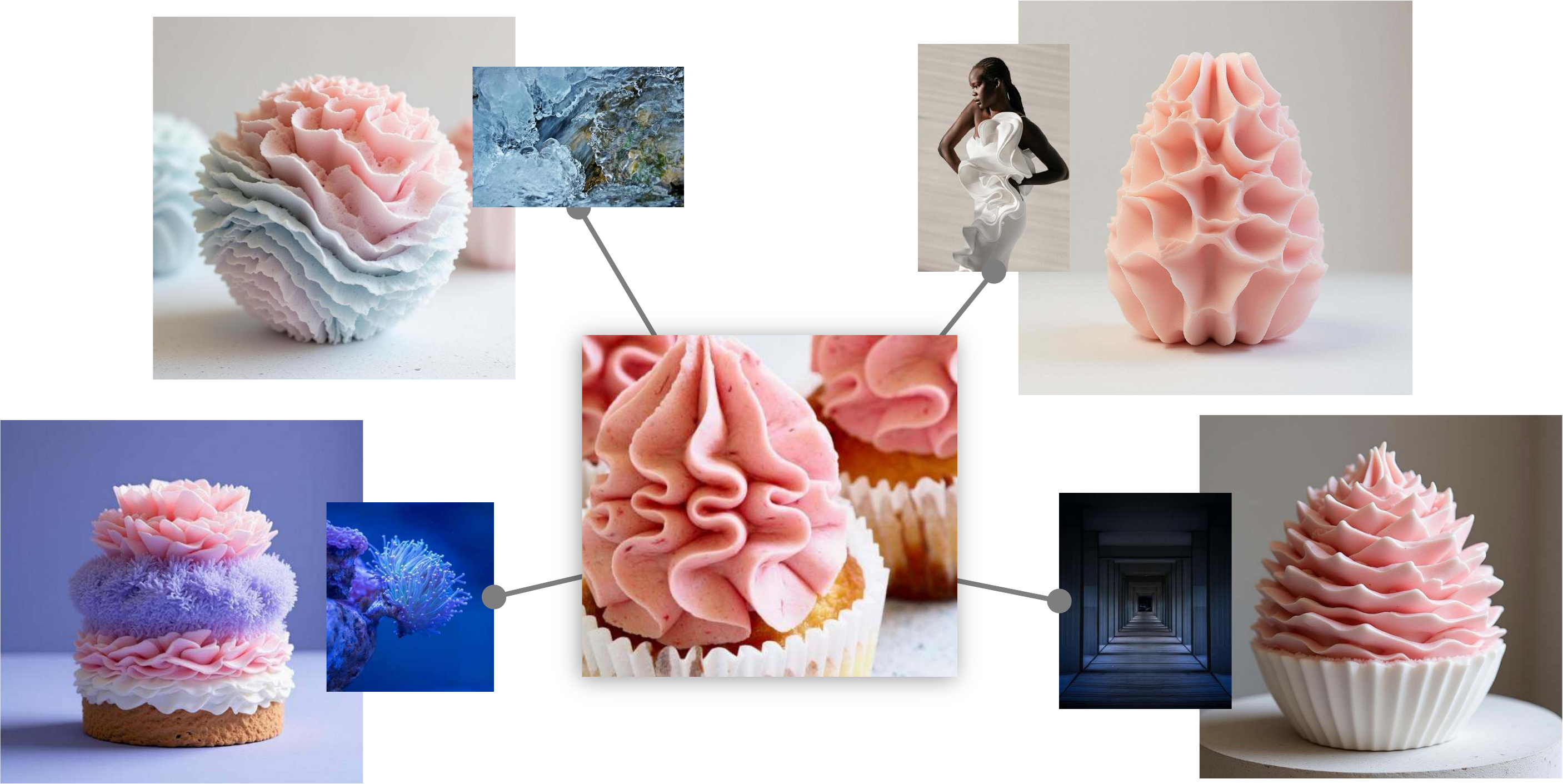} 
    \caption{Our method supports \emph{visual exploration} by generating non-trivial combinations from image pairs. These hybrids blend visual cues across domains to support early-stage ideation, without requiring users to specify their intent in text.
Here, a cupcake combined with four different visual references yields diverse transformations, from crystalline mineral layering and coral-like textures (left) to sculptural fabric forms and architectural folded surfaces (right).}
\vspace{0.1cm}
        \label{fig:teaser}
\end{teaserfigure}

\addtocontents{toc}{\protect\setcounter{tocdepth}{-10}}

\begin{abstract}
While generative models have become powerful tools for image synthesis, they are typically optimized for executing carefully crafted textual prompts, offering limited support for the open-ended visual exploration that often precedes idea formation. In contrast, designers frequently draw inspiration from loosely connected visual references, seeking emergent connections that spark new ideas. We introduce \emph{Inspiration Seeds}, a generative framework that shifts image generation from final execution to exploratory ideation. Given two input images, our model produces diverse, visually coherent compositions that reveal latent relationships between inputs, without relying on user-specified text prompts. Our approach is feed-forward, trained on synthetic triplets of decomposed visual aspects derived entirely through visual means: we use CLIP Sparse Autoencoders to extract editing directions in CLIP latent space and isolate concept pairs. By removing the reliance on language and enabling fast, intuitive recombination, our method supports visual ideation at the early and ambiguous stages of creative work.
\noindent Code and interactive demo are available at {\color{cyan!70!black}\href{https://kfirgoldberg.github.io/InspirationSeeds/}{\texttt{kfirgoldberg.github.io/InspirationSeeds/}}}.

\end{abstract}

\maketitle

\section{Introduction} 
\begin{figure}
    \centering
    \includegraphics[width=1\linewidth]{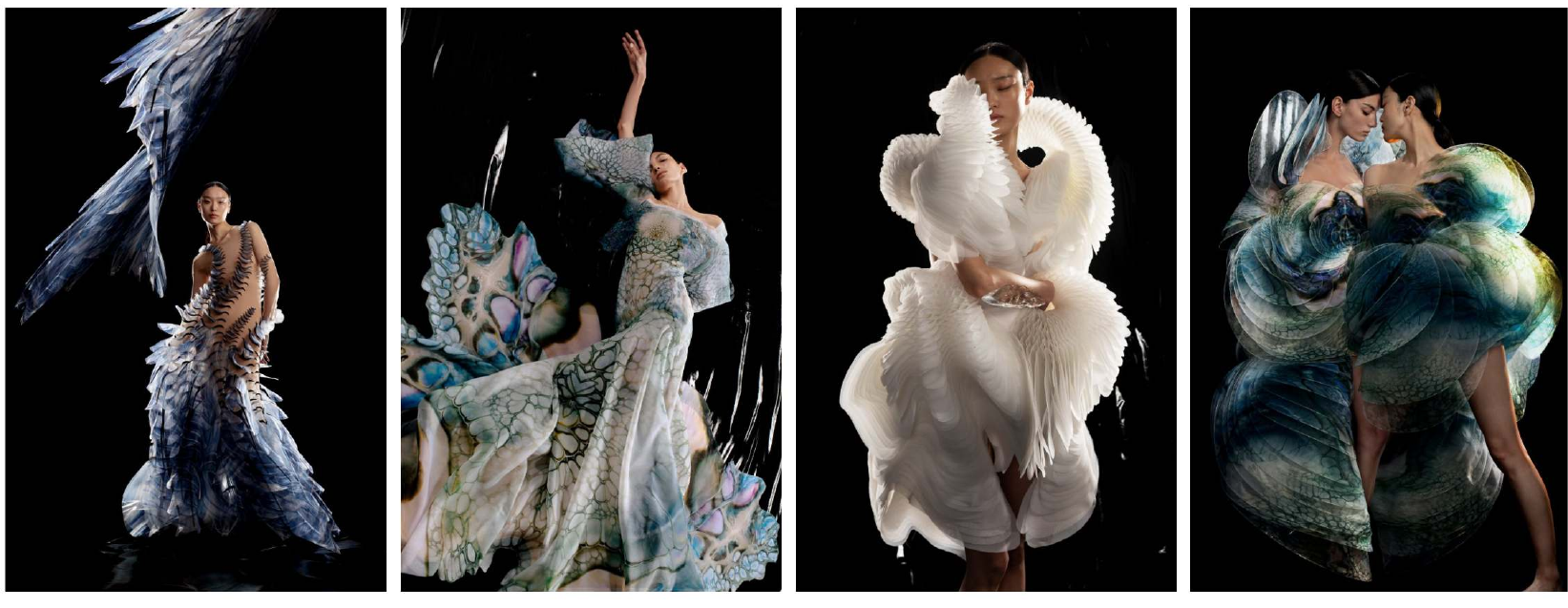}
    \caption{Dresses from Iris van Herpen's Sensory Seas collection (2020), inspired by a resemblance between deep-sea hydrozoans and neural structures.  Surfacing such unique connections is key to producing original designs.}
    \label{fig:iris_example}
\end{figure}

Ideas rarely arrive fully formed. Exploration and inspiration are key to the design process: creators explore by sketching, assembling inspiration boards, and observing artworks, natural phenomena, and abstract forms \cite{goldschmidt1991, ECKERT2000523}. Often, when examining a curated set of references, designers notice unexpected connections between familiar elements. An example of such a connection is shown in \Cref{fig:iris_example}, where fashion designer Iris van Herpen combines visual aspects of deep-sea organisms and neural structures in non-trivial ways, inspiring her collection of dresses \cite{irisvanherpen2020}.
These moments of recognition and inspiration lead to new ideas. However, perceiving such hidden visual qualities is challenging and often requires design experience and a creative eye to see beyond obvious connections \cite{goldschmidt1991, ECKERT2000523, gentner1983structure,tversky2011visualizing}.

Recent generative models offer new opportunities to support visual creation \cite{epstein2023art, mazzone2019art}, but they are typically used in a very specific way.
Most text-to-image models \cite{nanobanana2025, labs2025flux1kontextflowmatching} are designed to execute well-specified ideas through detailed prompts. As a result, they often come into play only after an idea has already been formed and verbalized. This leaves little support for the earlier, exploratory phase of creation, where ideas are still vague, intuitive, and primarily visual \cite{jonson2005design, kim2002fuzzy, arnheim1969visualthinking}.

In this paper, we propose a new perspective on the role of generative models in visual creation: using them as tools for \textit{visual exploration} rather than for producing final, polished images. From this perspective, the output of the model is not an endpoint, but an intermediate representation that can spark new ideas. To support this goal, we introduce \emph{Inspiration Seeds}, a model that takes two images as input and produces multiple visual combinations designed to surface visual relationships that are difficult to articulate verbally — revealing deep and sometimes surprising connections between the visual qualities of the inputs. 

\begin{figure}[b]
    \centering
    \includegraphics[width=1\linewidth]{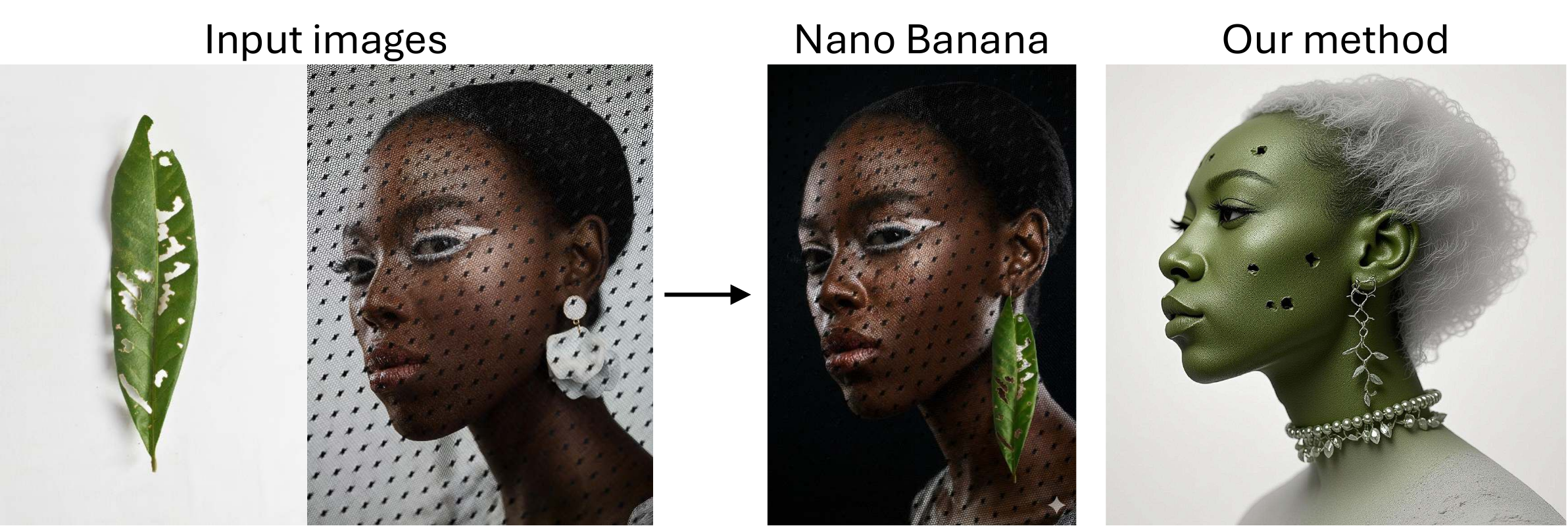}
    \caption{Trivial vs. non-trivial visual combinations. Given a leaf and a portrait, Nano Banana produces a trivial combination by replacing the earring with a leaf. Our method surfaces deeper connections: the leaf's decay pattern appears in the skin, and its aged quality carries over to the subject.}
    \label{fig:trivial_comparison}
\end{figure}

Current image generators and editing tools, even leading ones like Nano Banana \cite{nanobanana2025}, tend to produce trivial combinations even when prompted to be ``creative'', defaulting to straightforward edits as shown in \Cref{fig:trivial_comparison} (replacing an earring with a leaf). This is expected given the distribution of typical image edits these models were trained on. Generating more unexpected results typically requires careful prompt engineering and repeated intervention, which runs counter to the fluid, non-verbal nature of visual exploration \cite{suwa1997unexpected}.
Our method is designed explicitly to surface non-trivial connections without relying on text: in \Cref{fig:trivial_comparison}, the leaf's decay pattern, green tones, and aged quality carry over to the subject in unexpected ways. Such outputs can suggest new creative directions, particularly when users do not yet know what they want to create.

A key challenge in learning non-trivial visual combinations is obtaining suitable training data: triplets of two visual concepts and a corresponding non-obvious combination. Manually curating such data at scale is impractical, as it would require identifying and annotating pairs of visual concepts together with non-obvious combinations that are difficult to articulate explicitly. A natural alternative is to generate training data automatically using existing image decomposition methods and train a model to invert this process. 
However, most existing decomposition methods focus on specific, well-defined relationships, such as decomposing images into explicit object-level components~\cite{avrahami2023bas} or style–content separation~\cite{frenkel2024blora}. 
While effective for their respective goals, such formulations are inherently limited to a fixed vocabulary of relationships, making them ill-suited for learning the open-ended, non-literal combinations we target.

To go beyond these limitations, we require a data generation pipeline that avoids explicitly specifying the relationship during decomposition, allowing relevant visual aspects to emerge \emph{implicitly} from the image itself. A well-suited conceptual direction is the implicit decomposition proposed in InspirationTree \cite{inspirationtree23}, where the division into concepts is determined during optimization rather than prescribed in advance. However, this approach is designed for single-object decomposition and relies on textual inversion~\cite{gal2022textualinversion}, requiring multiple images per concept and costly per-image optimization, and often exhibiting optimization-related instabilities.

To address this, we retain the core idea of implicit decomposition while removing the reliance on costly, instable optimization. Our key insight is that the latent representations of pretrained vision–language models already encode multiple, partially disentangled visual concepts within a single image. Specifically, we propose a decompositon approach that uses CLIP Sparse Autoencoders~\cite{SAEBlog} to extract salient visual factors from each image. These factors are then grouped into coherent visual aspects and then used to define opposing directions in CLIP space that emphasize different visual aspects of the image, enabling a single image to be decomposed into complementary visual views.
This decomposition is entirely visual, requires no textual annotations, and provides a fast and scalable foundation for learning non-literal image composition.

We use our decomposition approach to construct a large-scale dataset of non-literal image decompositions and fine-tune an image-conditioned generative model \cite{labs2025flux1kontextflowmatching} to learn the inverse mapping. 
Leveraging the model's strong visual prior, our approach enables it to capture non-obvious relationships directly from data and generalize to unseen inputs. Importantly, sampling with different random seeds yields varied combinations, each surfacing distinct visual connections.
We evaluate on diverse image pairs and show that our method generates non-trivial, visually coherent combinations that reveal deeper relationships between inputs, outperforming leading models that favor literal composition. We additionally propose a description-complexity metric to evaluate this challenging task. We hope this work invites further research into generative models as tools for visual exploration and ideation.

\section{Related Work}

\paragraph{Design and Modeling Inspiration}
The ability to perceive connections between previously unrelated ideas and recombine prior knowledge in new ways is often integral to the design process and to generating new ideas \cite{Bonnardel2005TowardsSE, WILKENFELD200121, Runco2012TheSD}. In practice, this process is typically exploratory: designers and artists work with collections of visual elements and references to probe relationships and directions before a concrete concept is fully articulated \cite{ECKERT2000523}. This phase of ideation is inherently visual and associative, relying on perceived form rather than precise semantic descriptions.
Motivated by this, prior work has proposed computational tools to support ideation by facilitating the organization and comparison of visual material \cite{ImageSense2020, MetaMap2021, MoodCubes2022, Koch2019MayAD}. While effective for navigating existing examples, such systems primarily operate on curated content.

\paragraph{Image Generation and Personalization}
Recent progress in image generation has led to models capable of synthesizing high-quality images that closely follow user instructions \cite{ramesh2022hierarchical, nichol2021glidetp, rombach2022sd, saharia2022photorealistic, blackforestlabs2024flux, wu2025qwen}. These advances have established text-to-image generation as a powerful interface for visual content creation, enabling detailed control over content, style, and composition.
However, most existing generative models are optimized for execution rather than exploration. They assume that the desired outcome can be articulated through a well-specified textual prompt, offering limited support for the earlier creative phase in which ideas are still forming and primarily visual~\cite{jonson2005design, kim2002fuzzy}. In such settings, users often lack the language to precisely describe what they seek and instead rely on visual cues, references, and associations.
Personalization techniques~\cite{gal2022textualinversion,Ruiz2022DreamBoothFT} extend text-to-image models by allowing them to incorporate specific user-provided concepts into the generation process. While highly effective for reproducing known concepts, these approaches are not designed to encourage novel visual recombination and are less suited for exploring alternative interpretations or combining multiple visual elements in open-ended ways.

\paragraph{Concept Decomposition}
Decomposing images into meaningful visual components is inherently ill-posed, as high-level visual aspects are often entangled and do not correspond to explicit spatial regions or predefined categories.
Prior work has explored decomposition along fixed axes — extracting objects via mask-guided personalization \cite{avrahami2023bas, kumari2023customdiffusion, TokenVerse25}, modeling predefined attributes \cite{Xu2024CusConceptCV, lee2024languageinformed}, or separating style from content \cite{frenkel2024blora, ZipLoRA24, ngweta2023simple, Gatys2015ANA, Cross-ImageAlaluf24}. While effective for their intended purposes, these methods rely on predetermined decomposition dimensions rather than discovering new visual factors.
InspirationTree \cite{inspirationtree23} takes a different approach by decomposing a visual concept into unexpected, hierarchical visual attributes. However, it relies on textual inversion \cite{gal2022textualinversion}, requiring multiple images of the target concept across different views and backgrounds, as well as hours of optimization per concept. This process is often unstable, making it ill-suited for our setting, where we aim to decompose single images, which may not depict concrete, isolated objects, and to operate efficiently at scale.

Recent advances in mechanistic interpretability offer a promising alternative. Sparse Autoencoders (SAEs), originally proposed to identify monosemantic features in language models \cite{cunningham2023sparseautoencodershighlyinterpretable}, have been applied to CLIP \cite{Radford2021LearningTV}, decomposing its representations into sparse, interpretable visual factors \cite{fry2024towards, SAEBlog, zaigrajew2025msae}. Building on this, our approach leverages SAE-derived features to decompose arbitrary images into interpretable concepts in a single forward pass, without per-image optimization or concept-specific training.

\paragraph{Visually Inspired Generation}
Early work on interactive evolutionary computation showed that rich visual artifacts can emerge through iterative selection, without requiring users to explicitly specify their goals \cite{Sims1991, Takagi2001}. Picbreeder \cite{secretan2008picbreeder} extended this paradigm to collaborative online settings, enabling open-ended exploration of large design spaces. Similarly, DeepDream \cite{Mordvintsev2015} revealed that the internal representations of neural networks can serve as a substrate for visual discovery.
More recently, generative models have been explored as tools for visual inspiration, helping users discover new ideas through alternative interpretations or novel combinations of visual concepts \cite{hertzmann2018can, elhoseiny2019creativity, Oppenlaender_2022, white2020ganbreeder}. 
Several approaches use vision–language guidance to learn novel concepts within broader visual categories \cite{richardson2024conceptlab, lee2024languageinformed}, while others focus on visually conditioned generation, where models are guided by image embeddings rather than text.
Methods such as IP-Adapter \cite{ye2023ip} enable manipulation in embedding space and have been used to define composition rules over visual concepts \cite{richardson2025pops, dorfman2025ip, richardson2025piece}.
However, these approaches typically rely on predefined operations, leaving open the challenge of enabling open-ended, non-literal visual exploration driven purely by visual input.

\begin{figure*}[h]
    \centering
    \includegraphics[width=1\linewidth]{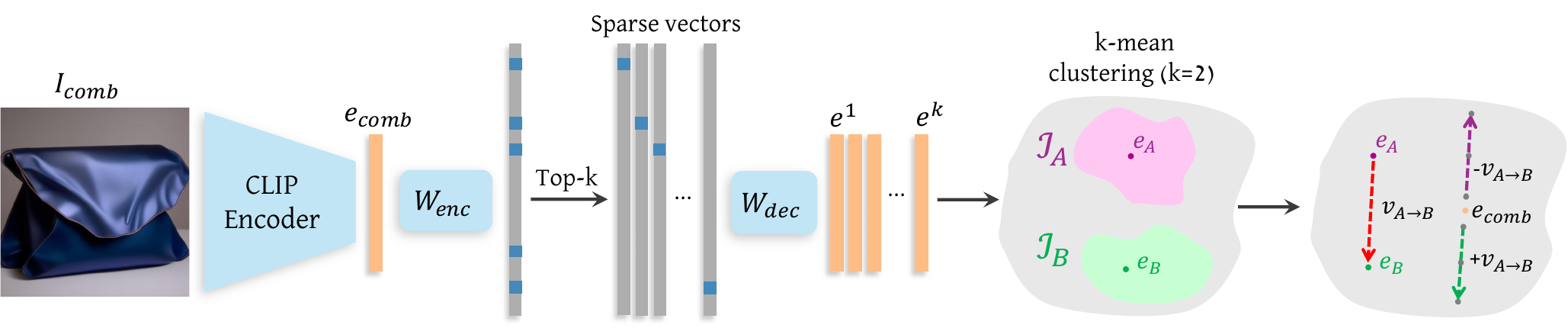}
    \vspace{-0.5cm}
    \caption{Overview of our image decomposition pipeline. Given an image $I_{comb}$, we encode it via CLIP and pass the embedding through an SAE encoder $W_{enc}$. We retain the top-k activations as sparse vectors, and decode them back to CLIP space via $W_{dec}$. We then cluster the resulting vectors into two groups using k-means. The editing direction $v_{A\to B}$ is computed as the difference between cluster centroids. Moving $e_{comb}$ in opposite directions along this axis and decoding via Kandinsky yields two images $I_A$ and $I_B$ that emphasize distinct visual aspects of the original image.}
    \label{fig:pipeline}
\end{figure*}

\section{Preliminaries}
\label{sec:preliminaries}

\paragraph{Sparse Autoencoders for CLIP}
Neural networks often exhibit polysemanticity, where individual neurons respond to multiple, semantically distinct concepts. This phenomenon arises from ``superposition'', in which more features are encoded than there are available representational dimensions~\cite{cunningham2023sparseautoencodershighlyinterpretable}. As a result, individual activation dimensions are difficult to interpret in isolation. Sparse Autoencoders (SAEs) address this by learning an overcomplete and sparse factorization of activations into interpretable features.
Given an activation vector $\mathbf{a} \in \mathbb{R}^n$ from a network layer, an SAE learns an encoder $\mathbf{W}_{\text{enc}} \in \mathbb{R}^{m \times n}$ and a decoder $\mathbf{W}_{\text{dec}} \in \mathbb{R}^{n \times m}$ where $m \gg n$:
\begin{equation}
    \mathbf{h} = \sigma(\mathbf{W}_{\text{enc}}\mathbf{a} + \mathbf{b}_{\text{enc}}), \quad \hat{\mathbf{a}} = \mathbf{W}_{\text{dec}} \mathbf{h} + \mathbf{b}_{\text{dec}},
\end{equation}
The SAE is trained with a sparse reconstruction loss $\mathcal{L}_{\text{SAE}} = \|\mathbf{a} - \hat{\mathbf{a}}\|_2^2 + \lambda \|\mathbf{h}\|_1$, 
which encourages $\mathbf{h}$ to activate only a small subset of features for any given input. Each column of $\mathbf{W}_{\text{dec}}$ corresponds to a learned feature direction, while the sparse coefficients $\mathbf{h}$ indicate which features are present in the activation.

SAEs have been recently applied to CLIP's latent space \cite{SAEBlog}, decomposing its representations into sparse, interpretable visual factors \cite{fry2024towards, zaigrajew2025msae}. In this process, CLIP’s 1280-dimensional embeddings are mapped with a dedicated $\mathbf{W}_{\text{enc}}$ into a sparse 163k-dimensional space, where individual features correspond to visual concepts, enabling separation of meaningful image aspects. We utilize the SAE-derived features to decompose arbitrary images into interpretable concepts in a single forward pass, without per-image optimization or concept-specific training.

\paragraph{FLUX.1 Kontext}
FLUX.1 Kontext~\shortcite{labs2025flux1kontextflowmatching} is a rectified flow model that unifies image generation and editing.
Kontext processes concatenated image and text token sequences through a Multimodal Diffusion Transformer~\cite{peebles2023scalable,esser2024scaling}, supporting both generation and in-context editing within a single architecture. Its strong prior on both generation and context understanding makes it a natural candidate for our base model.

\section{Method}
Our goal is to design a model that takes two images as input and generates multiple visual combinations that reveal non-trivial connections between them, without relying on textual supervision or user-provided instructions.
We formulate this task as training an image-to-image model $f_\theta(I_A, I_B) \rightarrow I_{\text{comb}}$, which receives two images and outputs a combined image.
We fine-tune Flux.1 Kontext~\cite{labs2025flux1kontextflowmatching}, a large pretrained model for image generation and editing, to perform this visual composition task.
A central challenge is obtaining suitable training data: triplets $(I_A, I_B, I_{\text{comb}})$, where the combination reflects a meaningful visual relationship rather than a superficial one.
Manually constructing such data at scale is impractical.
Our key insight is to invert this problem: instead of searching for image pairs that combine well, we start from visually rich images and decompose them into two constituent visual aspects.
The original image then serves as a ground-truth combination, providing natural supervision for training.

\subsection{Image Pool Construction}
\label{subsec:pool}
Our decomposition approach requires images that intentionally combine multiple distinct visual aspects within a single image. Typical single-object photographs may vary in color, pose, or shape, but they rarely contain several independently meaningful visual ideas that can later be separated and recombined. To obtain such images at scale, we utilize off-the-shelf text-to-image models \cite{reve2024,
seedream2025seedream40nextgenerationmultimodal, blackforestlabs2024flux,gutflaish2025generatingimage1000words} with two prompting strategies that serve complementary purposes. 

\begin{figure}[t]
\centering
\setlength{\tabcolsep}{0.5pt}
\small

\begin{tabular}{ccccc}

$I_A$ & & $I_{comb}$ & & $I_B$ \\
[2pt]

\includegraphics[width=0.2\linewidth]{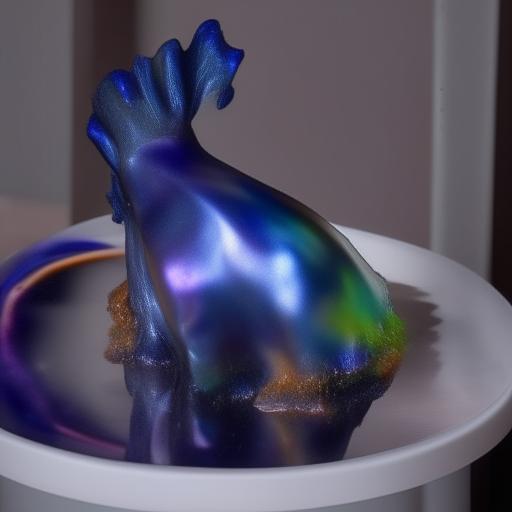} &
\includegraphics[width=0.2\linewidth]{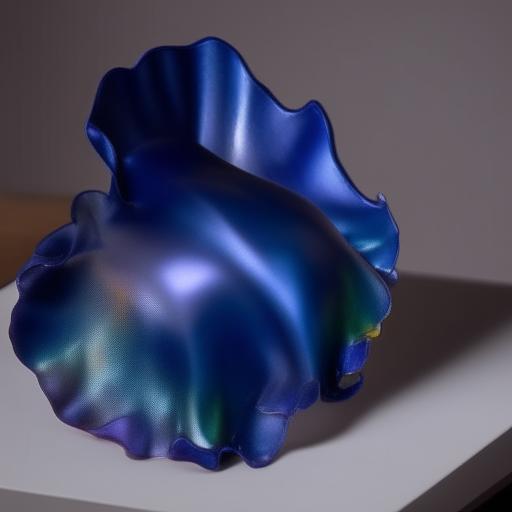} &
\includegraphics[width=0.2\linewidth]{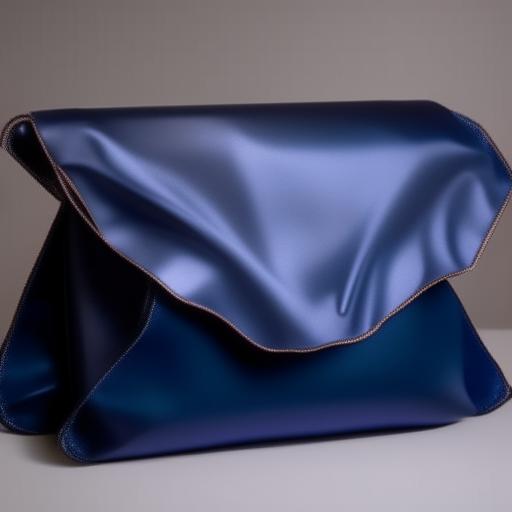} &
\includegraphics[width=0.2\linewidth]{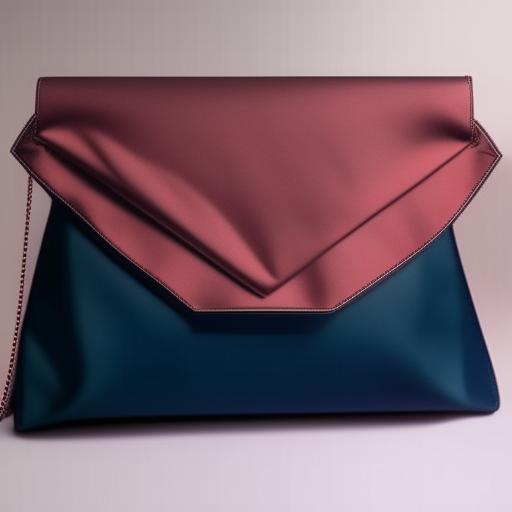} &
\includegraphics[width=0.2\linewidth]{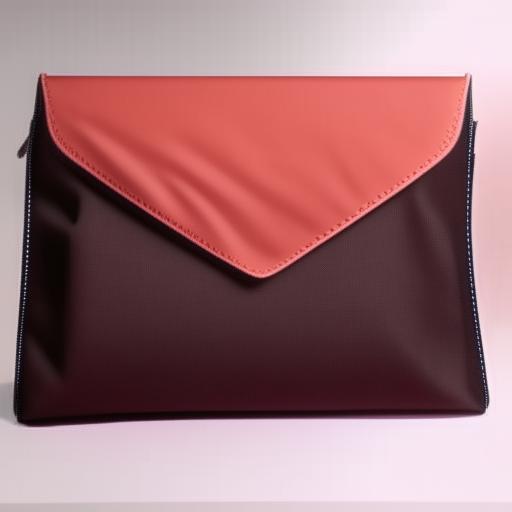} \\

\includegraphics[width=0.2\linewidth]{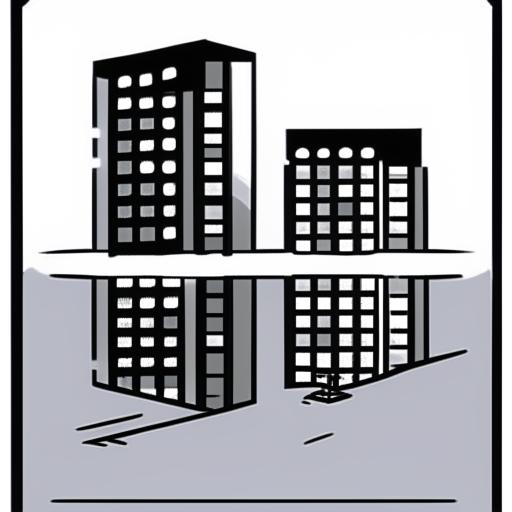} &
\includegraphics[width=0.2\linewidth]{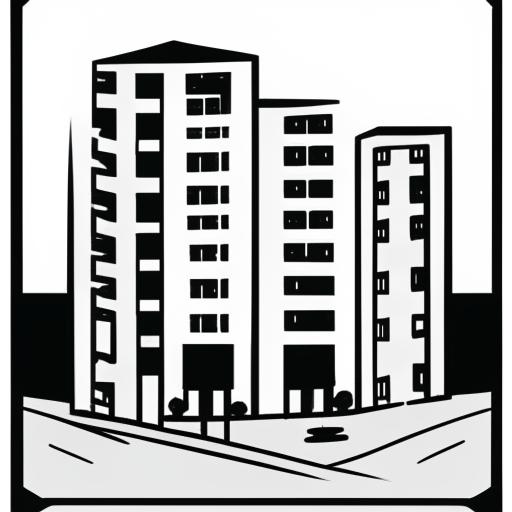} &
\includegraphics[width=0.2\linewidth]{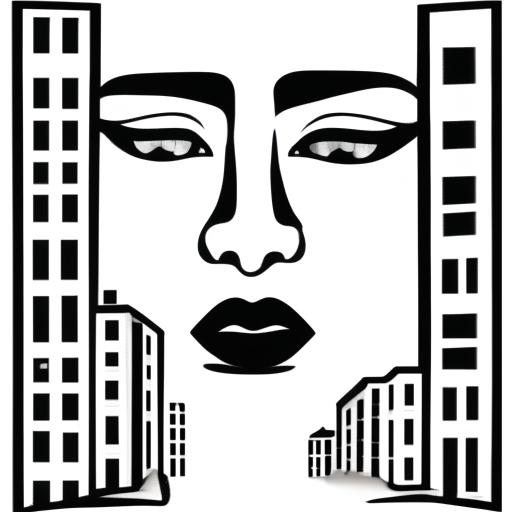} &
\includegraphics[width=0.2\linewidth]{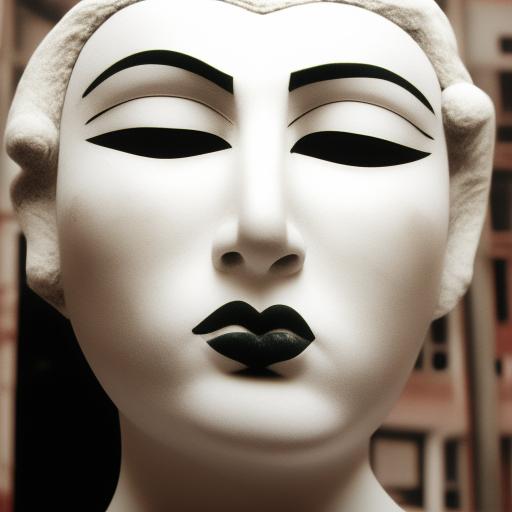} &
\includegraphics[width=0.2\linewidth]{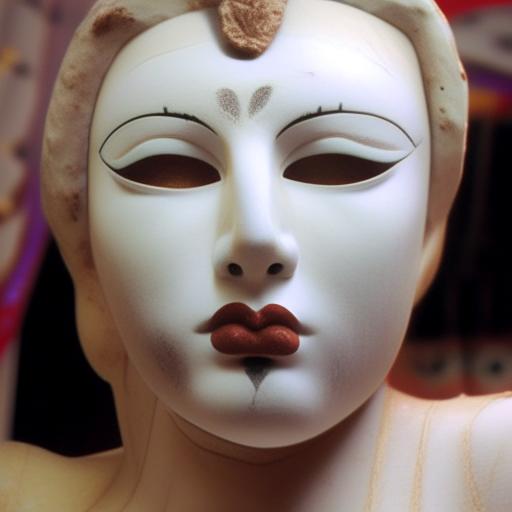} \\

\includegraphics[width=0.2\linewidth]{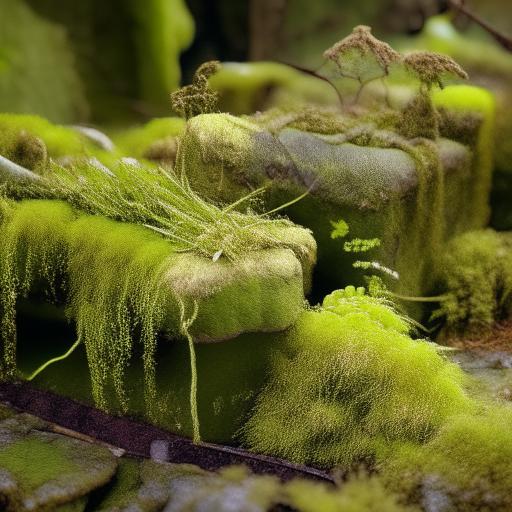} &
\includegraphics[width=0.2\linewidth]{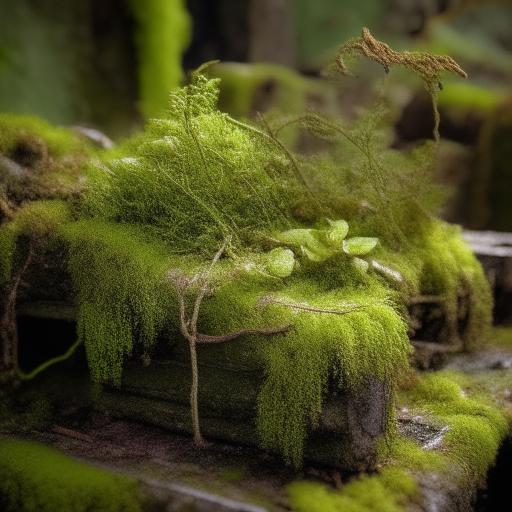} &
\includegraphics[width=0.2\linewidth]{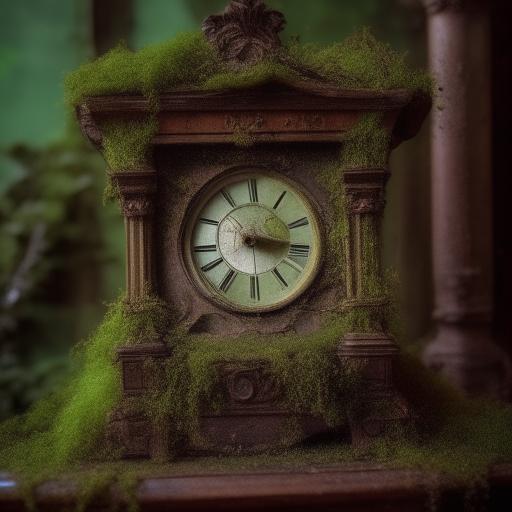} &
\includegraphics[width=0.2\linewidth]{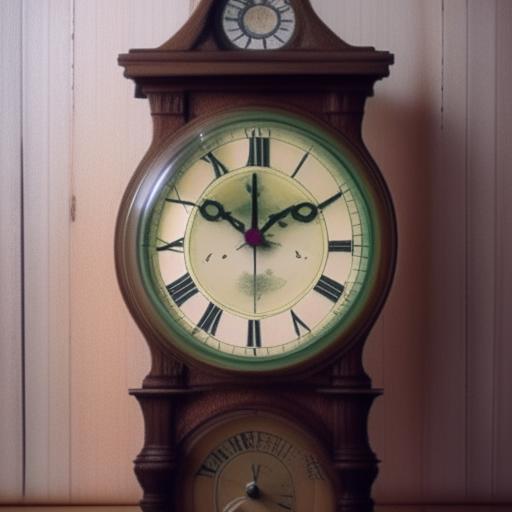} &
\includegraphics[width=0.2\linewidth]{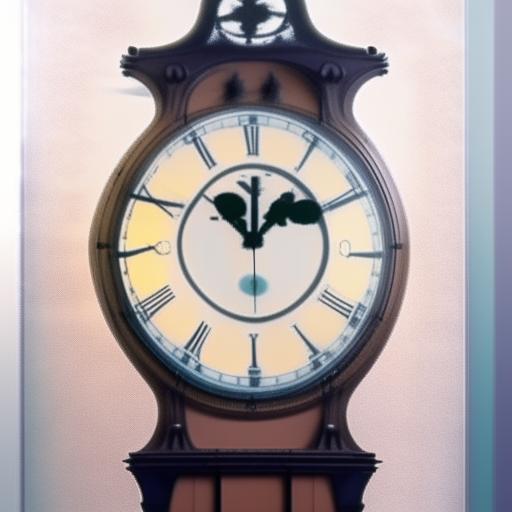} \\

\end{tabular}
\caption{Decomposition examples for varying $\lambda$ in \cref{eq:interp}. Each row shows the input image $I_{comb}$ (center) decomposed into two visual aspects $I_A$ (left) and $I_B$ (right). Intermediate columns show $\lambda = 0.5$. At $\lambda = 1$, the two aspects are well-separated while maintaining image quality.}
\label{fig:decomposition_exmaples}
\vspace{-0.2cm}
\end{figure}

First, we construct templated prompts that explicitly specify multiple visual properties such as material, color, shape, and context (e.g., \textit{``\{adj\} made of \{material\}, \{medium\}...''}). This process produces ``multi-attribute'' images, making them reliably decomposable (an example is shown in \Cref{fig:decomposition_exmaples}, first row, middle). 

Second, to encourage more creative visual concepts that are less trivially decomposed, we generate intentionally vague prompts (e.g., ``a place that never was'') and use Gemini \shortcite{gemini25_2025} to expand them into multiple distinct visual interpretations, yielding semantically related but visually diverse images. This results in a pool of visually rich images designed to support decomposition into non-trivial visual aspects.
See supplementary material for more details.

Having constructed images that bundle multiple visual aspects, we next decompose each image into its constituent aspects.

\subsection{Image Decomposition via CLIP SAEs}
\label{subsec:sae}
Decomposing a single image into multiple meaningful visual aspects is a highly non-trivial task, as there is no canonical way to separate an image into constituent components and high-level visual aspects are often entangled. While this problem has been explored in prior work \cite{inspirationtree23, avrahami2023bas, kumari2023customdiffusion, frenkel2024blora}, most existing approaches rely on text-to-image personalization optimization \cite{gal2022textualinversion,Ruiz2022DreamBoothFT}, which is time-consuming and typically decomposes images into explicit sub-objects rather than more abstract or non-obvious visual aspects.

To address this, we formulate decomposition as controlled editing in CLIP latent space, where linear directions correspond to meaningful visual transformations \cite{Radford2021LearningTV}. Rather than relying on predefined attributes or textual supervision, our approach derives image-specific decomposition axes directly from the visual content of each image.

Our pipeline is illustrated in \Cref{fig:pipeline}. Given a source image $I_{comb}$ from the set described above, we encode it with CLIP to obtain a 1280-dimensional embedding $e_{comb} = \text{CLIP}(I_{comb})$. Our goal is to produce two edited embeddings, $e_A$ and $e_B$, that separate $e_{comb}$ into two dominant visual aspects. We formulate this as finding an editing direction $v_{A \to B}$ in CLIP latent space such that moving $e_{comb}$ in opposite directions along this vector (\Cref{fig:pipeline}, right) yields the desired separation.

To identify the editing directions, we leverage CLIP Sparse Autoencoders (SAEs) \cite{SAEBlog}, which expose interpretable visual attributes from CLIP embeddings.
Given $e_{comb}$, we first encode it using the SAE encoder $W_{enc}$ to obtain a sparse 163k-dimensional vector.
Highly activated entries often correspond to interpretable visual features, but they are not fully disentangled—multiple features can capture closely related attributes with subtle variations.
Therefore, rather than selecting the top-2 highly activated individual features, we aim to find two groups of features that represent two different meaningful visual aspects.
We construct a set of sparse vectors for the top-$k$ $k=32$) activated features, by preserving the feature's activation magnitude and zeroing out all others.

We then decode each sparse vector back into CLIP space using $W_{dec}$, producing a set of vectors $\{e^1, \ldots, e^k\}$ (shown in orange). We cluster these vectors using $k$-means with $k{=}2$, yielding index sets $\mathcal{I_A}$ and $\mathcal{I_B}$. To improve cluster coherence, we retain the $50\%$ of vectors in each cluster closest to the centroid.

From the filtered clusters we compute an editing direction as the difference between the two cluster centroids:
\begin{equation}
e_A = \frac{1}{|\mathcal{I_A}|} \sum_{j \in \mathcal{I_A} e^j}e^j, \quad e_B = \frac{1}{|\mathcal{I_B}|} \sum_{j \in \mathcal{I_B} e^j}e^j, \quad v_{A \to B} = e_B - e_A.
\end{equation}
We then produce two edited embeddings by moving the original embedding in opposite directions:
\begin{equation}
e_{comb \to A} = e_{comb} - \lambda \, v_{A \to B}, \quad e_{comb \to B} = e_{comb} + \lambda \, v_{A \to B}.
\label{eq:interp}
\end{equation}
Finally, we generate images $I_A$ and $I_B$ from these embeddings using the Kandinsky model \cite{Razzhigaev2023KandinskyAI}, which was designed to support CLIP embedding conditioning. 
\Cref{fig:decomposition_exmaples} illustrates how interpolating in CLIP space along the editing direction $v_{A \to B}$ affects the generated images for different values of $\lambda$. We use $\lambda = 1$ for the final dataset construction.

\subsection{Training}
\label{subsec:train}
Our final synthetic image pool consists of 2085 images from which we produce 2085 triplets $(I_A,I_B,I_{comb})$ using our decomposition pipeline. Using this set we can now train a model to perform the inverse task: given two images, produce a combination that captures visual aspects of both.
In practice, we fine-tune Flux.1 Kontext \cite{labs2025flux1kontextflowmatching} using LoRA with a rank of 32. 
The input images are resized to $512 \times 512$ pixels and placed on a $1024 \times 1024$ canvas: $I_A$ in the top-left corner and $I_B$ in the bottom-right, with the remaining area filled with white. The model is trained to generate $I_{comb}$ conditioned on this canvas.
To avoid textual bias during training and inference, we use a fixed prompt: \textit{``Combine the element in the top left with the element in the bottom right to create a single object inspired by both of them.''} We tune the model for $15k$ steps using the Ostris AI-Toolkit~\shortcite{ostrisAIToolkit}.
At inference, given any two images, the model can generate multiple combinations by varying the random seed, surfacing different visual relationships between the inputs.

\begin{figure}[t]
    \centering
    \setlength{\tabcolsep}{0.5pt}
    \newcommand{\imgw}{0.16\linewidth}
    \renewcommand{\arraystretch}{0}

    \begin{tabular}{@{}cc@{\hspace{8pt}}cccc@{}}
    \toprule
    \multicolumn{2}{c}{Inputs} & \multicolumn{4}{c}{Results under different seeds} \\
    \midrule

    \includegraphics[width=\imgw,height=\imgw,keepaspectratio=false]{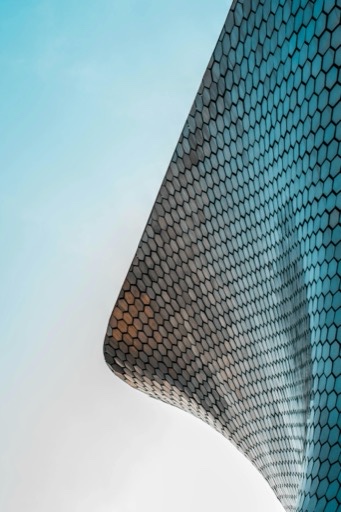} &
    \includegraphics[width=\imgw,height=\imgw,keepaspectratio=false]{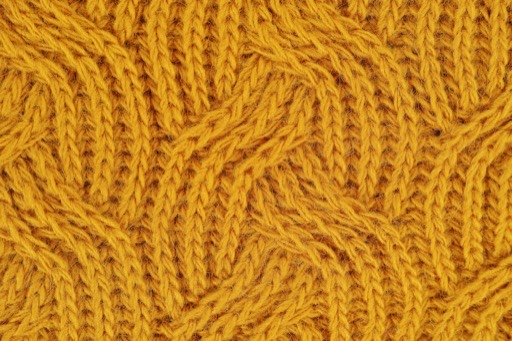} &
    \includegraphics[width=\imgw]{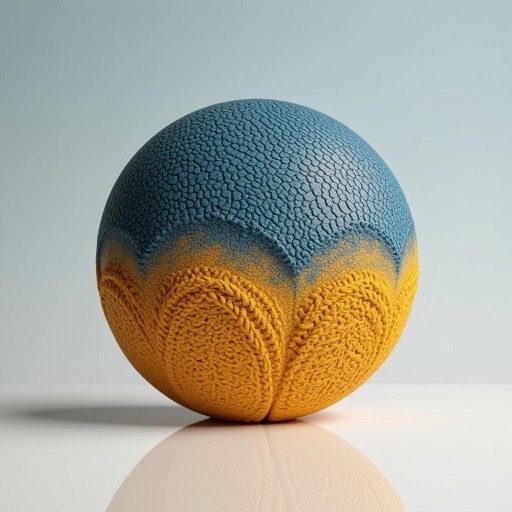} &
    \includegraphics[width=\imgw]{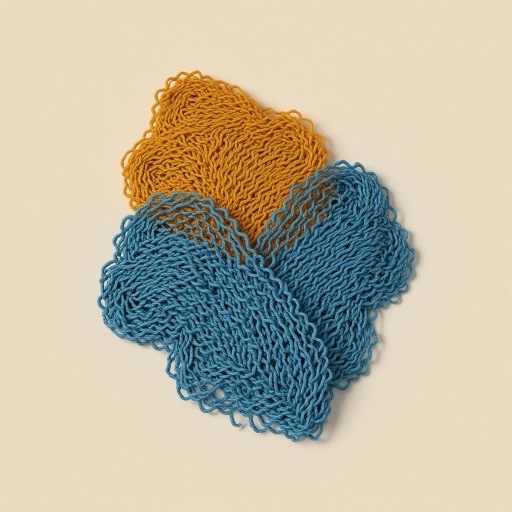} &
    \includegraphics[width=\imgw]{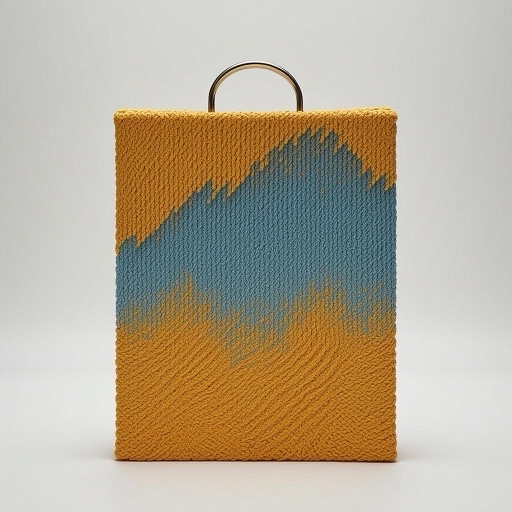} &
        \includegraphics[width=\imgw]{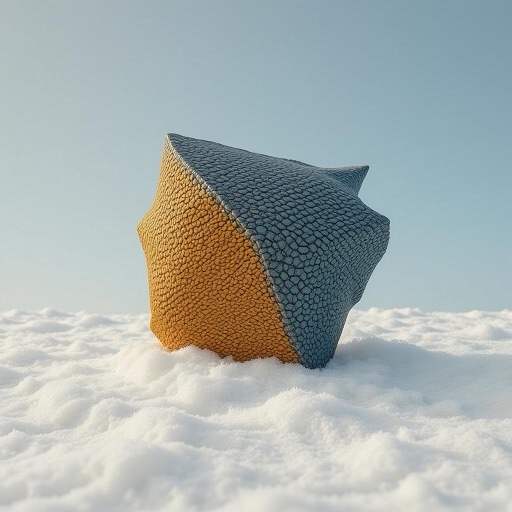} \\
    
    \includegraphics[width=\imgw,height=\imgw,keepaspectratio=false]{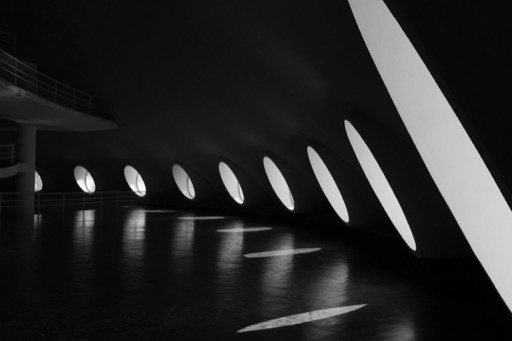} &
    \includegraphics[width=\imgw,height=\imgw,keepaspectratio=false]{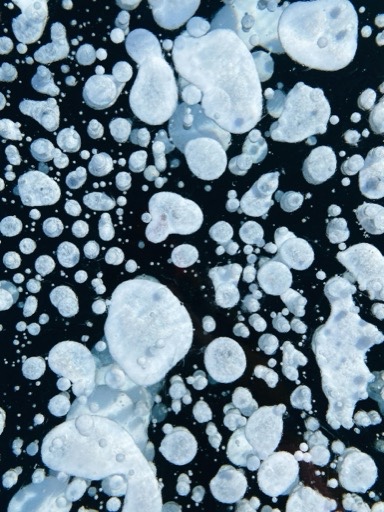} &
    \includegraphics[width=\imgw]{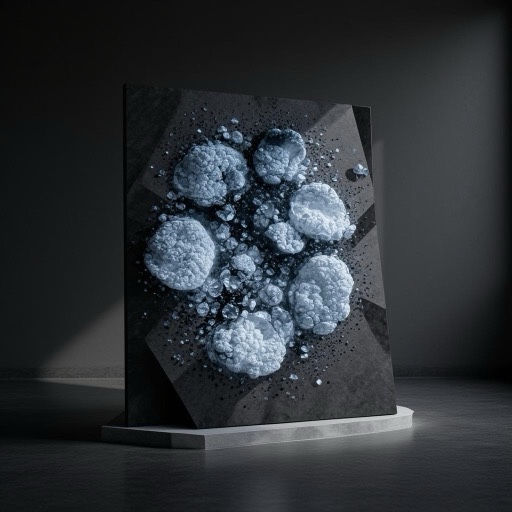} &
    \includegraphics[width=\imgw]{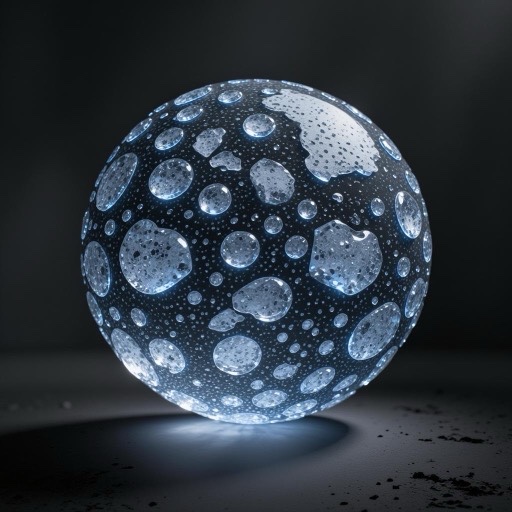} &
    \includegraphics[width=\imgw]{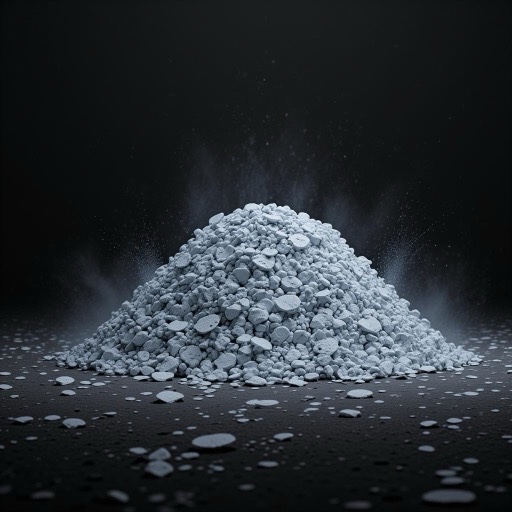} &
        \includegraphics[width=\imgw]{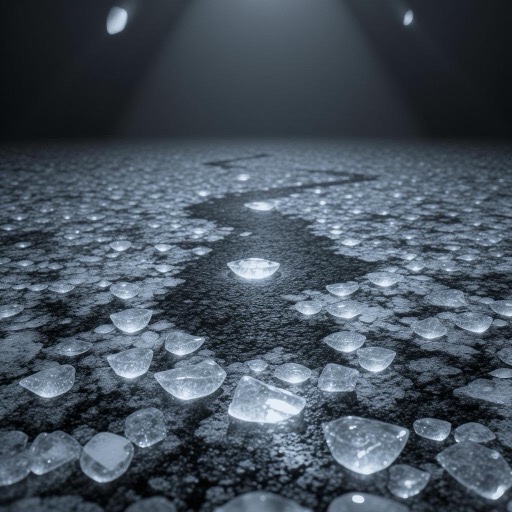} \\

    \includegraphics[width=\imgw,height=\imgw,keepaspectratio=false]{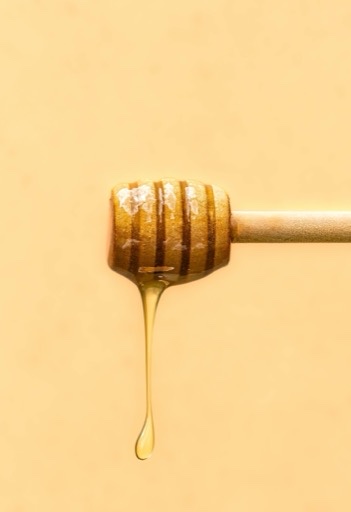} &
    \includegraphics[width=\imgw,height=\imgw,keepaspectratio=false]{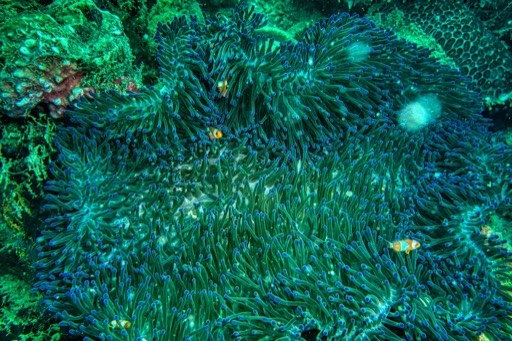} &
    \includegraphics[width=\imgw]{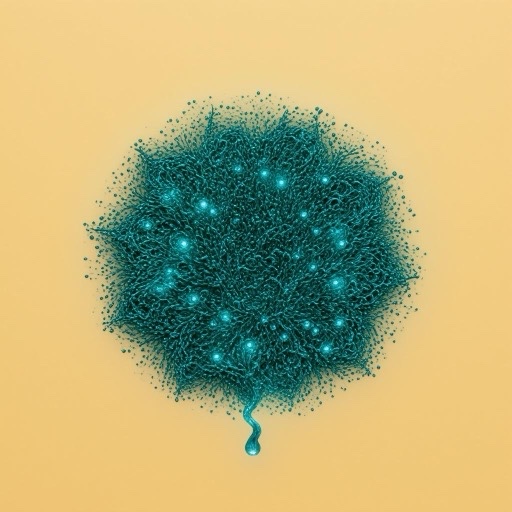} &
    \includegraphics[width=\imgw]{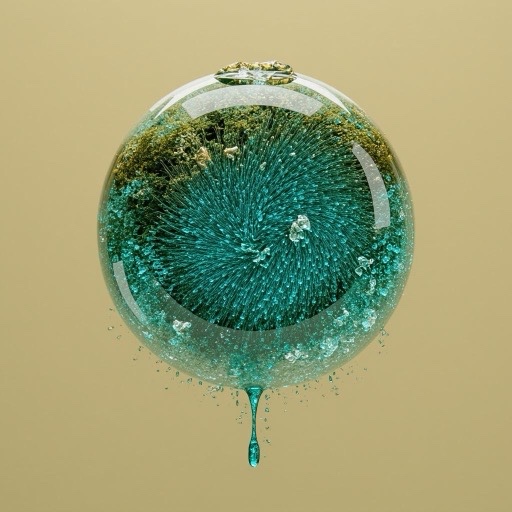} &
    \includegraphics[width=\imgw]{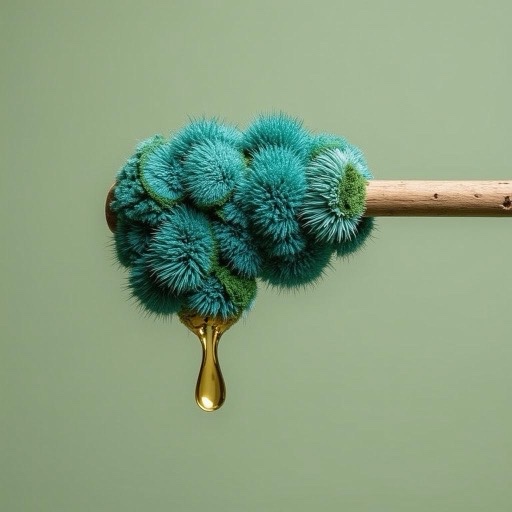} &
        \includegraphics[width=\imgw]{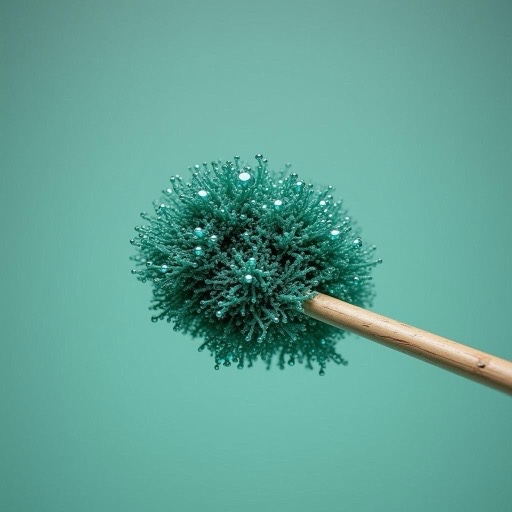} \\        
    \end{tabular}
    \vspace{-0.2cm}
    \caption{Visual Combinations under different seeds. For the same pair of input images our model can produce different visual combinations just by varying the seed, without any explicit guidance.
    }
    \vspace{-0.2cm}
    \label{fig:results_multiple_seeds}
\end{figure}

\begin{figure*}
    \centering
    \includegraphics[width=1\linewidth]{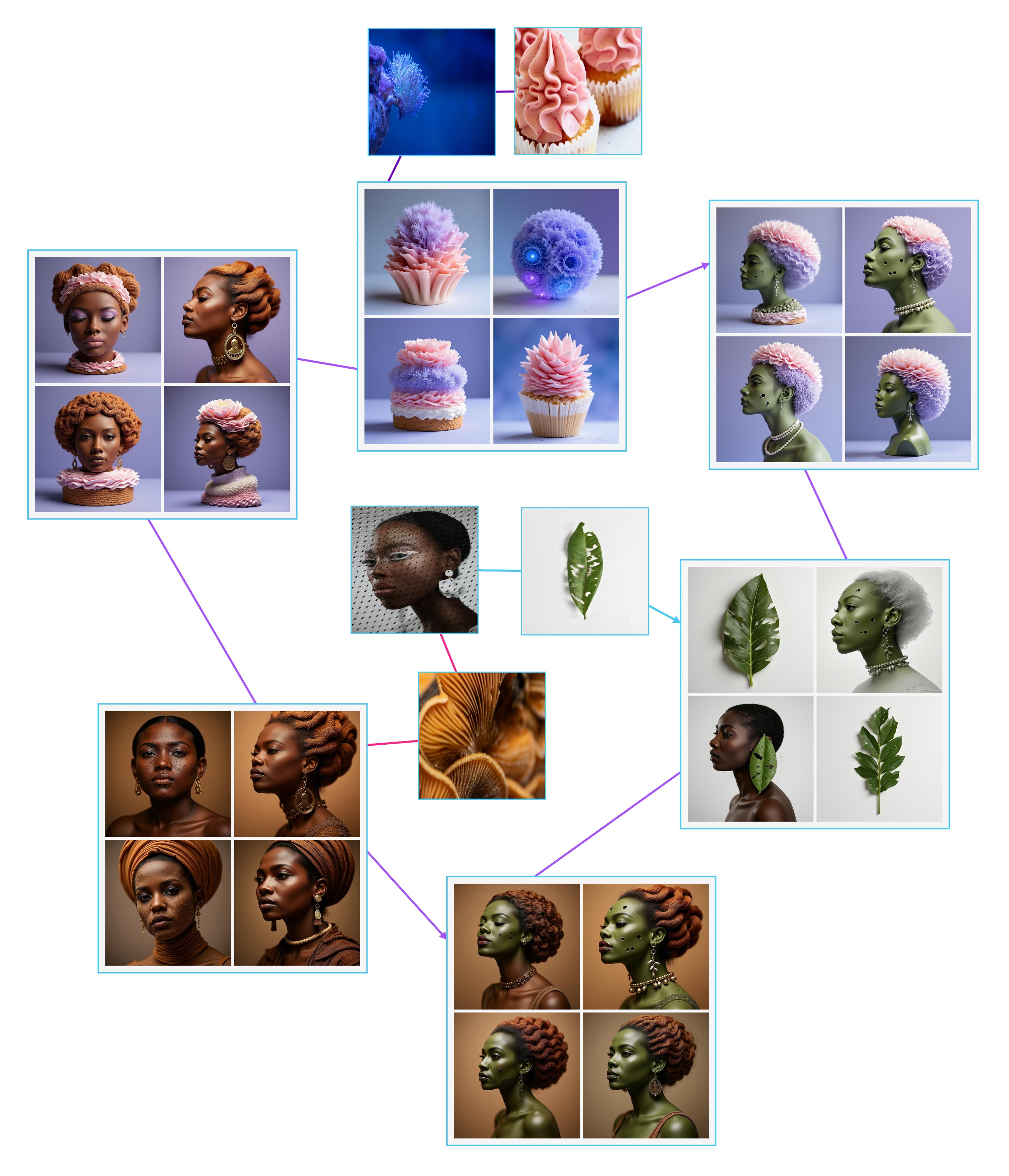}
    \vspace{-0.5cm}
    \caption{Iterative exploration. Outputs can serve as inputs for further combination. A cupcake paired with coral produces frosting with organic, anemone-like texture (top grid). A portrait paired with a leaf yields green skin and botanical patterns (middle grid); paired with jellyfish produces bioluminescent figures (bottom left grid); paired with fungi creates warm tones and sculptural, ruffled hair (bottom middle grid). The rightmost grids show further iterations: combining the portrait-leaf output with the cupcake-coral output produces figures with green skin and fluffy pink hair (top right); combining with the fungi output yields warm-toned portraits with layered, textured hair (bottom right). Each iteration accumulates visual qualities from multiple sources.}
    \label{fig:ours_demo_2}
    \vspace{-1cm}
\end{figure*}

\begin{figure*}
    \centering
    \includegraphics[width=1\linewidth]{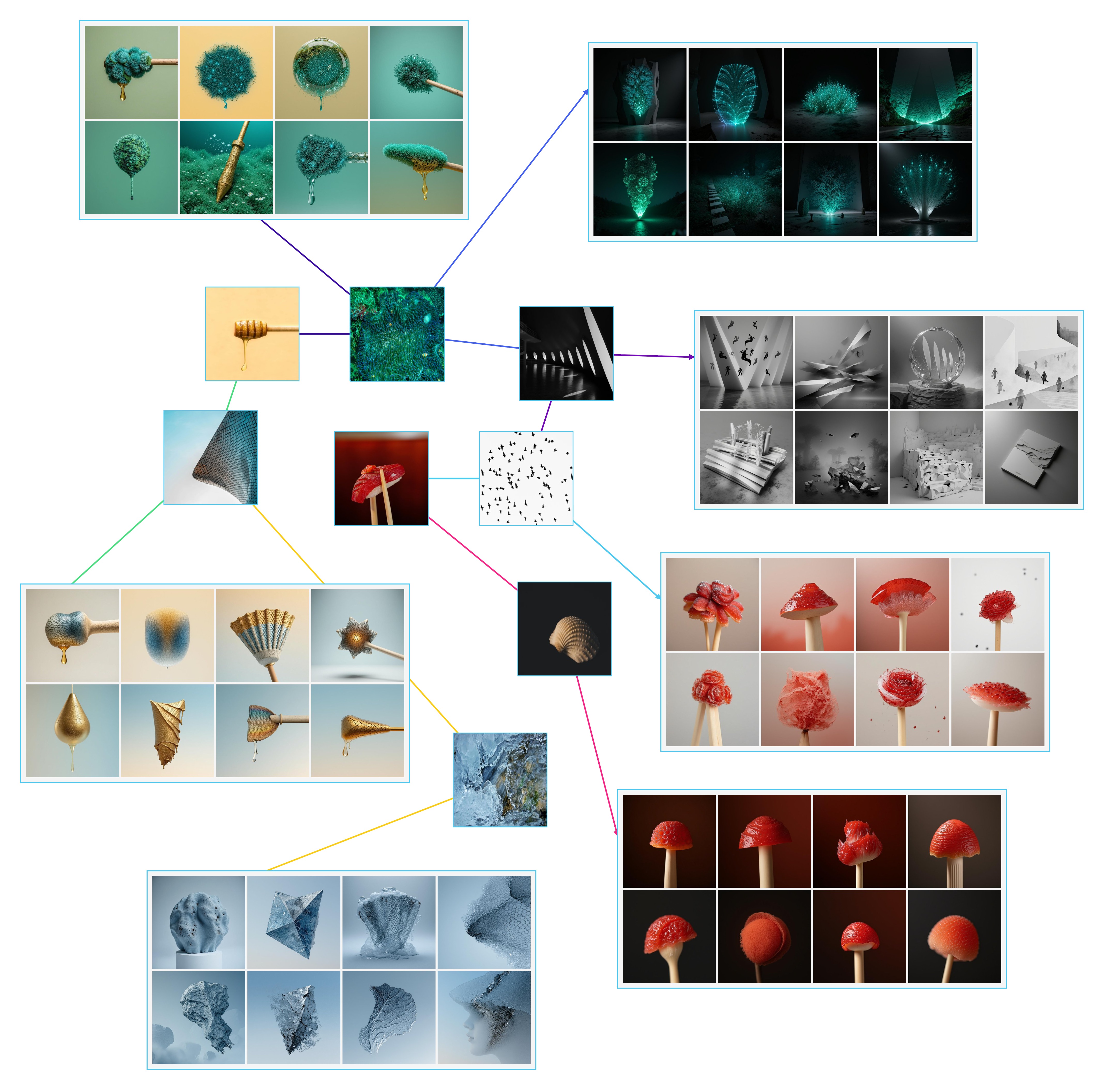}
    \vspace{-0.3cm}
    \caption{Exploration canvas. We present results in the format of an infinite canvas, reflecting how we envision the method being used in practice. Input images (center) are paired with different references, producing grids of outputs generated with varying random seeds. The canvas structure allows users to browse combinations, compare variations, and branch out from promising results, supporting open-ended exploration rather than converging on a single output.}
    \vspace{1cm}
    \label{fig:ours_demo}
\end{figure*}

\section{Experiments}
\subsection{Visual Exploration Results}
We first present qualitative examples that illustrate how our method can facilitate visual exploration. \Cref{fig:teaser,fig:results_multiple_seeds,fig:ours_demo,fig:ours_demo_2}, show results generated by our method. In \Cref{fig:teaser}, a cupcake combined with four different references (a mineral texture, a flowing dress, a sea anemone, and a receding corridor) yields distinct visual transformations, with the frosting adopting crystalline layering, fabric-like folds, organic branching, or architectural geometry depending on the input. In~\Cref{fig:results_multiple_seeds} we show how our method can implicitly produce different interpretations of the given input images when varying the seed, a property well-suited to exploratory workflows.

\begin{figure*}[t]
    \centering
    \setlength{\tabcolsep}{3pt}
    \renewcommand{\arraystretch}{0}
    
    \begin{tabular}{cc@{\hspace{18pt}}c@{\hspace{8pt}}c@{\hspace{8pt}}c@{\hspace{8pt}}c}
    \toprule
    \multicolumn{2}{c}{Inputs} & Flux.1 Kontext & Qwen-Image-2511 & Nano Banana & Ours \\
    \midrule

    \includegraphics[width=0.15\textwidth,height=0.15\textwidth,keepaspectratio=false]{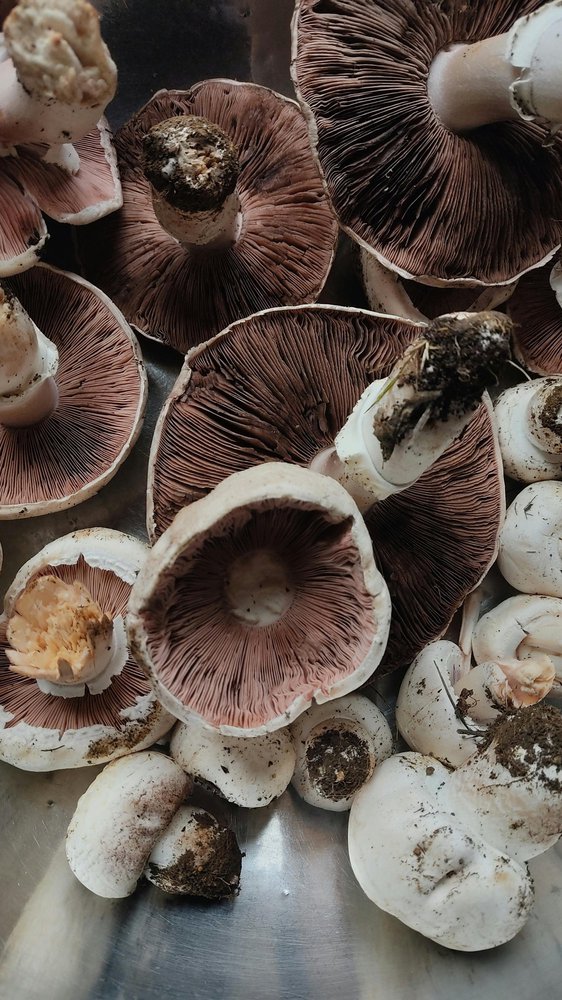} &
    \includegraphics[width=0.15\textwidth,height=0.15\textwidth,keepaspectratio=false]{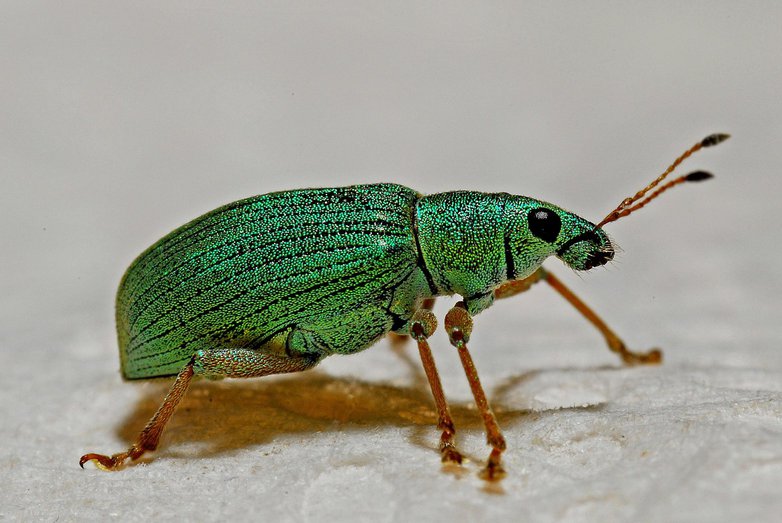} &
    \includegraphics[width=0.15\textwidth]{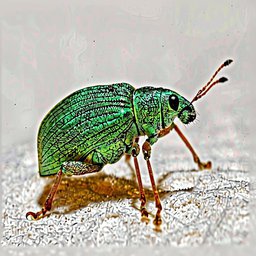} &
    \includegraphics[width=0.15\textwidth]{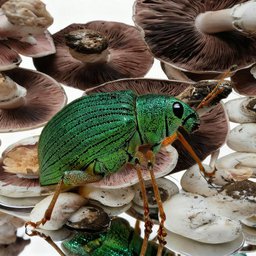} &
    \includegraphics[width=0.15\textwidth]{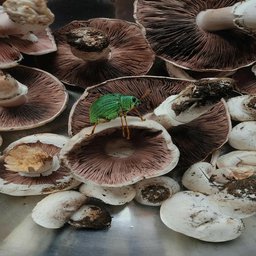} &
        \includegraphics[width=0.15\textwidth]{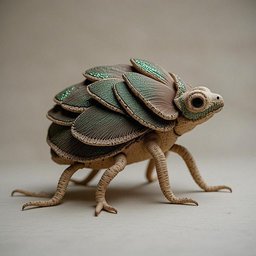} 
     \\
    \midrule

    \includegraphics[width=0.15\textwidth,height=0.15\textwidth,keepaspectratio=false]{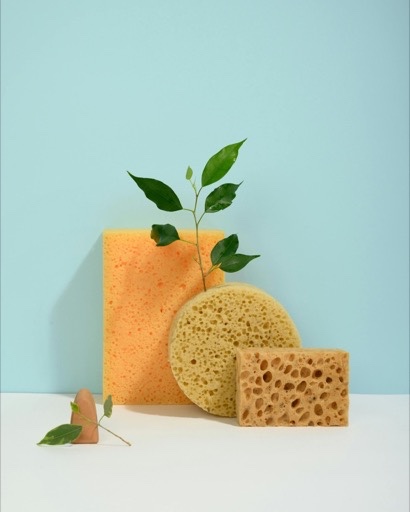} &
    \includegraphics[width=0.15\textwidth,height=0.15\textwidth,keepaspectratio=false]{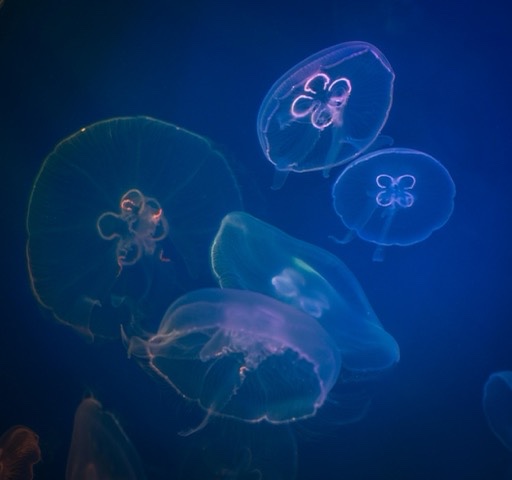} &
    \includegraphics[width=0.15\textwidth]{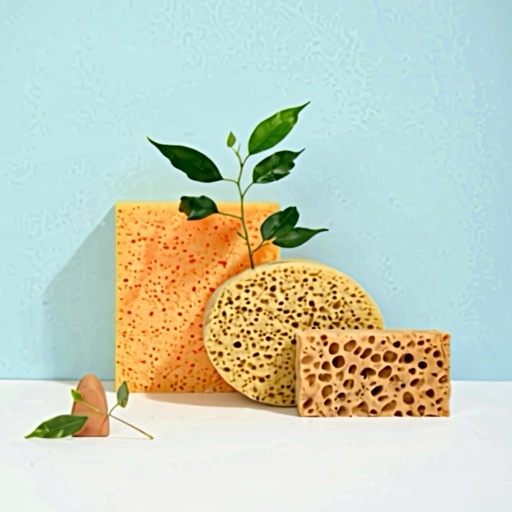} &
    \includegraphics[width=0.15\textwidth]{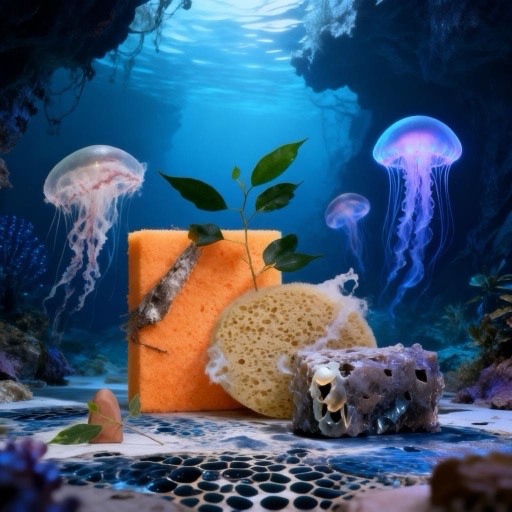} &
    \includegraphics[width=0.15\textwidth]{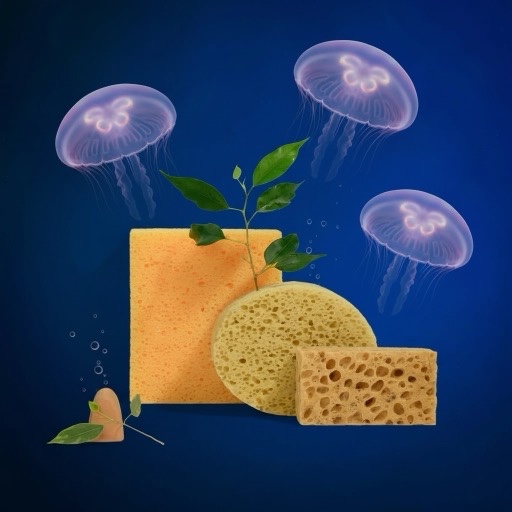} &
        \includegraphics[width=0.15\textwidth]{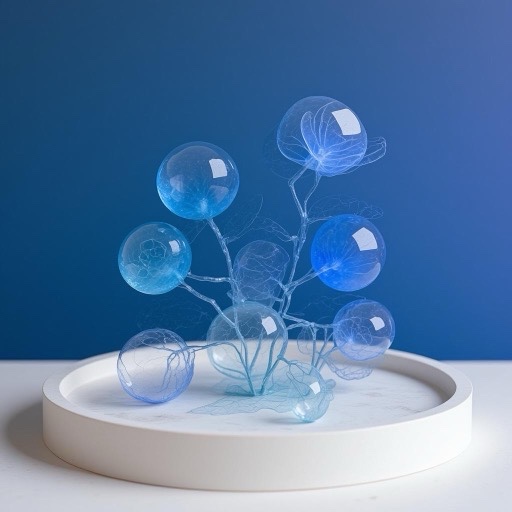}
     \\
    \midrule

    \includegraphics[width=0.15\textwidth,height=0.15\textwidth,keepaspectratio=false]{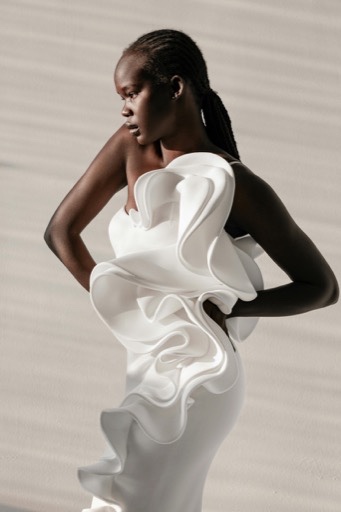} &
    \includegraphics[width=0.15\textwidth,height=0.15\textwidth,keepaspectratio=false]{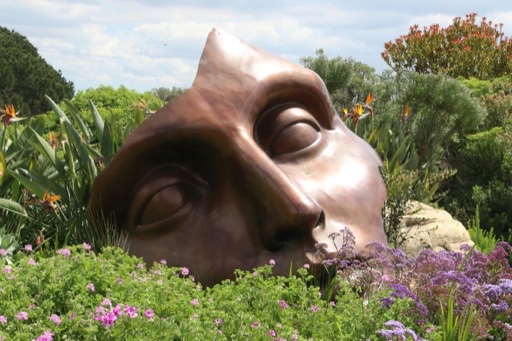}
    &
    \includegraphics[width=0.15\textwidth]{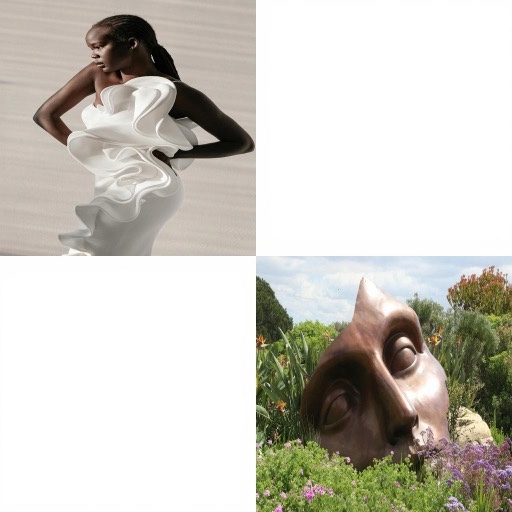} &
    \includegraphics[width=0.15\textwidth]{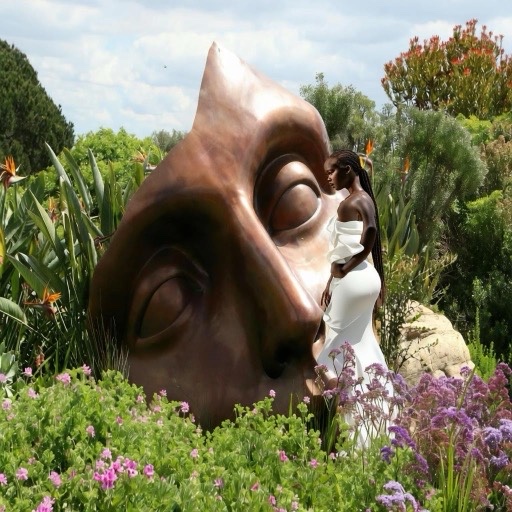} &
    \includegraphics[width=0.15\textwidth]{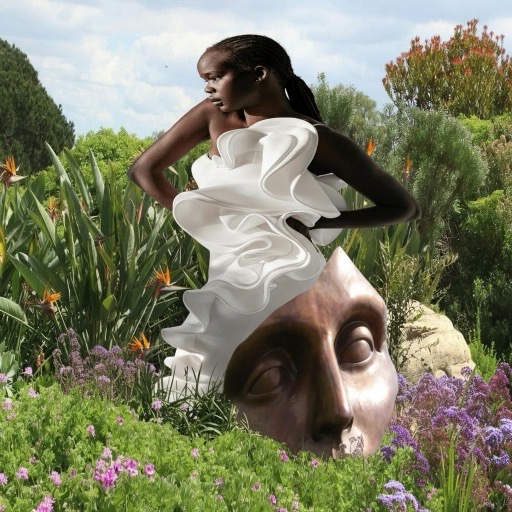} &
    \includegraphics[width=0.15\textwidth]{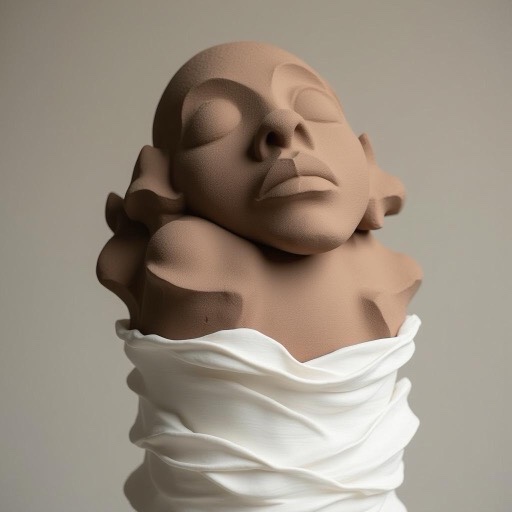} 
     \\

    \bottomrule
    \end{tabular}
    \caption{Qualitative comparison of visual combinations. Baseline methods often produce trivial combinations: direct copying of the inputs (e.g., Flux reproducing the input layout in the first two rows and copying the grid input in the third row), or object insertion (e.g., Nano Banana inserting the insect intro the mushrooms scene in the first row).
    In contrast, our method produces images in which visual cues from both inputs are integrated into a single coherent form. }
    \vspace{-0.2cm}
    \label{fig:qual_single_image}
\end{figure*}

In \Cref{fig:ours_demo} and \Cref{fig:ours_demo_2}, we present additional results in the form of an exploration canvas, reflecting how we envision the method being used in practice. This canvas illustrates a potential workflow where a user might collect reference images, combine them in different pairings, and branch out from promising results. 
\Cref{fig:ours_demo} shows diverse input images (a honey dipper, a woven mesh, a mineral texture, a flock of birds, a strawberry mushroom, a shell) connected to grids of multiple outputs generated from their pairings. For example, a honey dipper paired with underwater flora transforms into mossy organic forms (top left); paired with a woven mesh, it takes on golden shell-like qualities (middle left). 
Using different seeds provides diversity, which is key to supporting exploration: rather than producing a single ``correct'' combination, the model generates a space of options that users can browse, allowing unexpected connections to emerge without requiring users to articulate what they are looking for.
More results and an interactive demo are available in the supplementary material.

\subsection{Evaluating Visual Combinations}
Here we evaluate our method's ability to produce meaningful, non-trivial visual combinations. We curate a benchmark of 41 images spanning six categories (architecture, fashion, food, nature, sea creatures, and other), sourced from Pexels, to cover a range of concepts, styles, structures, and materials. We randomly sample 99 cross-category pairs, ensuring each image appears in at least one pair. 

Since no existing method is explicitly designed to generate non-trivial visual combinations from image pairs, we compare against the strongest publicly available image-conditioned generation models. 
Specifically, we evaluate Flux.1 Kontext \cite{labs2025flux1kontextflowmatching}, a large-scale image editing model that also serves as our backbone; Qwen-Image-2511 \cite{wu2025qwen}, a recent multimodal model with strong visual understanding capabilities; and Nano Banana \cite{nanobanana2025}, Google's image generation and editing model. 
For Flux.1 Kontext, we use the same constant input prompt used in our training, matching its single input design.
For Qwen and Nano Banana, we provide both images along with the prompt: \textit{``Combine the two images into a novel and non-trivial image inspired by them.''} For all methods, we generate four random outputs per pair using different seeds, resulting in 396 images per method in total.

Representative results are shown in \Cref{fig:qual_single_image}. For visualization clarity, we display one output per method for each input pair. All generated samples, including all seeds, are provided in the supplementary material.
Flux.1 Kontext tends to copy the input images, either completely or by reproducing its input grid-like arrangement. Qwen-Image-2511 often defaults to trivial combinations. Nano Banana performs best among the baselines, however, it often defaults to object-level placement without transferring deeper visual qualities.
In contrast, our method produces coherent, non-trivial combinations, integrating visual aspects from both inputs. The beetle takes on the mushroom's layered patterns, the sponge and jellyfish merge into delicate, bubble-like forms, and the portrait and sculpture blend into a figure where skin and fabric share the same materiality. These connections are not immediately obvious, they require a close look, and invite interpretation, which is what makes them useful for creative work. They surface relationships a user might not have thought to look for.

\begin{figure*}[t]
\centering
\setlength{\tabcolsep}{1pt}
\renewcommand{\arraystretch}{1.0}

\begin{tabular}{c c c c c c c c c}
\toprule
Inputs &
\multicolumn{2}{c}{Flux.1 Kontext} &
\multicolumn{2}{c}{Qwen-Image-2511} &
\multicolumn{2}{c}{Nano Banana} &
\multicolumn{2}{c}{Ours} \\
\midrule \noalign{\vskip-8pt}

\raisebox{-\height}{\begin{tabular}[t]{@{}c@{}}
\includegraphics[width=0.065\linewidth,height=0.065\linewidth]{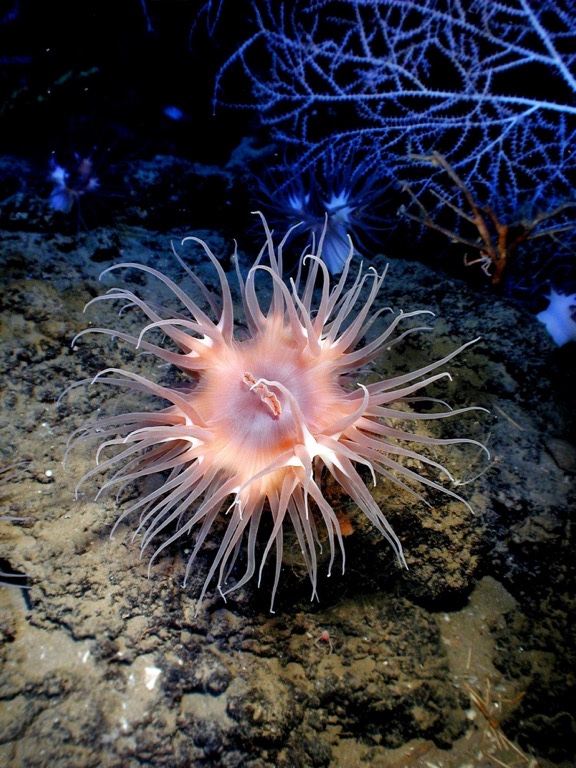} \\
\includegraphics[width=0.065\linewidth,height=0.065\linewidth]{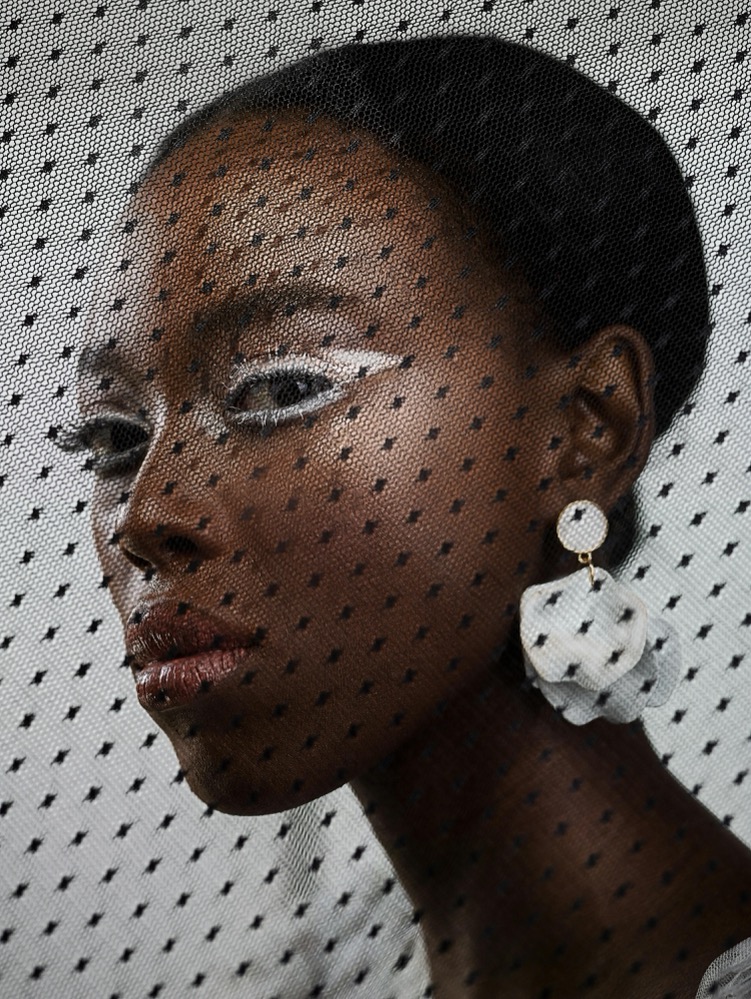}
\end{tabular}}
&
\raisebox{-\height}
{\begin{tabular}[t]{@{}c@{}}
\includegraphics[width=0.1\linewidth,height=0.1\linewidth]{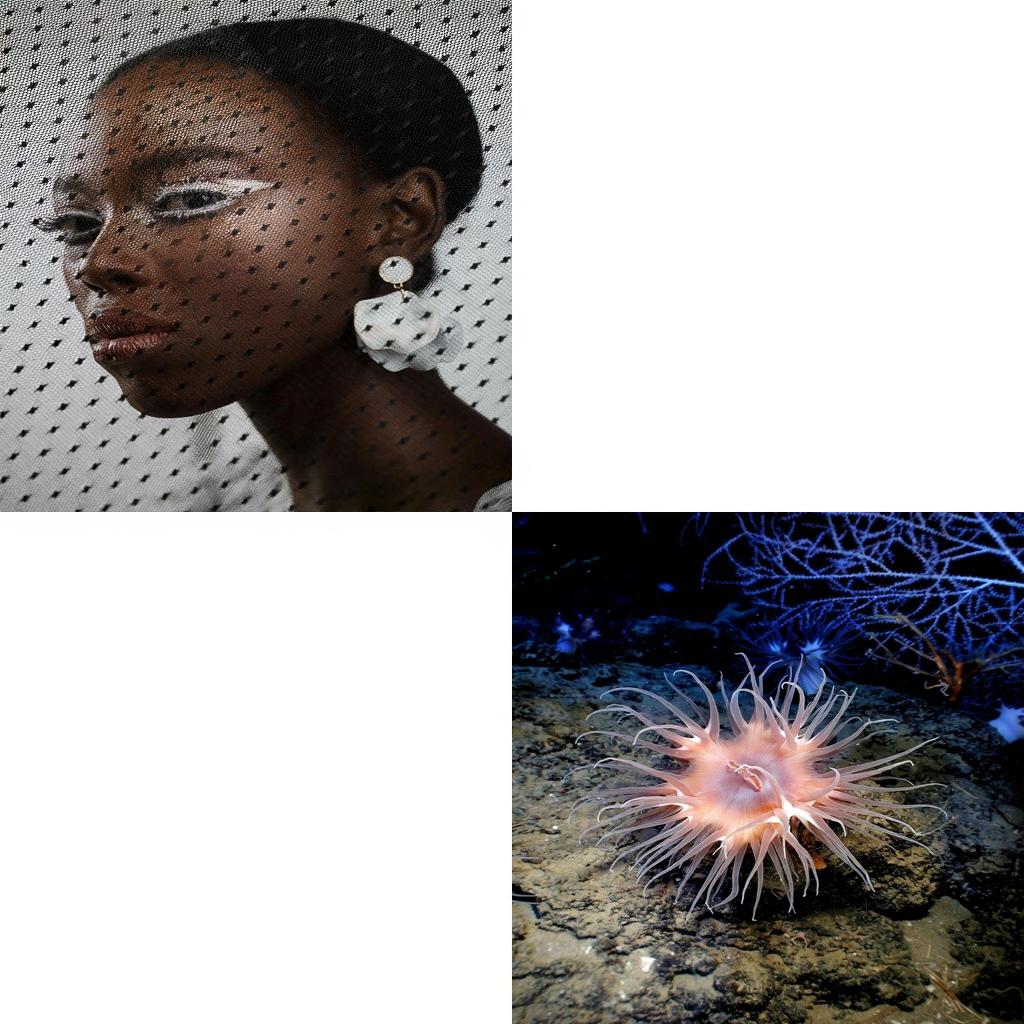} \\[-4pt]
\scriptsize(3 words)
\end{tabular}}
&
\raisebox{-\height}{\parbox[t]{0.05\linewidth}{\raggedright\scriptsize
\textbullet~copy entire grid
}}
&
\raisebox{-\height}
{\begin{tabular}[t]{@{}c@{}}
\includegraphics[width=0.1\linewidth,height=0.1\linewidth]{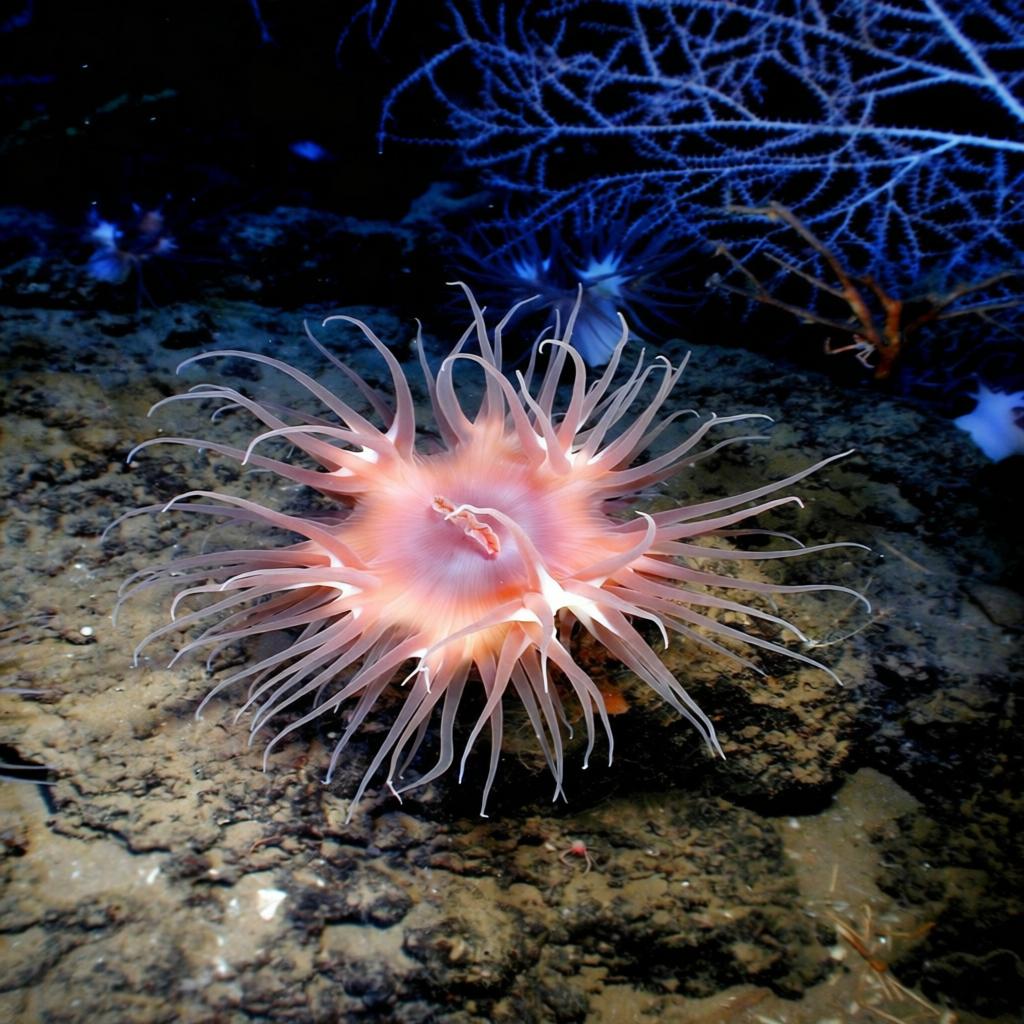} \\[-4pt]
\scriptsize(3 words)
\end{tabular}}
&
\raisebox{-\height}{\parbox[t]{0.05\linewidth}{\raggedright\scriptsize
\textbullet~copy $\langle$image2$\rangle$
}}
&
\raisebox{-\height}
{\begin{tabular}[t]{@{}c@{}}
\includegraphics[width=0.1\linewidth,height=0.1\linewidth]{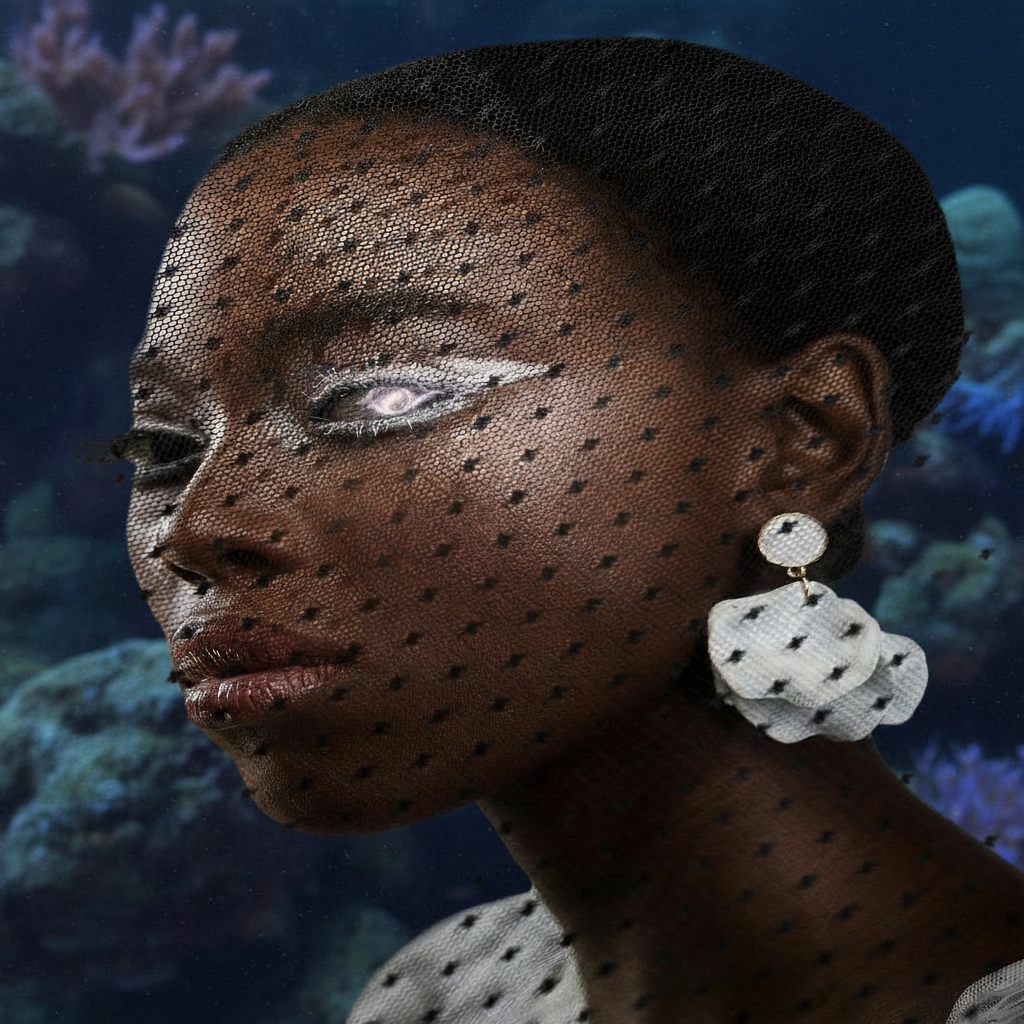} \\[-4pt]
\scriptsize(46 words)
\end{tabular}}
&
\raisebox{-\height}{\parbox[t]{0.13\linewidth}{\raggedright\scriptsize
\textbullet~Extract the woman with the dotted veil and earring from image 1.
\textbullet~Place the extracted woman into the background from image 2.
\textbullet~Modify the white eyeliner of the woman to incorporate the pinkish glow from the center of the anemone in image 2.
}}
&
\raisebox{-\height}
{\begin{tabular}[t]{@{}c@{}}
\includegraphics[width=0.1\linewidth,height=0.1\linewidth]{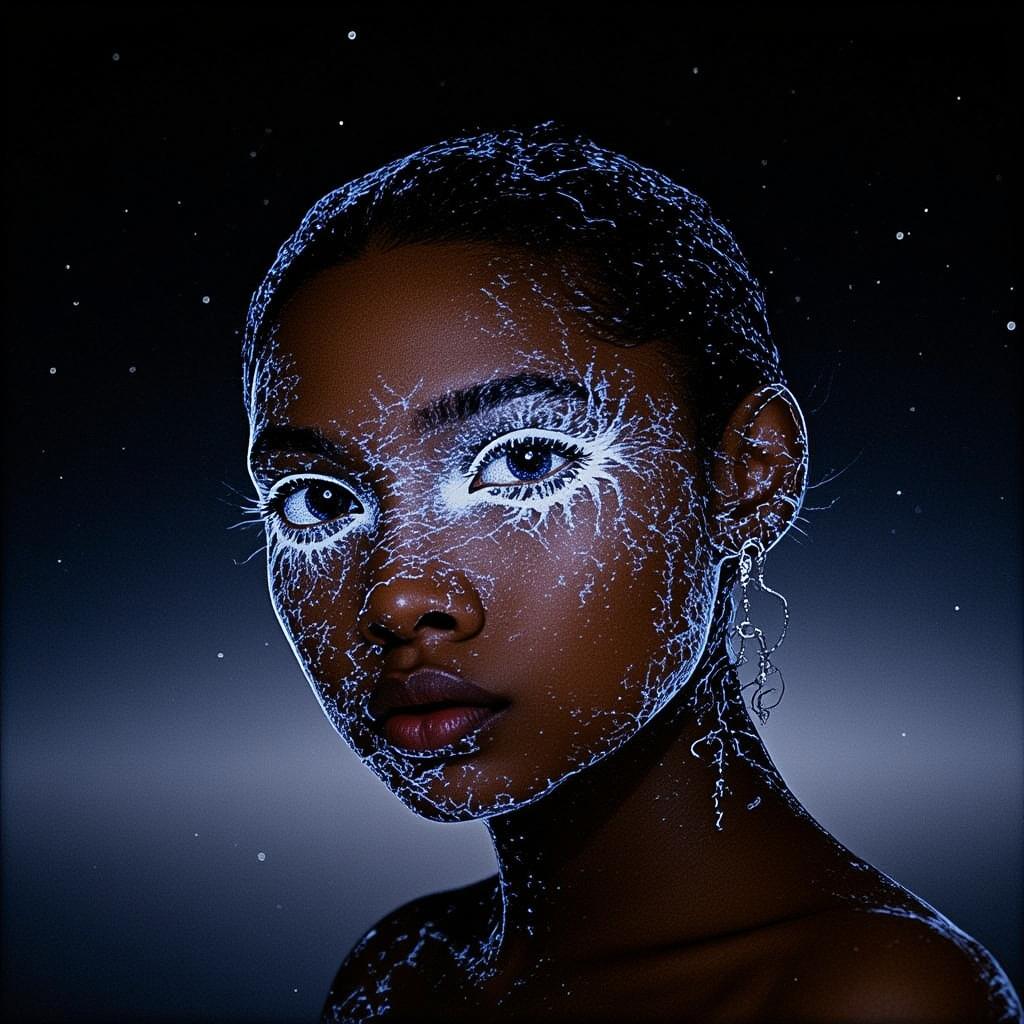} \\[-4pt]
\scriptsize(67 words)
\end{tabular}}
&
\raisebox{-\height}{\parbox[t]{0.23\linewidth}{\raggedright\scriptsize
\textbullet~Extract the woman's head from image1, removing the dotted veil.
\textbullet~Apply the intricate, glowing patterns and texture of the anemone and coral from image2 onto her skin, hair, and earring.
\textbullet~Replace the background with the dark, deep-sea environment from image2, including its subtle glow and ambient particles.
\textbullet~Enhance the white eye makeup from image1 to glow and integrate with the new organic patterns.
}}
\\

\bottomrule
\end{tabular}
\vspace{-0.2cm}
\caption{Description complexity analysis. We show the LLM's descriptions for reconstructing the target image from the two inputs. One can see that as the connection becomes more complex and non-literal, the LLM naturally scales to longer and more sophisticated descriptions.}
\label{fig:caption_length_comparison_single}
\end{figure*}

\paragraph{Quantitative Evaluation.}
Standard perceptual similarity metrics such as CLIP cosine similarity \cite{Radford2021LearningTV} or DreamSim \cite{fu2023dreamsim} reward visual similarity.  In our setting, this means outputs that simply preserve or insert elements from the inputs score higher than those that transform and recombine them. Moreover, such metrics are not designed to measure whether a combination is non-trivial or unique.
We therefore propose using \emph{description complexity} as an alternative measure. We observe that trivial combinations can often be explained in a few words (``place object A into scene B''), whereas non-trivial combinations require longer descriptions to articulate.
This observation aligns with research linking description length to complexity. Specifically, Kolmogorov complexity formalizes the idea that an object's complexity corresponds to the length of its shortest description \cite{kolmogorov1965, li1997vitanyi}, and Sun and Firestone \cite{sun2021speaking} showed that verbal description length tracks the information-theoretic complexity of visual stimuli.

\begin{table}[t]
\centering
\small
\caption{Caption length comparison for combination complexity. We measure the word count of VLM-generated descriptions explaining how to recreate the output from the inputs. Higher word counts indicate more complex, non-trivial combinations. We also report the percentage of outputs classified as trivial patterns.}
\label{tab:caption_length_comparison}
\vspace{-0.3cm}
\begin{tabular}{lcccc}
\toprule
Method & Word Count & Copy & Insertion & Split \\
\hline
Flux.1 Kontext & $23.5 \pm 21.4$ & $2.8\%$ & $0.3\%$ & $85.4\%$ \\
Qwen-Image & $37.4 \pm 19.2$ & $16.2\%$ & $18.9\%$ & $10.6\%$ \\
Nano Banana & $42.9 \pm 15.6$ & $9.1\%$ & $19.7\%$ & $0.3\%$ \\
\midrule
Ours & $\mathbf{54.8 \pm 12.5}$ & $\mathbf{2.3\%}$ & $\mathbf{0.0\%}$ & $\mathbf{1.5\%}$ \\
\bottomrule
\end{tabular}
\end{table}

To apply this, we prompt Gemini 2.5 Flash to describe how each output image could be reconstructed from its two source images, using a fixed instruction format across all methods (see supplementary material for more details). We use word count as a proxy for the complexity of the visual relationship.
The average word counts across all 99 image pairs are shown in \Cref{tab:caption_length_comparison}. Our method elicits longer descriptions on average compared to all baselines.
\Cref{fig:caption_length_comparison_single} illustrates the textual descriptions obtained by the LLM across different methods.

We additionally analyze the types of relationships recognized by Gemini, counting observable patterns such as copying (output nearly identical to one input), insertion (placing one element into the other scene), or split composition (inputs placed side by side or in a grid), as demonstrated in \Cref{fig:qual_single_image}. As shown in \Cref{tab:caption_length_comparison}, Flux.1 Kontext defaults to split compositions in most cases, while Qwen and Nano Banana often resort to insertion. Our method rarely triggers any of these categories, indicating that the combinations it produces do not reduce to simple operations.

\vspace{-0.3pt}
\paragraph{User Study.}
To provide additional support that description length serves as a meaningful proxy for combination complexity, we conduct a user study with 35 participants. Each participant was shown an output image alongside its two inputs and asked to classify the relationship between them. The options were: (1) near-duplicate; (2) element insertion; (3) texture or structure transfer; (4) other relationship not captured by the above; and (5) unrelated. We sampled 25 outputs stratified by description length, comprising 11 images from our method and 7 each from Nano Banana and Qwen.

\begin{figure}
    \centering
    \includegraphics[width=0.8\linewidth]{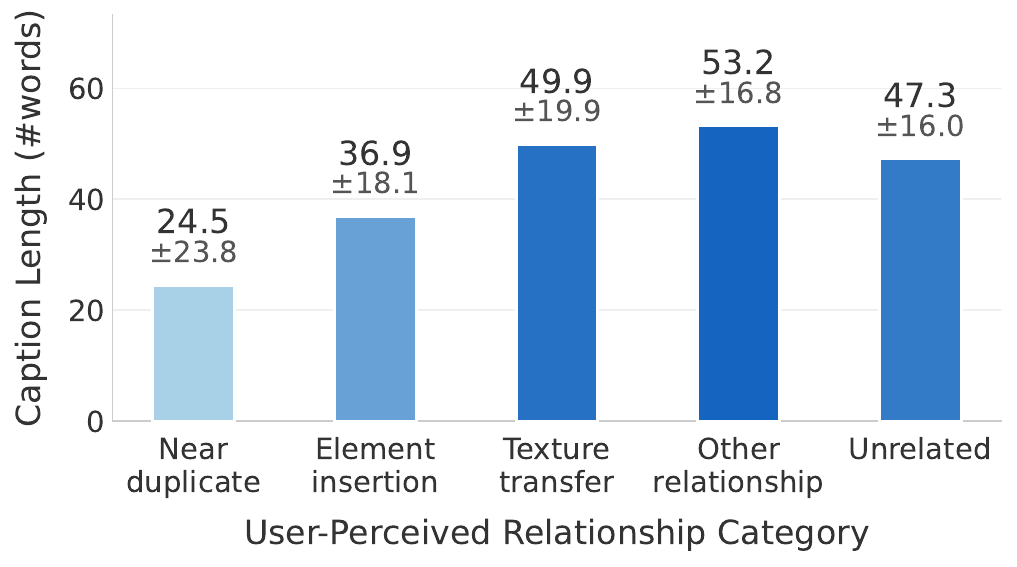}
    \vspace{-0.3cm}
    \caption{User study results. Simple relationships such as duplication or insertion require less words than more complex ones such as texture transferring or other non-canonical relationships.
    }
    \vspace{-0.1cm}
    \label{fig:user_study_results}
\end{figure}

In \Cref{fig:user_study_results} we report the average description length of images assigned by participants to each category. If description length captures combination complexity, we would expect images classified as more complex relationships to have longer descriptions. Indeed, description length increases with combination complexity. Images associated with trivial relationships such as duplication or insertion are more easily described, whereas texture and structure transfer requires explicit specification of which visual properties are extracted and how they are mapped onto another structure. Outputs categorized as ``other relationship'' similarly demand longer explanations, as they involve transformations that are not captured by predefined operations.

\subsection{Decomposition Results}
\begin{figure}
    \centering
    \setlength{\tabcolsep}{5pt}
    \newcommand{\imgw}{0.13\linewidth}
    \renewcommand{\arraystretch}{0.1}

    \begin{tabular}{@{}cccccc@{}}
    \toprule
    & Input & Ours & Ins. Tree & I2I & T2I \\
    \midrule
    \rotatebox{90}{Comp 1} &
    \includegraphics[width=\imgw,height=\imgw,keepaspectratio=false]{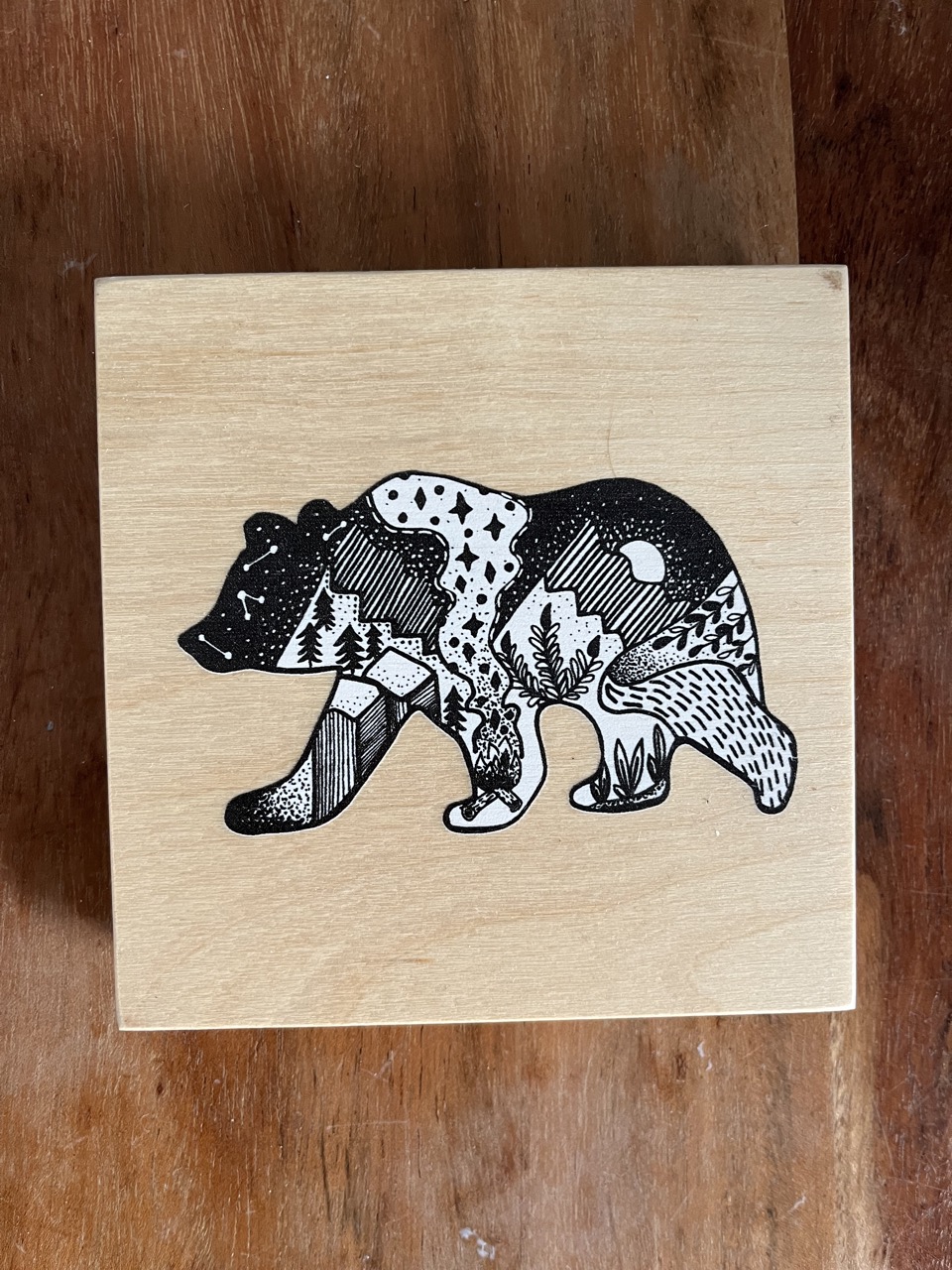} &
    \includegraphics[width=\imgw]{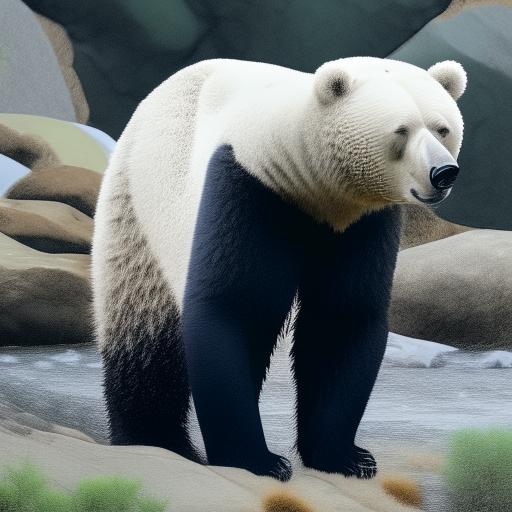} &
    \includegraphics[width=\imgw]{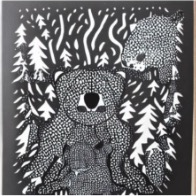} &
    \includegraphics[width=\imgw]{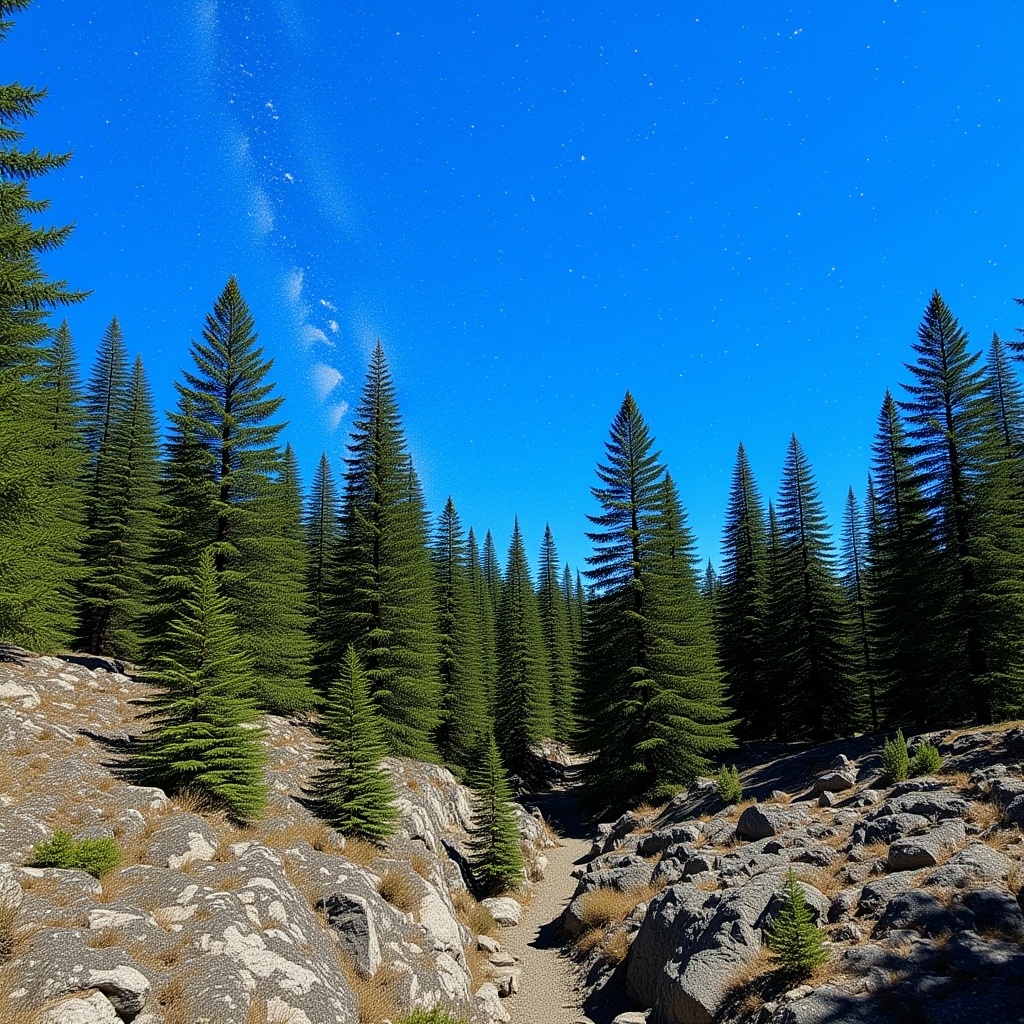} &
    \includegraphics[width=\imgw]{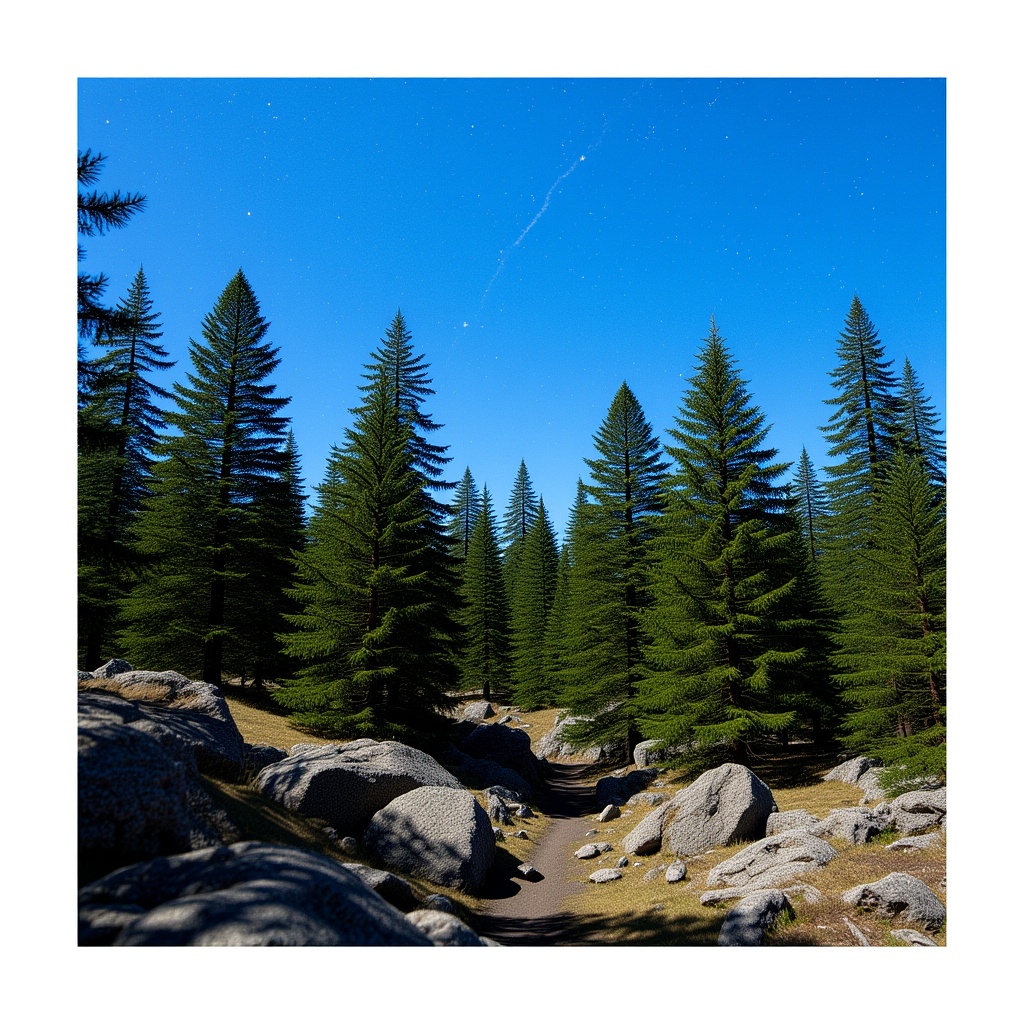} \\
    \rotatebox{90}{Comp 2} &
    &
    \includegraphics[width=\imgw]{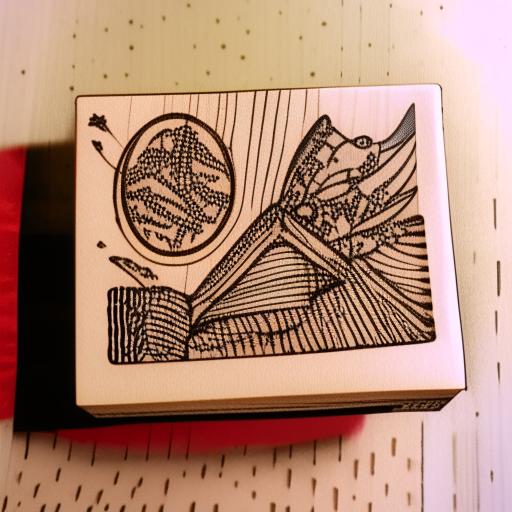} &
    \includegraphics[width=\imgw]{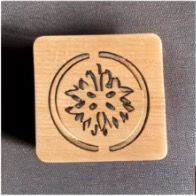} &
    \includegraphics[width=\imgw]{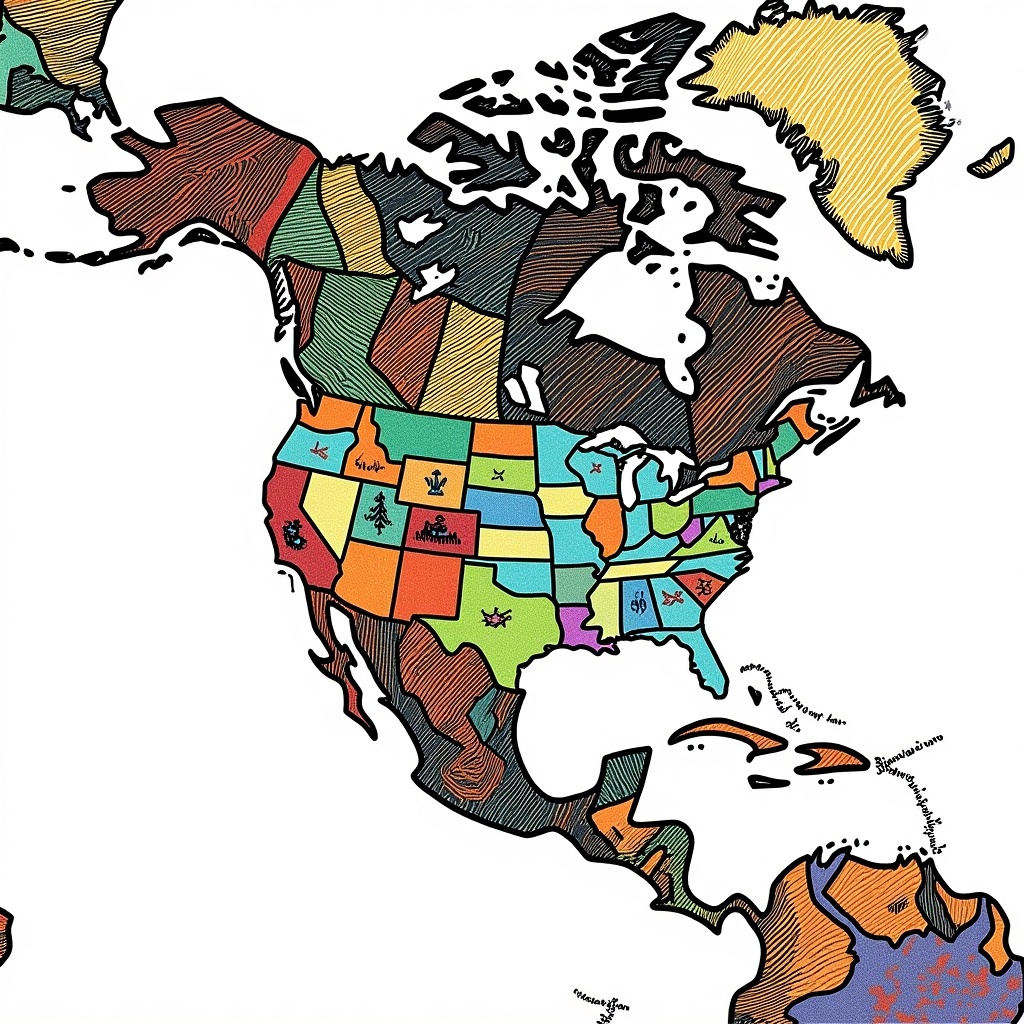} &
    \includegraphics[width=\imgw]{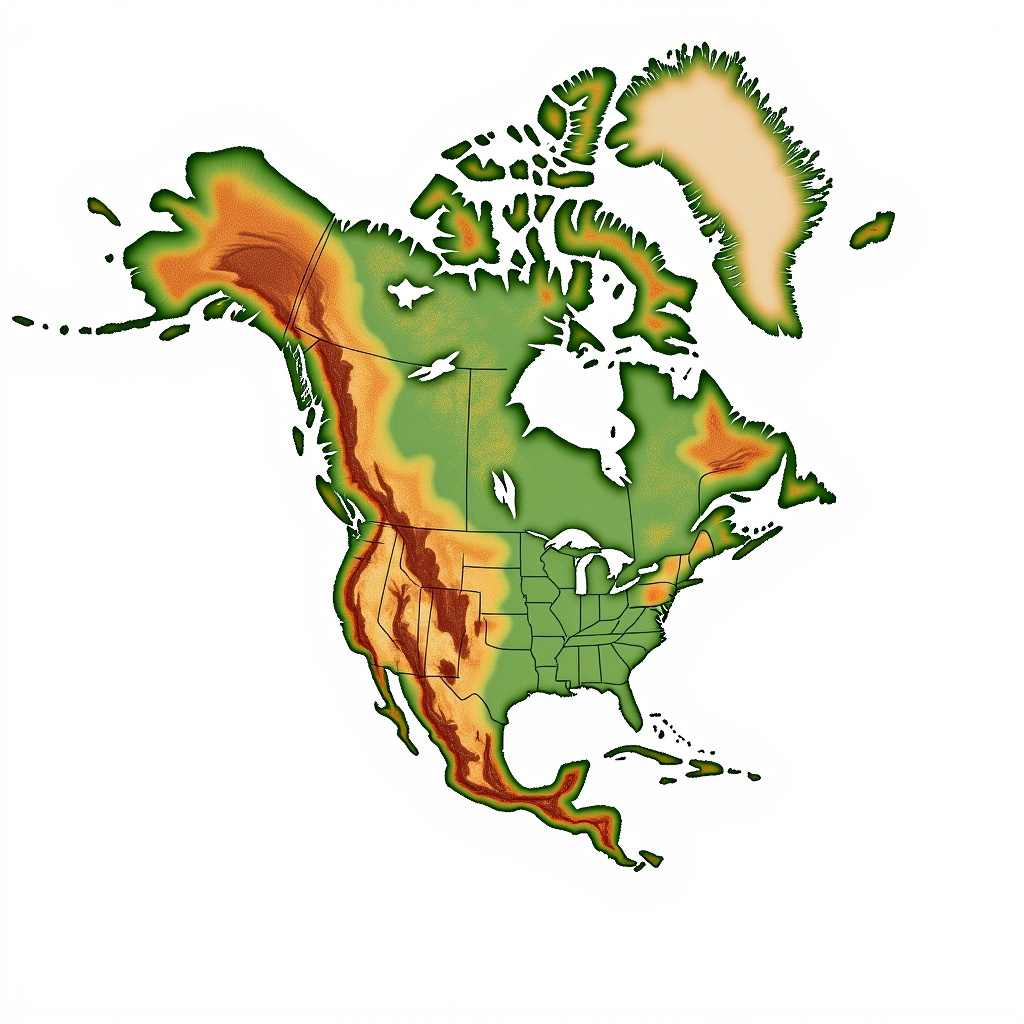} \\
    \midrule
    \rotatebox{90}{Comp 1} &
    \includegraphics[width=\imgw,height=\imgw,keepaspectratio=false]{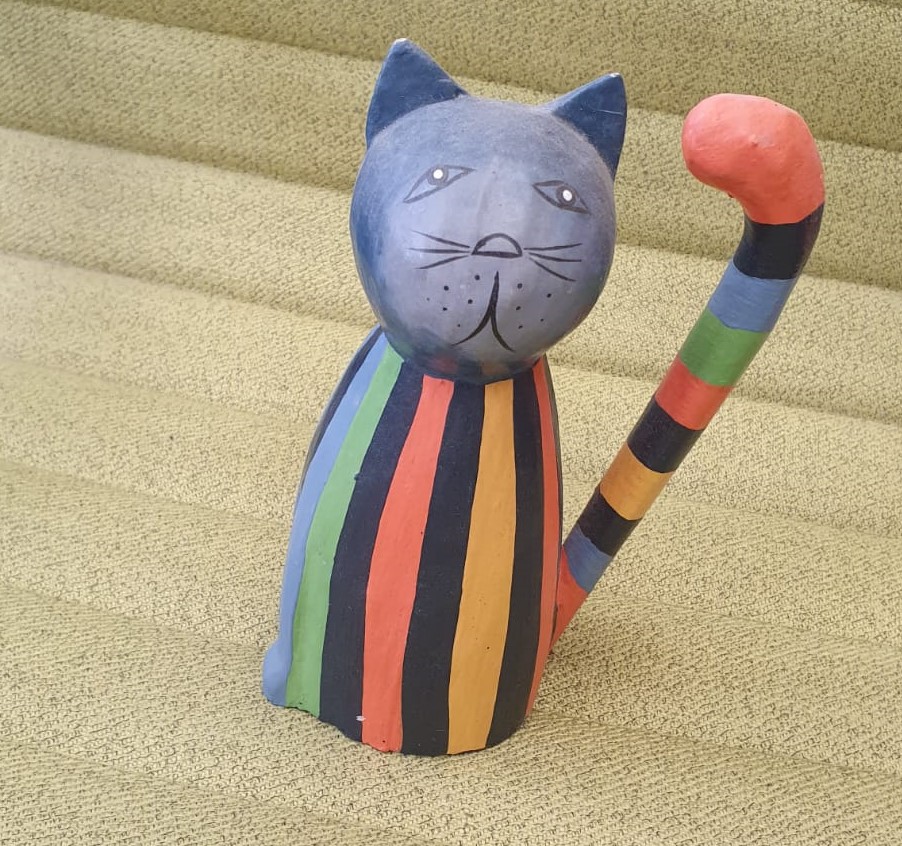} &
    \includegraphics[width=\imgw]{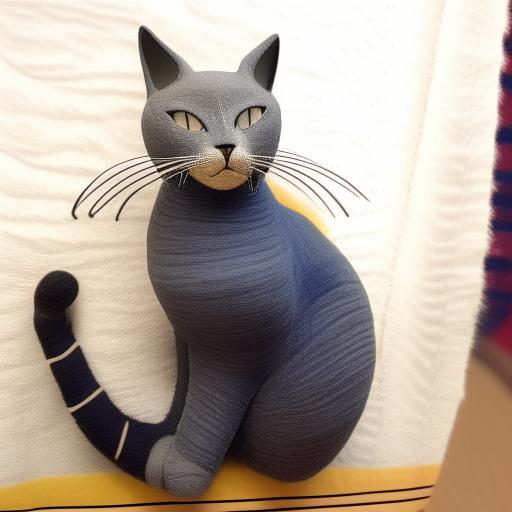} &
    \includegraphics[width=\imgw]{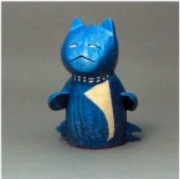} &
    \includegraphics[width=\imgw]{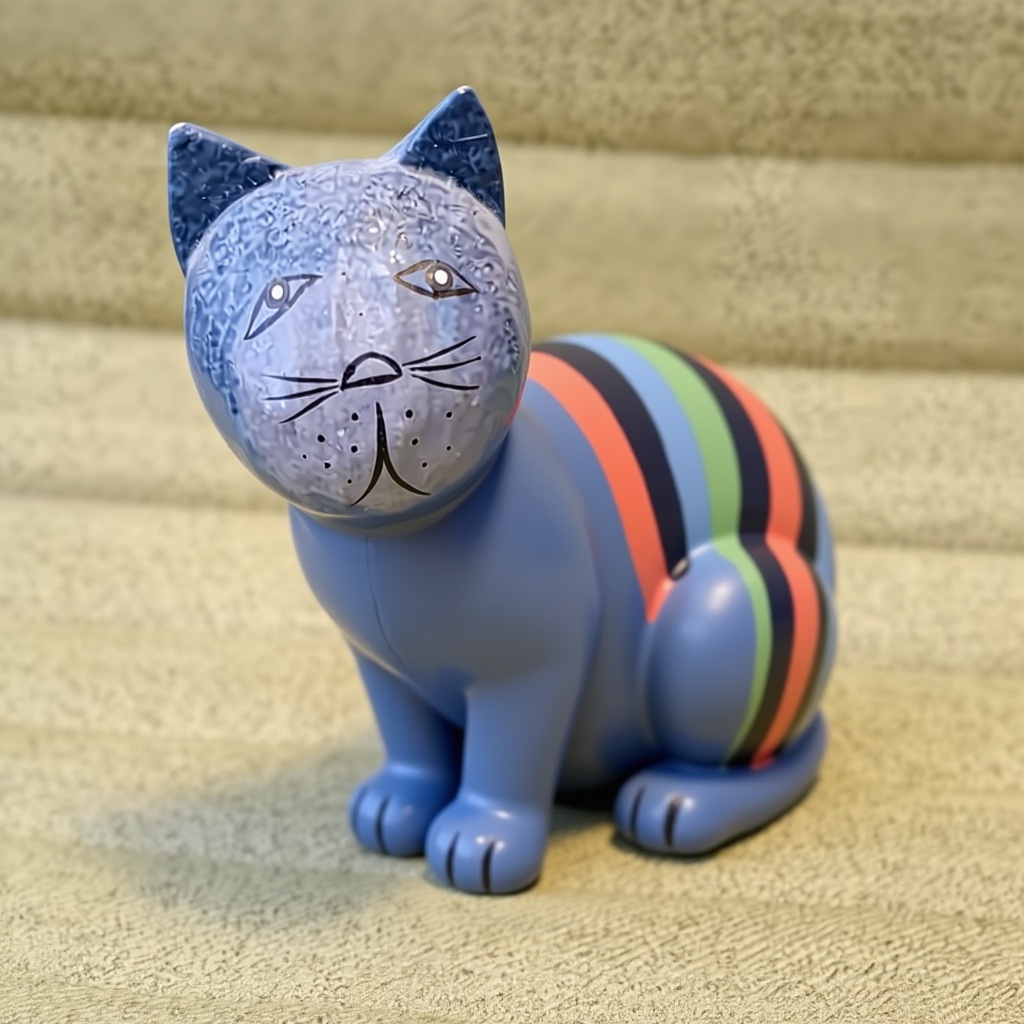} &
    \includegraphics[width=\imgw]{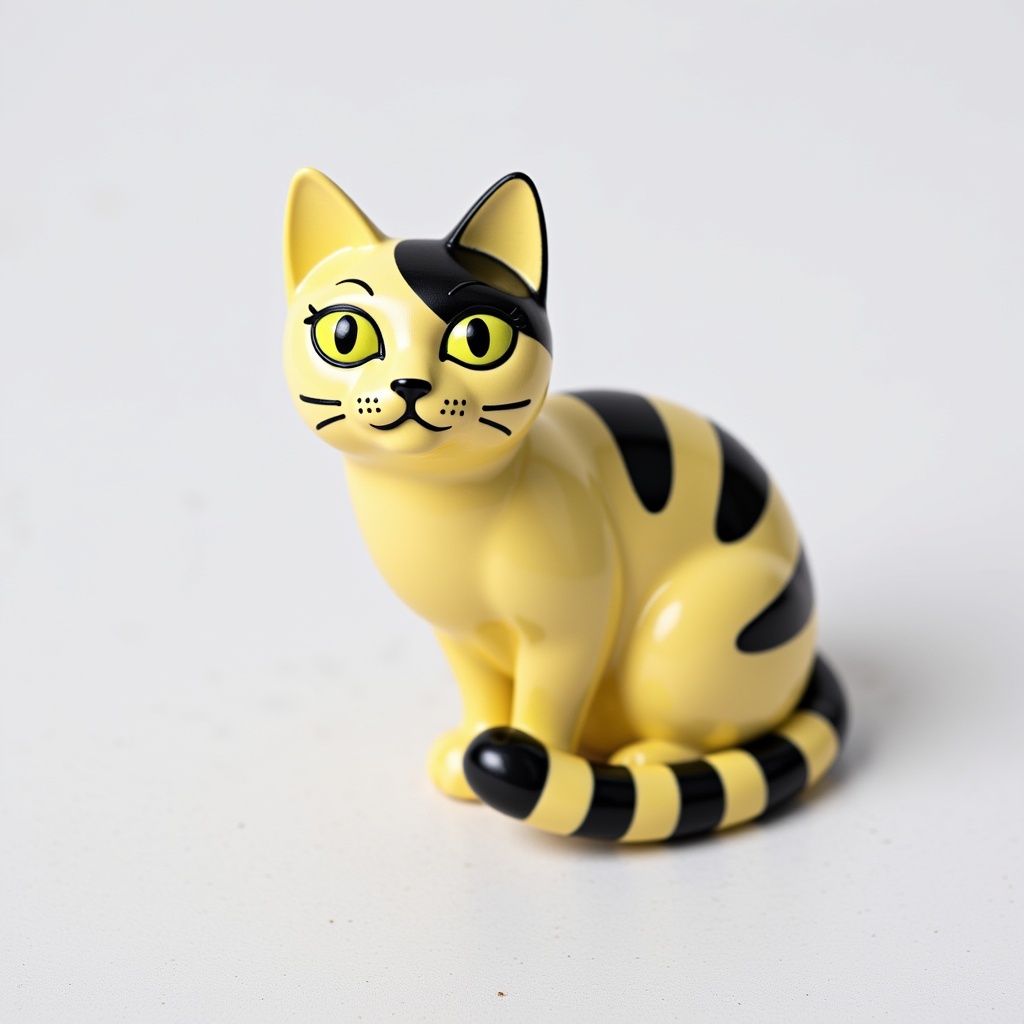} \\
    \rotatebox{90}{Comp 2} &
    &
    \includegraphics[width=\imgw]{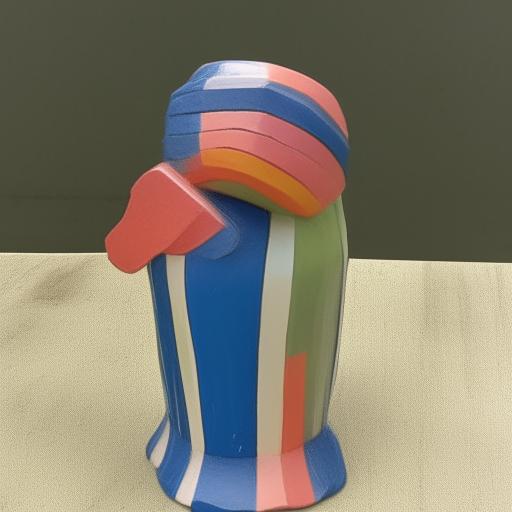} &
    \includegraphics[width=\imgw]{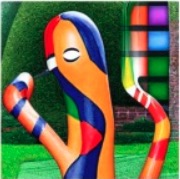} &
    \includegraphics[width=\imgw]{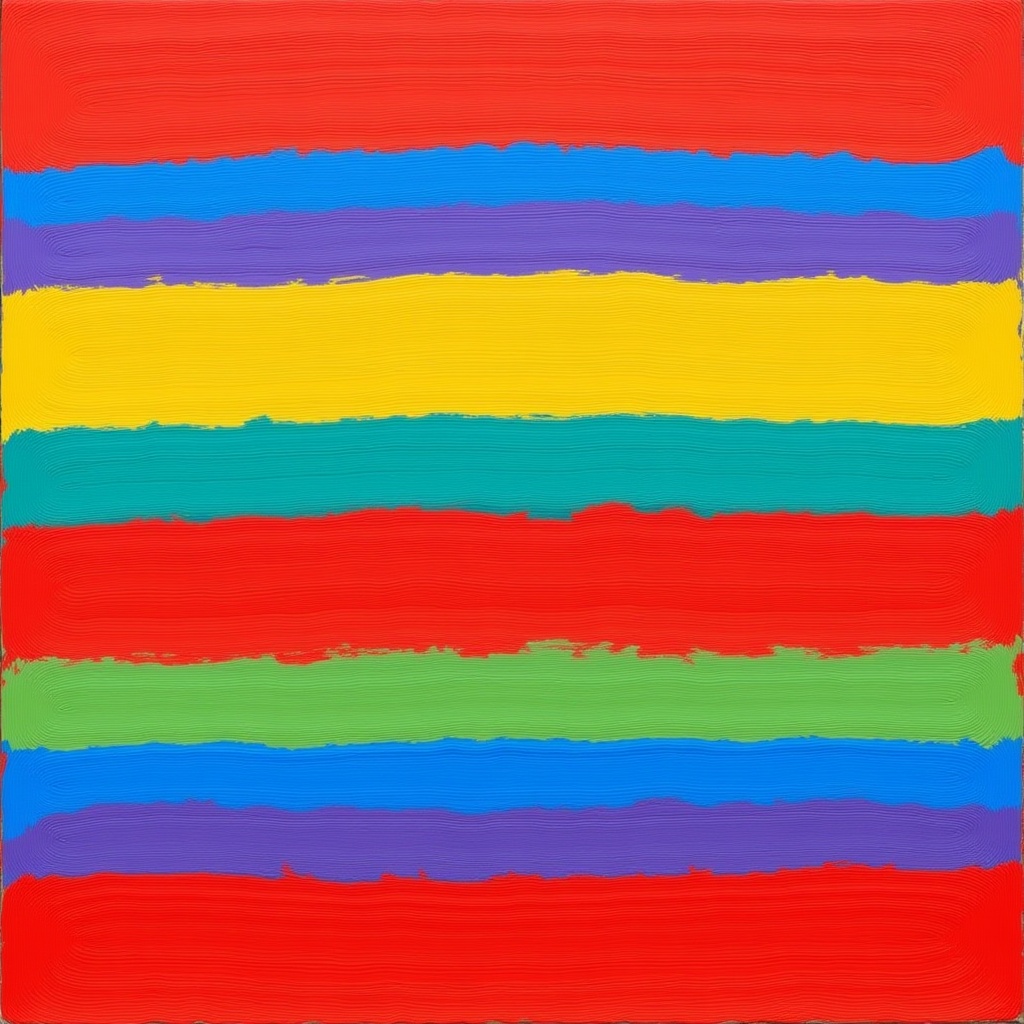} &
    \includegraphics[width=\imgw]{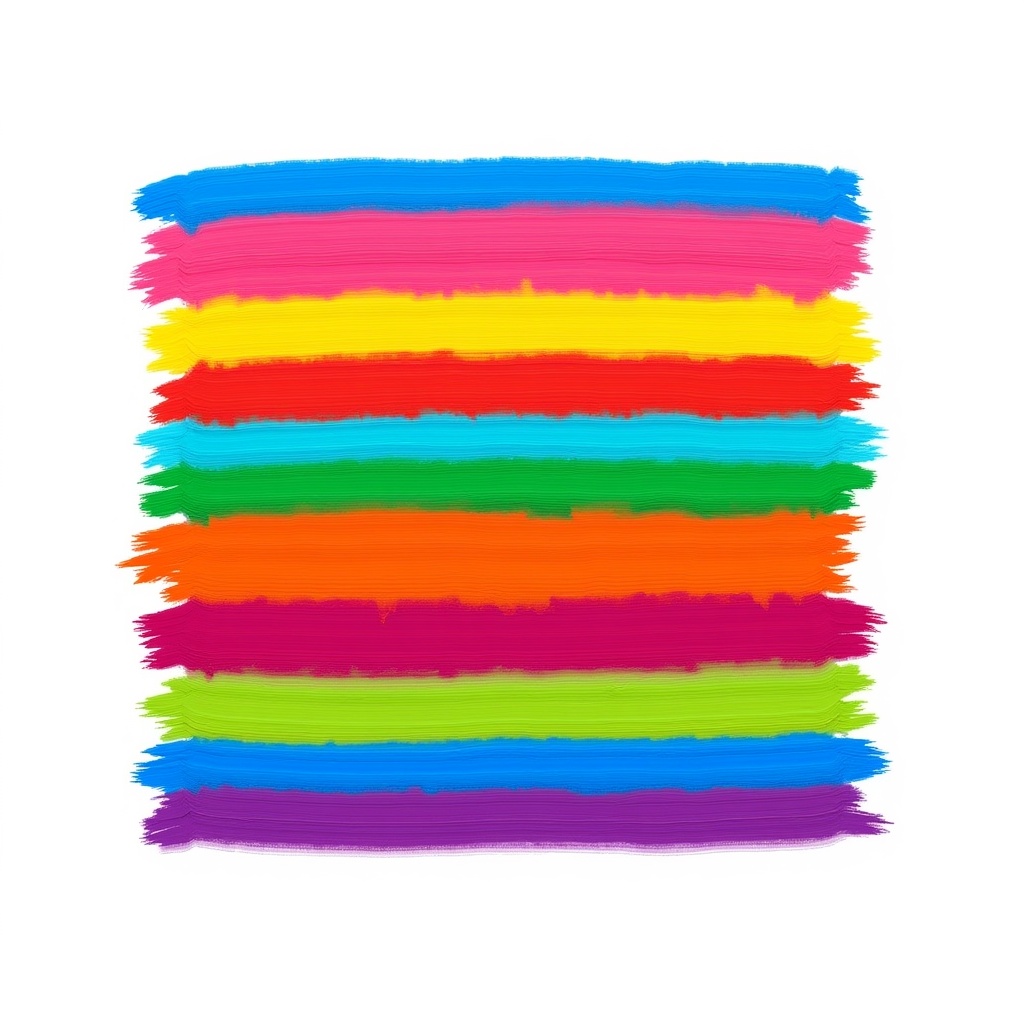} \\
    \bottomrule
    \end{tabular}
    \caption{Decomposition results comparison. Given an input image, we decompose it into two components using four methods. Our approach produces components that capture distinct visual aspects while maintaining semantic relevance. Inspiration Tree relies on time-consuming optimization and requires multiple input images of the concept to converge. T2I and I2I methods often fail to adhere to the visual qualities of the input.}
    \label{fig:decomposition_results_comparison}
\end{figure}

Our decomposition technique is central to our approach as it determines what kinds of relationships the model learns to combine. 
InspirationTree \cite{inspirationtree23} is the only existing method for implicit decomposition beyond style-content separation. However, it is designed for decomposing images of single objects and relies on textual inversion, requiring 4-5 images of the same concept from different viewpoints, over an hour of optimization per concept, and multiple runs to handle instability (as noted by the authors). This makes it impractical for our setting, where we aim to decompose arbitrary images efficiently. 
We therefore construct an alternative, feed-forward baseline to evaluate our approach. 
Given an input image, we prompt Qwen3-VL-8B-Instruct to describe two possible inspiration sources that could have been combined to form it. Then, we generate images from these descriptions using Flux.1 Kontext in two settings: from text alone (T2I), and with the input image as conditioning (I2I). We illustrate the results of these two baselines and InspirationTree in~\Cref{fig:decomposition_results_comparison}. Our method significancy improves over the baselines while producing meaningful decomposition results similar to those of InspirationTree from just a single image and in a feed-forward manner.

Next, we evaluate our method compared to the proposed baselines on 915 images from our synthetic image pool (described in \Cref{subsec:pool}). InspirationTree is omitted from this evaluation due to its significant runtime. Intuitively, A good decomposition should produce two components that are both related to the input but distinct from each other. Thus, we compute DreamSim \cite{fu2023dreamsim} similarity between each component and the input, and report their harmonic mean, which penalizes decompositions where one component matches but the other does not. The results in \Cref{tab:decomposition_dreamsim_comparison} show that our method achieves the highest harmonic mean, with high similarity of both generated components to the input while achieving low average similarity between the two components themselves. The baselines tend to produce one similar component and one unrelated.

\begin{table}
\centering
\small
\caption{Decomposition quality measured by DreamSim similarity. We report similarity between each component and the input (Comp1/Comp2 $\leftrightarrow$ Orig), similarity between components (Comp1 $\leftrightarrow$ Comp2), and the harmonic mean of component-to-input similarities. A good decomposition should have high harmonic mean (both components related to input) and low component similarity (components are distinct).}
\vspace{-0.25cm}
\label{tab:decomposition_dreamsim_comparison}
\begin{tabular}{lcccc}
\hline
Method & Comp1 & Comp2 & Comp1 & Harmonic  \\
& $\leftrightarrow$ Input & $\leftrightarrow$ Input & $\leftrightarrow$ Comp2 & Mean $\uparrow$ \\
\hline
Base T2I & $0.51 \pm 0.18$ & $0.25 \pm 0.11$ & $0.24 \pm 0.10$ & $0.31 \pm 0.10$ \\
Base I2I & $0.69 \pm 0.24$ & $0.32 \pm 0.17$ & $0.28 \pm 0.15$ & $0.40 \pm 0.16$ \\
Ours & $\mathbf{0.55 \pm 0.14}$ & $\mathbf{0.56 \pm 0.14}$ & $\mathbf{0.31 \pm 0.14}$ & $\mathbf{0.53 \pm 0.10}$ \\
\hline
\end{tabular}
\vspace{-0.4cm}
\end{table}

\section{Limitations}
While our method enables visual exploration through non-trivial image combinations, it has several limitations. First, certain failure modes in the combination can occur: one input may dominate the composition, the spatial structure of one input may persist across seeds, or the combination may remain trivial (e.g., simply inserting one object into the other). 
Second, our method currently supports only two input images; extending to multiple inputs could enable richer combinations that better reflect how designers draw from many references simultaneously. Third, users have limited control over the combination process; enabling continuous control over how much of each input to incorporate would improve interaction. Finally, generation currently takes approximately 30 seconds per image; reducing this to near real-time would significantly benefit the interactive experience.

\section{Discussion and Conclusions}
We presented \emph{Inspiration Seeds}, a method for generating non-trivial visual combinations from pairs of images. Unlike existing approaches that execute well-specified ideas, our method is designed to support the earlier, exploratory phase of visual work — surfacing unexpected connections between visual concepts without requiring users to articulate what they are looking for.
Central to our approach is a decomposition technique using CLIP SAEs that enables automatic training data generation without predefined relationship categories, allowing our model to learn open-ended visual combinations beyond fixed transformations such as style transfer or object insertion.
Finally, we introduced a new evaluation framework based on description complexity, grounded in research linking description length to cognitive complexity. Our experiments show that our method produces combinations that require richer descriptions than those generated by other methods, indicating deeper integration of visual aspects.
We hope this work opens new directions for generative models that support exploratory settings in visual domains, enhancing visual ideation while keeping the human creator at the center.

\begin{acks}
We thank Yuval Alaluf for providing feedback on early versions of our manuscript. 
This work was partially supported by Hyundai Motor Co/MIT Agreement dated 2/22/2023, Hasso Plattner Foundation/MIT Agreement dated 11/02/2022, and IBM/MIT Agreement No. W1771646. The sponsors had no role in the experimental design or analysis, the decision to publish, or manuscript preparation. The authors have no competing interests to report.

\end{acks}

\bibliographystyle{ACM-Reference-Format}
\bibliography{main}

@misc{irisvanherpen2020,
  author = {{Iris van Herpen}},
  title = {Sensory Seas Collection},
  year = {2020},
  howpublished = {\url{https://www.irisvanherpen.com/collections/sensory-seas}},
  note = {Haute Couture Spring/Summer 2020}
}

@article{goldschmidt1991,
  author = {Goldschmidt, Gabriela},
  title = {The Dialectics of Sketching},
  journal = {Creativity Research Journal},
  volume = {4},
  number = {2},
  pages = {123--143},
  year = {1991}
}

@article{epstein2023art,
  author = {Epstein, Ziv and Hertzmann, Aaron},
  title = {Art and the Science of Generative AI},
  journal = {Science},
  volume = {380},
  number = {6650},
  pages = {1110--1111},
  year = {2023}
}

@article{mazzone2019art,
  author = {Mazzone, Marian and Elgammal, Ahmed},
  title = {Art, Creativity, and the Potential of Artificial Intelligence},
  journal = {Arts},
  volume = {8},
  number = {1},
  pages = {26},
  year = {2019}
}

@inproceedings{rombach2022sd,
  author = {Rombach, Robin and Blattmann, Andreas and Lorenz, Dominik and Esser, Patrick and Ommer, Bj{\"o}rn},
  title = {High-Resolution Image Synthesis with Latent Diffusion Models},
  booktitle = {CVPR},
  year = {2022}
}

@article{jonson2005design,
  author = {Jonson, Ben},
  title = {Design Ideation: The Conceptual Sketch in the Digital Age},
  journal = {Design Studies},
  volume = {26},
  number = {6},
  pages = {613--624},
  year = {2005}
}

@article{kim2002fuzzy,
  author = {Kim, Jongbae and Wilemon, David},
  title = {Focusing the Fuzzy Front-End in New Product Development},
  journal = {R\&D Management},
  volume = {32},
  number = {4},
  pages = {269--279},
  year = {2002}
}

@article{suwa1997unexpected,
  author = {Suwa, Masaki and Tversky, Barbara},
  title = {What Do Architects and Students Perceive in Their Design Sketches? A Protocol Analysis},
  journal = {Design Studies},
  volume = {18},
  number = {4},
  pages = {385--403},
  year = {1997}
}

@misc{nanobanana2025,
  author = {{Google DeepMind}},
  title = {Nano Banana (Gemini 2.5 Flash Image)},
  year = {2025},
  howpublished = {\url{https://deepmind.google/models/gemini-image/flash/}},
  note = {Accessed: 2025}
}

@misc{blackforestlabs2024flux,
  title        = {FLUX.1: A Family of Open-Weight Text-to-Image Models},
  author       = {{Black Forest Labs}},
  year         = {2024},
  howpublished = {\url{https://blackforestlabs.ai}},
  note         = {Accessed 2024}
}

@misc{SAEBlog,
title={Interpreting and Steering Features in Images},
author = {Daujotas, G.},
year = {2024},
howpublished = {\url{https://www.lesswrong.com/posts/Quqekpvx8BGMMcaem/interpreting-and-steering-features-in-images}
}
}

@inproceedings{esser2024scaling,
  title={Scaling rectified flow transformers for high-resolution image synthesis},
  author={Esser, Patrick and Kulal, Sumith and Blattmann, Andreas and Entezari, Rahim and M{\"u}ller, Jonas and Saini, Harry and Levi, Yam and Lorenz, Dominik and Sauer, Axel and Boesel, Frederic and others},
  booktitle={Forty-first international conference on machine learning},
  year={2024}
}

@article{wu2025qwen,
  title={Qwen-image technical report},
  author={Wu, Chenfei and Li, Jiahao and Zhou, Jingren and Lin, Junyang and Gao, Kaiyuan and Yan, Kun and Yin, Sheng-ming and Bai, Shuai and Xu, Xiao and Chen, Yilei and others},
  journal={arXiv preprint arXiv:2508.02324},
  year={2025}
}

@misc{cunningham2023sparseautoencodershighlyinterpretable,
      title={Sparse Autoencoders Find Highly Interpretable Features in Language Models}, 
      author={Hoagy Cunningham and Aidan Ewart and Logan Riggs and Robert Huben and Lee Sharkey},
      year={2023},
      eprint={2309.08600},
      archivePrefix={arXiv},
      primaryClass={cs.LG},
      url={https://arxiv.org/abs/2309.08600}, 
}

@misc{labs2025flux1kontextflowmatching,
      title={FLUX.1 Kontext: Flow Matching for In-Context Image Generation and Editing in Latent Space},
      author={Black Forest Labs and Stephen Batifol and Andreas Blattmann and Frederic Boesel and Saksham Consul and Cyril Diagne and Tim Dockhorn and Jack English and Zion English and Patrick Esser and Sumith Kulal and Kyle Lacey and Yam Levi and Cheng Li and Dominik Lorenz and Jonas Müller and Dustin Podell and Robin Rombach and Harry Saini and Axel Sauer and Luke Smith},
      year={2025},
      eprint={2506.15742},
      archivePrefix={arXiv},
      primaryClass={cs.GR},
      url={https://arxiv.org/abs/2506.15742},
}

@article{inspirationtree23,
author = {Vinker, Yael and Voynov, Andrey and Cohen-Or, Daniel and Shamir, Ariel},
title = {Concept Decomposition for Visual Exploration and Inspiration},
year = {2023},
issue_date = {December 2023},
publisher = {Association for Computing Machinery},
address = {New York, NY, USA},
volume = {42},
number = {6},
issn = {0730-0301},
url = {https://doi.org/10.1145/3618315},
doi = {10.1145/3618315},
journal = {ACM Trans. Graph.},
month = dec,
articleno = {241},
numpages = {13}
}

@inproceedings{kumari2023customdiffusion,
  author = {Kumari, Nupur and Zhang, Bingliang and Zhang, Richard and Shechtman, Eli and Zhu, Jun-Yan},
  title = {Multi-Concept Customization of Text-to-Image Diffusion},
  booktitle = {CVPR},
  year = {2023}
}

@inproceedings{frenkel2024blora,
  author = {Frenkel, Yarden and Vinker, Yael and Shamir, Ariel and Cohen-Or, Daniel},
  title = {Implicit Style-Content Separation using B-LoRA},
  booktitle = {ECCV},
  year = {2024}
}

@inproceedings{gal2022textualinversion,
  author = {Gal, Rinon and Alaluf, Yuval and Atzmon, Yuval and Patashnik, Or and Bermano, Amit Haim and Chechik, Gal and Cohen-Or, Daniel},
  title = {An Image is Worth One Word: Personalizing Text-to-Image Generation using Textual Inversion},
  booktitle = {ICLR},
  year = {2023}
}

@inproceedings{Razzhigaev2023KandinskyAI,
  title={Kandinsky: an Improved Text-to-Image Synthesis with Image Prior and Latent Diffusion},
  author={Anton Razzhigaev and Arseniy Shakhmatov and Anastasia Maltseva and V.Ya. Arkhipkin and Igor Pavlov and Ilya Ryabov and Angelina Kuts and Alexander Panchenko and Andrey Kuznetsov and Denis Dimitrov},
  booktitle={Conference on Empirical Methods in Natural Language Processing},
  year={2023},
  url={https://api.semanticscholar.org/CorpusID:263671912}
}

@inproceedings{avrahami2023bas,
  author = {Avrahami, Omri and Aberman, Kfir and Fried, Ohad and Cohen-Or, Daniel and Lischinski, Dani},
  title = {Break-A-Scene: Extracting Multiple Concepts from a Single Image},
  year = {2023},
  isbn = {9798400703157},
  publisher = {Association for Computing Machinery},
  address = {New York, NY, USA},
  url = {https://doi.org/10.1145/3610548.3618154},
  doi = {10.1145/3610548.3618154},        
  booktitle = {SIGGRAPH Asia 2023 Conference Papers},
  articleno = {96},
  numpages = {12},
  location = {, Sydney, NSW, Australia, },
  series = {SA '23}
}

@article{Ruiz2022DreamBoothFT,
  title={DreamBooth: Fine Tuning Text-to-Image Diffusion Models for Subject-Driven Generation},
  author={Nataniel Ruiz and Yuanzhen Li and Varun Jampani and Yael Pritch and Michael Rubinstein and Kfir Aberman},
  journal={2023 IEEE/CVF Conference on Computer Vision and Pattern Recognition (CVPR)},
  year={2022},
  pages={22500-22510},
  url={https://api.semanticscholar.org/CorpusID:251800180}
}

@inproceedings{Radford2021LearningTV,
  title={Learning Transferable Visual Models From Natural Language Supervision},
  author={Alec Radford and Jong Wook Kim and Chris Hallacy and Aditya Ramesh and Gabriel Goh and Sandhini Agarwal and Girish Sastry and Amanda Askell and Pamela Mishkin and Jack Clark and Gretchen Krueger and Ilya Sutskever},
  booktitle={International Conference on Machine Learning},
  year={2021},
  url={https://api.semanticscholar.org/CorpusID:231591445}
}

@article{sun2021speaking,
  title={Seeing and speaking: How verbal "description length" encodes visual complexity},
  author={Sun, Zekun and Firestone, Chaz},
  journal={Journal of Experimental Psychology: General},
  volume={151},
  number={1},
  pages={82--96},
  year={2022},
  publisher={APA}
}

@article{kolmogorov1965,
  title={Three approaches to the quantitative definition of information},
  author={Kolmogorov, Andrei N.},
  journal={Problems of Information Transmission},
  volume={1},
  number={1},
  pages={1--7},
  year={1965}
}

@book{li1997vitanyi,
  title={An Introduction to Kolmogorov Complexity and Its Applications},
  author={Li, Ming and Vit{\'a}nyi, Paul},
  year={1997},
  publisher={Springer},
  address={New York}
}

@article{richardson2025piece,
  title={Piece it Together: Part-Based Concepting with IP-Priors},
  author={Richardson, Elad and Goldberg, Kfir and Alaluf, Yuval and Cohen-Or, Daniel},
  journal={arXiv preprint arXiv:2503.10365},
  year={2025}
}

@inproceedings{richardson2025pops,
  title={pOps: Photo-inspired diffusion operators},
  author={Richardson, Elad and Alaluf, Yuval and Mahdavi-Amiri, Ali and Cohen-Or, Daniel},
  booktitle={Proceedings of the Special Interest Group on Computer Graphics and Interactive Techniques Conference Conference Papers},
  pages={1--12},
  year={2025}
}

@inproceedings{fu2023dreamsim,
  title={DreamSim: Learning New Dimensions of Human Visual Similarity using Synthetic Data},
  author={Fu, Stephanie and Tamir, Netanel and Sundaram, Shobhita and Chai, Lucy and Zhang, Richard and Dekel, Tali and Isola, Phillip},
  booktitle={Advances in Neural Information Processing Systems},
  year={2023}
}

@article{Bonnardel2005TowardsSE,
  title={Towards supporting evocation processes in creative design: A cognitive approach},
  author={Nathalie Bonnardel and Evelyne Cauzinille-Marm{\`e}che},
  journal={Int. J. Hum. Comput. Stud.},
  year={2005},
  volume={63},
  pages={422-435}
}

@article{WILKENFELD200121,
    title = {Similarity and Emergence in Conceptual Combination},
    journal = {Journal of Memory and Language},
    volume = {45},
    number = {1},
    pages = {21-38},
    year = {2001},
    issn = {0749-596X},
    doi = {https://doi.org/10.1006/jmla.2000.2772},
    url = {https://www.sciencedirect.com/science/article/pii/S0749596X00927724},
    author = {Merryl J. Wilkenfeld and Thomas B. Ward},
}

@article{Runco2012TheSD,
  title={The Standard Definition of Creativity},
  author={Mark A. Runco and Garrett J. Jaeger},
  journal={Creativity Research Journal},
  year={2012},
  volume={24},
  pages={92 - 96}
}

@article{ECKERT2000523,
    title = {Sources of inspiration: a language of design},
    journal = {Design Studies},
    volume = {21},
    number = {5},
    pages = {523-538},
    year = {2000},
    issn = {0142-694X},
    doi = {https://doi.org/10.1016/S0142-694X(00)00022-3},
    url = {https://www.sciencedirect.com/science/article/pii/S0142694X00000223},
    author = {Claudia Eckert and Martin Stacey}
}

@article{ImageSense2020,
    author = {Koch, Janin and Taffin, Nicolas and Beaudouin-Lafon, Michel and Laine, Markku and Lucero, Andr\'{e}s and Mackay, Wendy E.},
    title = {ImageSense: An Intelligent Collaborative Ideation Tool to Support Diverse Human-Computer Partnerships},
    year = {2020},
    issue_date = {May 2020},
    publisher = {Association for Computing Machinery},
    address = {New York, NY, USA},
    volume = {4},
    number = {CSCW1},
    url = {https://doi.org/10.1145/3392850},
    doi = {10.1145/3392850},
    journal = {Proc. ACM Hum.-Comput. Interact.},
    month = {may},
    articleno = {45},
    numpages = {27}
}

@inproceedings{MoodCubes2022,
    author = {Ivanov, Alexander and Ledo, David and Grossman, Tovi and Fitzmaurice, George and Anderson, Fraser},
    title = {MoodCubes: Immersive Spaces for Collecting, Discovering and Envisioning Inspiration Materials},
    year = {2022},
    isbn = {9781450393584},
    publisher = {Association for Computing Machinery},
    address = {New York, NY, USA},
    url = {https://doi.org/10.1145/3532106.3533565},
    doi = {10.1145/3532106.3533565},
    booktitle = {Designing Interactive Systems Conference},
    pages = {189–203},
    numpages = {15},
    location = {Virtual Event, Australia},
    series = {DIS '22}
}

@inproceedings{MetaMap2021,
    author = {Kang, Youwen and Sun, Zhida and Wang, Sitong and Huang, Zeyu and Wu, Ziming and Ma, Xiaojuan},
    title = {MetaMap: Supporting Visual Metaphor Ideation through Multi-Dimensional Example-Based Exploration},
    year = {2021},
    isbn = {9781450380966},
    publisher = {Association for Computing Machinery},
    address = {New York, NY, USA},
    url = {https://doi.org/10.1145/3411764.3445325},
    doi = {10.1145/3411764.3445325},
    booktitle = {Proceedings of the 2021 CHI Conference on Human Factors in Computing Systems},
    articleno = {427},
    numpages = {15},
    location = {Yokohama, Japan},
    series = {CHI '21}
}

@article{Koch2019MayAD,
  title={May AI?: Design Ideation with Cooperative Contextual Bandits},
  author={Janin Koch and Andr{\'e}s Lucero and Lena Hegemann and Antti Oulasvirta},
  journal={Proceedings of the 2019 CHI Conference on Human Factors in Computing Systems},
  year={2019}
}

@inproceedings{Nichol2021GLIDETP,
  title={GLIDE: Towards Photorealistic Image Generation and Editing with Text-Guided Diffusion Models},
  author={Alex Nichol and Prafulla Dhariwal and Aditya Ramesh and Pranav Shyam and Pamela Mishkin and Bob McGrew and Ilya Sutskever and Mark Chen},
  booktitle={International Conference on Machine Learning},
  year={2021},
  url={https://api.semanticscholar.org/CorpusID:245335086}
}

@article{saharia2022photorealistic,
  title={Photorealistic text-to-image diffusion models with deep language understanding},
  author={Saharia, Chitwan and Chan, William and Saxena, Saurabh and Li, Lala and Whang, Jay and Denton, Emily L and Ghasemipour, Kamyar and Gontijo Lopes, Raphael and Karagol Ayan, Burcu and Salimans, Tim and others},
  journal={Advances in Neural Information Processing Systems},
  volume={35},
  pages={36479--36494},
  year={2022}
}

@article{ramesh2022hierarchical,
  title={Hierarchical text-conditional image generation with clip latents},
  author={Ramesh, Aditya and Dhariwal, Prafulla and Nichol, Alex and Chu, Casey and Chen, Mark},
  journal={arXiv preprint arXiv:2204.06125},
  year={2022}
}

@inproceedings{dorfman2025ip,
  title={Ip-composer: Semantic composition of visual concepts},
  author={Dorfman, Sara and Cohen-Bar, Dana and Gal, Rinon and Cohen-Or, Daniel},
  booktitle={Proceedings of the Special Interest Group on Computer Graphics and Interactive Techniques Conference Conference Papers},
  pages={1--11},
  year={2025}
}

@inproceedings{hertzmann2018can,
  title={Can computers create art?},
  author={Hertzmann, Aaron},
  booktitle={Arts},
  volume={7},
  number={2},
  pages={18},
  year={2018},
  organization={MDPI}
}

@inproceedings{elhoseiny2019creativity,
  title={Creativity inspired zero-shot learning},
  author={Elhoseiny, Mohamed and Elfeki, Mohamed},
  booktitle={Proceedings of the IEEE/CVF international conference on computer vision},
  pages={5784--5793},
  year={2019}
}

@inproceedings{Oppenlaender_2022, 
   series={Academic Mindtrek 2022},
   title={The Creativity of Text-to-Image Generation},
   url={http://dx.doi.org/10.1145/3569219.3569352},
   DOI={10.1145/3569219.3569352},
   booktitle={Proceedings of the 25th International Academic Mindtrek Conference},
   publisher={ACM},
   author={Oppenlaender, Jonas},
   year={2022},
   month=nov, collection={Academic Mindtrek 2022} }

@article{richardson2024conceptlab,
  title={ConceptLab: Creative Concept Generation using VLM-Guided Diffusion Prior Constraints},
  author={Richardson, Elad and Goldberg, Kfir and Alaluf, Yuval and Cohen-Or, Daniel},
  journal={ACM Transactions on Graphics},
  volume={43},
  number={3},
  pages={1--14},
  year={2024},
  publisher={ACM New York, NY}
}

@inproceedings{
lee2024languageinformed,
title={Language-Informed Visual Concept Learning},
author={Sharon Lee and Yunzhi Zhang and Shangzhe Wu and Jiajun Wu},
booktitle={The Twelfth International Conference on Learning Representations},
year={2024},
url={https://openreview.net/forum?id=juuyW8B8ig}
}

@article{ye2023ip,
  title={Ip-adapter: Text compatible image prompt adapter for text-to-image diffusion models},
  author={Ye, Hu and Zhang, Jun and Liu, Sibo and Han, Xiao and Yang, Wei},
  journal={arXiv preprint arXiv:2308.06721},
  year={2023}
}

@article{zaigrajew2025msae,
  title={Interpreting CLIP with Hierarchical Sparse Autoencoders},
  author={Zaigrajew, Vladimir and Baniecki, Hubert and Biecek, Przemyslaw},
  journal={arXiv preprint arXiv:2502.20578},
  year={2025}
}

@misc{fry2024towards,
  author       = {Fry, Hugo},
  title        = {Towards Multimodal Interpretability: Learning Sparse Interpretable Features in Vision Transformers},
  year         = {2024},
  howpublished = {\url{https://www.lesswrong.com/posts/bCtbuWraqYTDtuARg/towards-multimodal-interpretability-learning-sparse}},
  note         = {LessWrong blog post},
}

@inproceedings{peebles2023scalable,
  title={Scalable diffusion models with transformers},
  author={Peebles, William and Xie, Saining},
  booktitle={Proceedings of the IEEE/CVF international conference on computer vision},
  pages={4195--4205},
  year={2023}
}

@article{hu2022lora,
  title={Lora: Low-rank adaptation of large language models.},
  author={Hu, Edward J and Shen, Yelong and Wallis, Phillip and Allen-Zhu, Zeyuan and Li, Yuanzhi and Wang, Shean and Wang, Lu and Chen, Weizhu and others},
  journal={ICLR},
  volume={1},
  number={2},
  pages={3},
  year={2022}
}

@misc{ostrisAIToolkit,
  author       = {{Ostris AI-Toolkit Contributors}},
  title        = {{Ostris AI-Toolkit}},
  year         = {2025},
  howpublished = {\url{https://github.com/ostris/ai-toolkit}},
  note         = {GitHub repository},
}

@misc{gutflaish2025generatingimage1000words,
      title={Generating an Image From 1,000 Words: Enhancing Text-to-Image With Structured Captions}, 
      author={Eyal Gutflaish and Eliran Kachlon and Hezi Zisman and Tal Hacham and Nimrod Sarid and Alexander Visheratin and Saar Huberman and Gal Davidi and Guy Bukchin and Kfir Goldberg and Ron Mokady},
      year={2025},
      eprint={2511.06876},
      archivePrefix={arXiv},
      primaryClass={cs.CV},
      url={https://arxiv.org/abs/2511.06876}, 
}

@misc{seedream2025seedream40nextgenerationmultimodal,
      title={Seedream 4.0: Toward Next-generation Multimodal Image Generation}, 
      author={Team Seedream and : and Yunpeng Chen and Yu Gao and Lixue Gong and Meng Guo and Qiushan Guo and Zhiyao Guo and Xiaoxia Hou and Weilin Huang and Yixuan Huang and Xiaowen Jian and Huafeng Kuang and Zhichao Lai and Fanshi Li and Liang Li and Xiaochen Lian and Chao Liao and Liyang Liu and Wei Liu and Yanzuo Lu and Zhengxiong Luo and Tongtong Ou and Guang Shi and Yichun Shi and Shiqi Sun and Yu Tian and Zhi Tian and Peng Wang and Rui Wang and Xun Wang and Ye Wang and Guofeng Wu and Jie Wu and Wenxu Wu and Yonghui Wu and Xin Xia and Xuefeng Xiao and Shuang Xu and Xin Yan and Ceyuan Yang and Jianchao Yang and Zhonghua Zhai and Chenlin Zhang and Heng Zhang and Qi Zhang and Xinyu Zhang and Yuwei Zhang and Shijia Zhao and Wenliang Zhao and Wenjia Zhu},
      year={2025},
      eprint={2509.20427},
      archivePrefix={arXiv},
      primaryClass={cs.CV},
      url={https://arxiv.org/abs/2509.20427}, 
}

@misc{gemini25_2025,
      title={Gemini 2.5: Pushing the Frontier with Advanced Reasoning, Multimodality, Long Context, and Next Generation Agentic Capabilities}, 
      author={{Gemini Team, Google}},
      year={2025},
      eprint={2507.06261},
      archivePrefix={arXiv},
      primaryClass={cs.CL},
      url={https://arxiv.org/abs/2507.06261},
}

@article{gentner1983structure,
title = {Structure-mapping: A theoretical framework for analogy},
journal = {Cognitive Science},
volume = {7},
number = {2},
pages = {155-170},
year = {1983},
issn = {0364-0213},
doi = {https://doi.org/10.1016/S0364-0213(83)80009-3},
url = {https://www.sciencedirect.com/science/article/pii/S0364021383800093},
author = {Dedre Gentner},
}

@article{tversky2011visualizing,
  title   = {Visualizing Thought},
  author  = {Tversky, Barbara},
  journal = {Topics in Cognitive Science},
  volume  = {3},
  number  = {3},
  pages   = {499--535},
  year    = {2011}
}

@book{arnheim1969visualthinking,
  title     = {Visual Thinking},
  author    = {Arnheim, Rudolf},
  year      = {1969},
  publisher = {University of California Press}
}

@article{TokenVerse25,
author = {Garibi, Daniel and Yadin, Shahar and Paiss, Roni and Tov, Omer and Zada, Shiran and Ephrat, Ariel and Michaeli, Tomer and Mosseri, Inbar and Dekel, Tali},
title = {TokenVerse: Versatile Multi-concept Personalization in Token Modulation Space},
year = {2025},
issue_date = {August 2025},
publisher = {Association for Computing Machinery},
address = {New York, NY, USA},
volume = {44},
number = {4},
issn = {0730-0301},
url = {https://doi.org/10.1145/3730843},
doi = {10.1145/3730843},
journal = {ACM Trans. Graph.},
month = jul,
articleno = {41},
numpages = {11}
}

@inproceedings{Cross-ImageAlaluf24,
author = {Alaluf, Yuval and Garibi, Daniel and Patashnik, Or and Averbuch-Elor, Hadar and Cohen-Or, Daniel},
title = {Cross-Image Attention for Zero-Shot Appearance Transfer},
year = {2024},
isbn = {9798400705250},
publisher = {Association for Computing Machinery},
address = {New York, NY, USA},
url = {https://doi.org/10.1145/3641519.3657423},
doi = {10.1145/3641519.3657423},
booktitle = {ACM SIGGRAPH 2024 Conference Papers},
articleno = {132},
numpages = {12},
location = {Denver, CO, USA},
series = {SIGGRAPH '24}
}

@article{Gatys2015ANA,
  title={A Neural Algorithm of Artistic Style},
  author={Leon A. Gatys and Alexander S. Ecker and Matthias Bethge},
  journal={ArXiv},
  year={2015},
  volume={abs/1508.06576},
  url={https://api.semanticscholar.org/CorpusID:13914930}
}

@article{Xu2024CusConceptCV,
  title={CusConcept: Customized Visual Concept Decomposition with Diffusion Models},
  author={Zhi Xu and Shaozhe Hao and Kai Han},
  journal={2025 IEEE/CVF Winter Conference on Applications of Computer Vision (WACV)},
  year={2024},
  pages={3678-3687},
  url={https://api.semanticscholar.org/CorpusID:273023065}
}

@inproceedings{ZipLoRA24,
author = {Shah, Viraj and Ruiz, Nataniel and Cole, Forrester and Lu, Erika and Lazebnik, Svetlana and Li, Yuanzhen and Jampani, Varun},
title = {ZipLoRA: Any Subject in Any Style by Effectively Merging LoRAs},
year = {2024},
isbn = {978-3-031-73231-7},
publisher = {Springer-Verlag},
address = {Berlin, Heidelberg},
url = {https://doi.org/10.1007/978-3-031-73232-4_24},
doi = {10.1007/978-3-031-73232-4_24},
booktitle = {Computer Vision -- ECCV 2024: 18th European Conference, Milan, Italy, September 29--October 4, 2024, Proceedings, Part I},
pages = {422--438},
numpages = {17},
location = {Milan, Italy}
}

@inproceedings{ngweta2023simple,
  title={Simple Disentanglement of Style and Content in Visual Representations},
  author={Ngweta, Lilian and others},
  booktitle={ICML},
  year={2023}
}

@misc{reve2024,
  author       = {{Reve AI}},
  title        = {Reve: Image Generation Platform},
  howpublished = {\url{https://www.reve.ai}},
  year         = {2024},
  note         = {Commercial image generation system}
}

@inproceedings{Sims1991,
  author = {Sims, Karl},
  title = {Artificial Evolution for Computer Graphics},
  booktitle = {SIGGRAPH '91 Proceedings},
  pages = {319--328},
  year = {1991}
}

@article{Takagi2001,
  author = {Takagi, Hideyuki},
  title = {Interactive Evolutionary Computation: Fusion of the Capabilities of EC Optimization and Human Evaluation},
  journal = {Proceedings of the IEEE},
  volume = {89},
  number = {9},
  pages = {1275--1296},
  year = {2001}
}

@inproceedings{secretan2008picbreeder,
  title     = {Picbreeder: Evolving Pictures Collaboratively Online},
  author    = {Secretan, Jimmy and Beato, Nicholas and D'Ambrosio, David B. and Rodriguez, Adelein and Campbell, Adam and Stanley, Kenneth O.},
  booktitle = {Proceedings of the Genetic and Evolutionary Computation Conference (GECCO)},
  year      = {2008}
}

@misc{Mordvintsev2015,
  author = {Mordvintsev, Alexander and Olah, Christopher and Tyka, Mike},
  title = {Inceptionism: Going Deeper into Neural Networks},
  howpublished = {Google Research Blog},
  year = {2015},
  url = {https://research.google/blog/inceptionism-going-deeper-into-neural-networks/}
}

@article{white2020ganbreeder,
  title   = {GANbreeder: Evolving Images Using Deep Generative Models},
  author  = {White, Tom},
  journal = {arXiv preprint arXiv:2009.08379},
  year    = {2020}
}

\clearpage
\appendix

\suppressfloats[t] %

\noindent\begin{minipage}{\textwidth}
\centering
{\LARGE\bfseries Inspiration Seeds: Learning Non-Literal Visual Combinations for Generative Exploration \\[0.3cm]
Supplementary Material}
\end{minipage}

\par\bigskip

\addtocontents{toc}{\protect\setcounter{tocdepth}{4}}

\makeatletter
\@starttoc{toc}
\makeatother

\par\bigskip

\newtcolorbox{promptbox}[1][]{%
  colback=gray!5,
  colframe=gray!60,
  fonttitle=\bfseries\small,
  title={#1},
  breakable,
  fontupper=\footnotesize,
  left=4pt, right=4pt, top=2pt, bottom=2pt,
  before skip=6pt, after skip=6pt,
}

\section{Visual Exploration Interface}
\begin{figure}
    \centering
    \vspace{3cm}
    \includegraphics[width=0.9\linewidth]{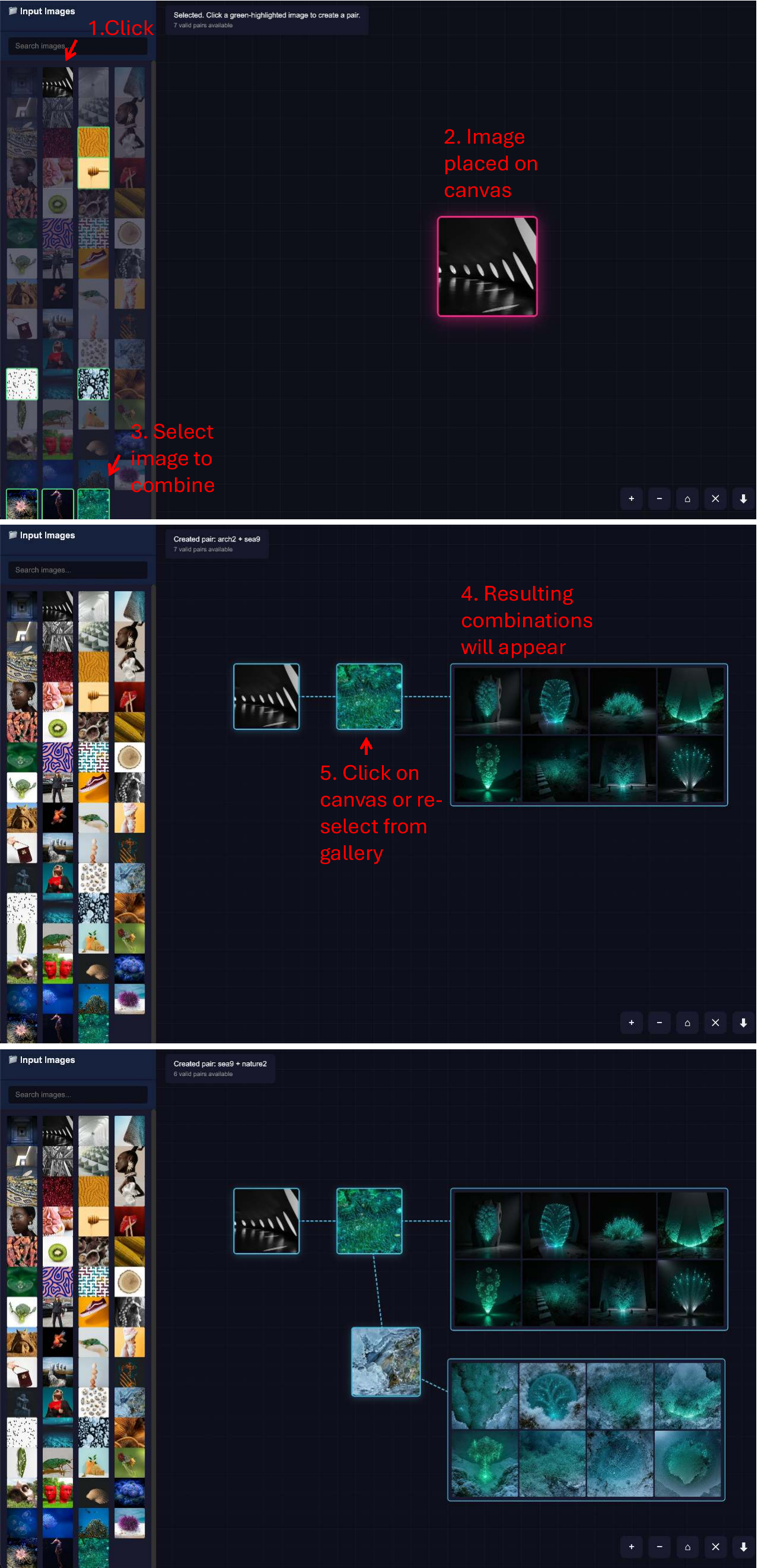}
    \caption{Interactive demo workflow. Users select an image from the gallery (1), which is placed on the canvas (2). Available pairing options are highlighted in green (3). After selecting a second image, the resulting visual combinations appear on the canvas (4). Users can click any result to continue exploring, or select a new image from the gallery (5).}
    \label{fig:demo_explained}
\end{figure}

We provide an interactive demonstration of our exploration canvas, illustrating how our method can support ideation and exploration in visual space. The demo allows users to freely combine images from a curated gallery and observe the resulting visual combinations generated by our model.
Our interactive demo is available at:
\\{\color{cyan!70!black}\href{https://kfirgoldberg.github.io/InspirationSeeds/static/viewer/index.html}{\texttt{\footnotesize kfirgoldberg.github.io/InspirationSeeds/static/viewer/index.html}}}.\\
\Cref{fig:demo_explained} illustrates the interaction workflow: (1) click an image in the gallery to select it, (2) the selected image is placed on the canvas, (3) available images that can be combined with the selection are highlighted in green—select one to create a pair, (4) the resulting combinations appear on the canvas, and (5) click on any result to use it as input for further exploration, or select a new image from the gallery to continue.
You can drag images and results to organize them in any structure you prefer.

\begin{figure*}
    \centering
    \includegraphics[width=1\linewidth]{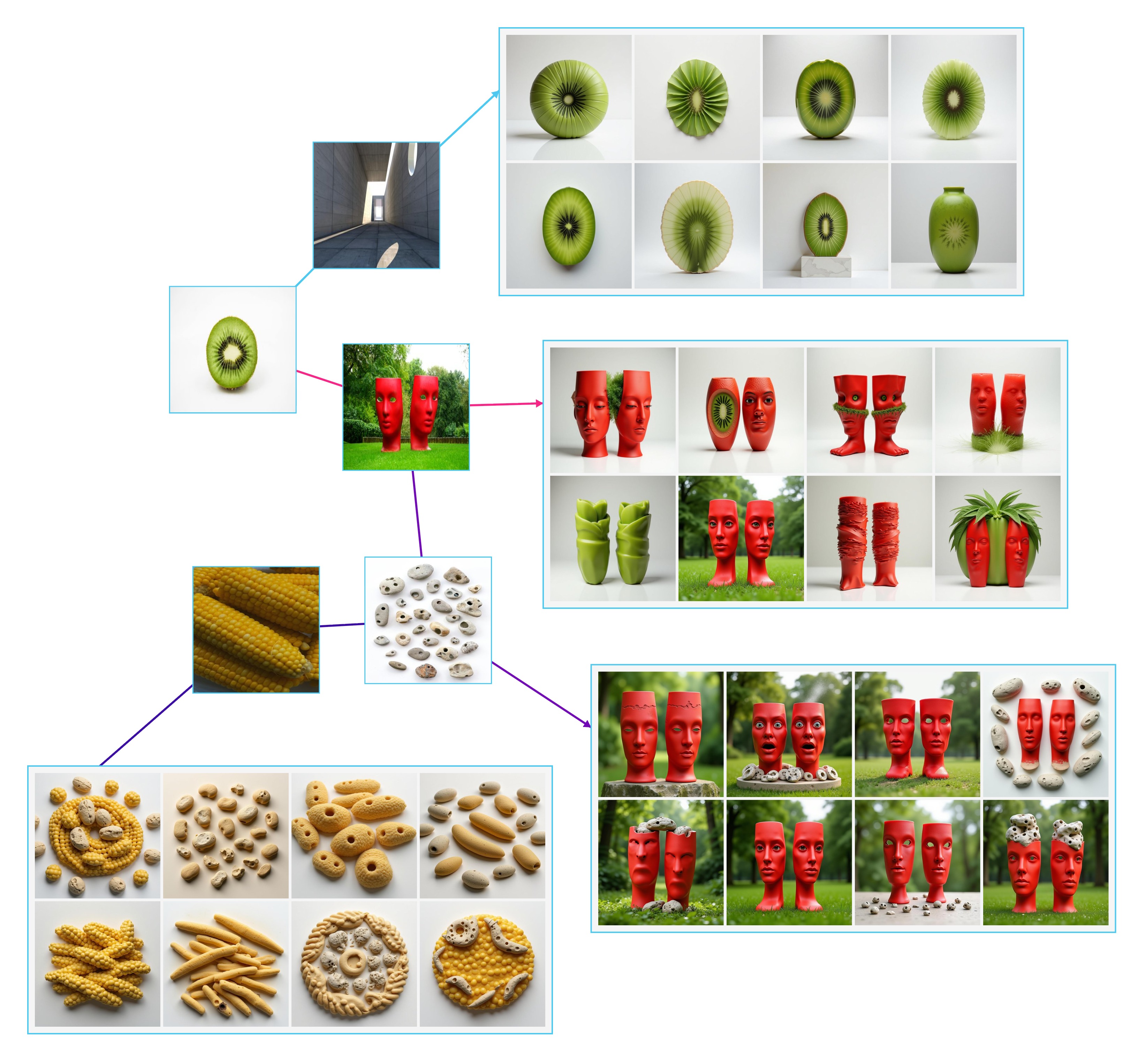}
    \caption{Exploration canvas showing visual combinations generated by our method.}
    \label{fig:canvas1}
\end{figure*}

\begin{figure*}
    \centering
    \includegraphics[width=1\linewidth]{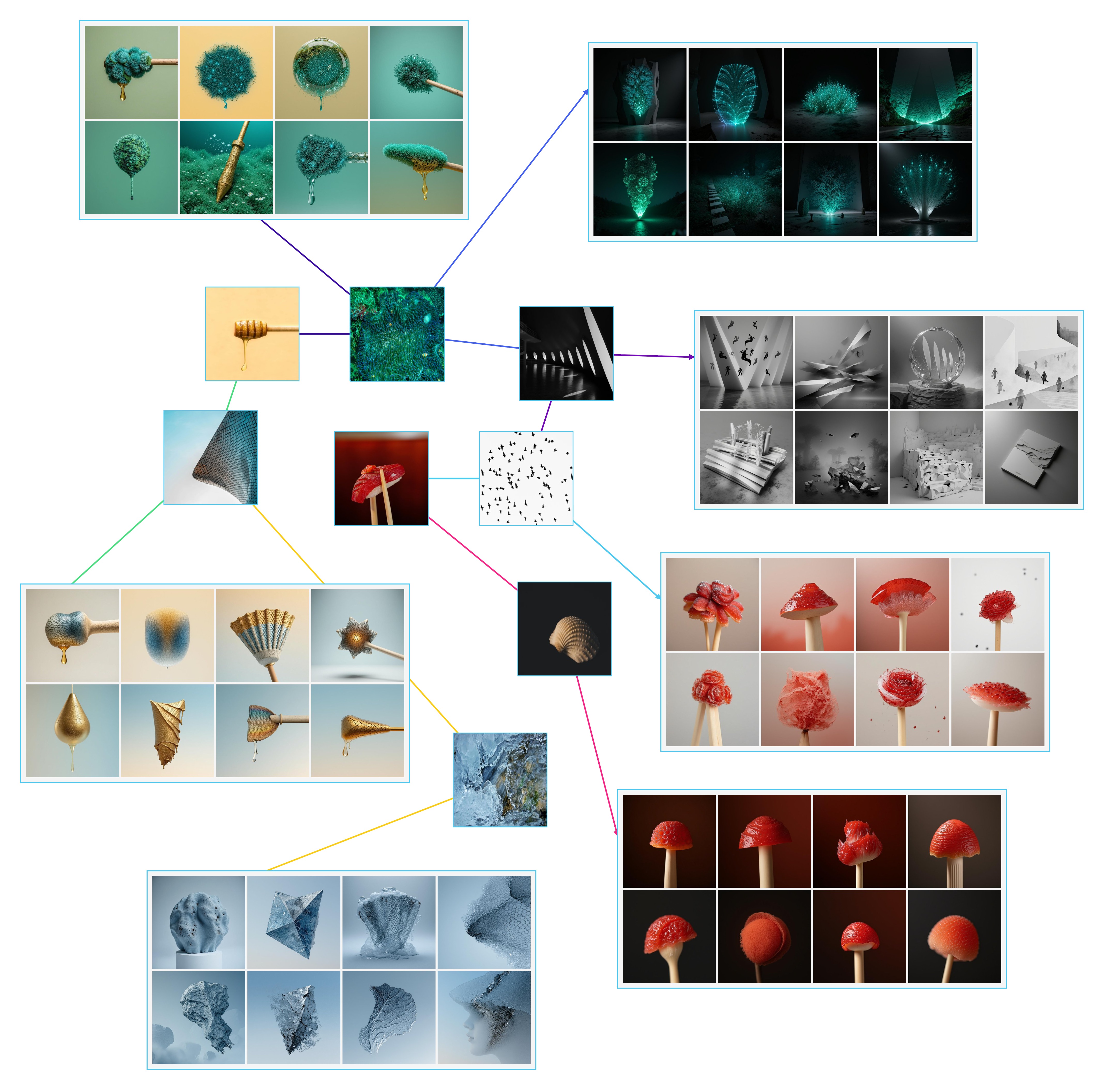}
    \caption{Exploration canvas showing visual combinations generated by our method.}
    \label{fig:canvas2}
\end{figure*}

\begin{figure*}
    \centering
    \includegraphics[width=1\linewidth]{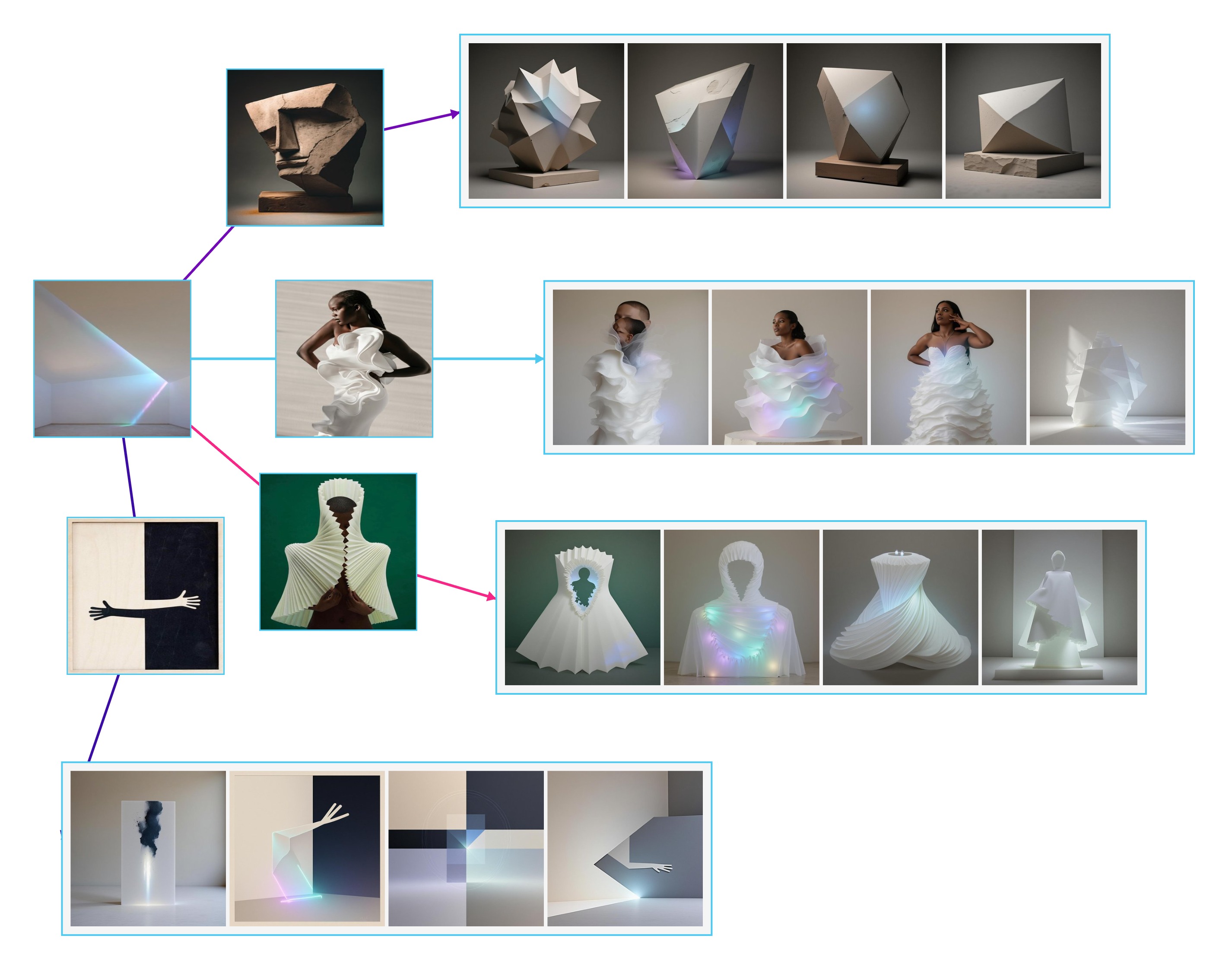}
    \caption{Exploration canvas showing visual combinations generated by our method.}
    \label{fig:canvas3}
\end{figure*}

\begin{figure*}
    \centering
    \includegraphics[width=1\linewidth]{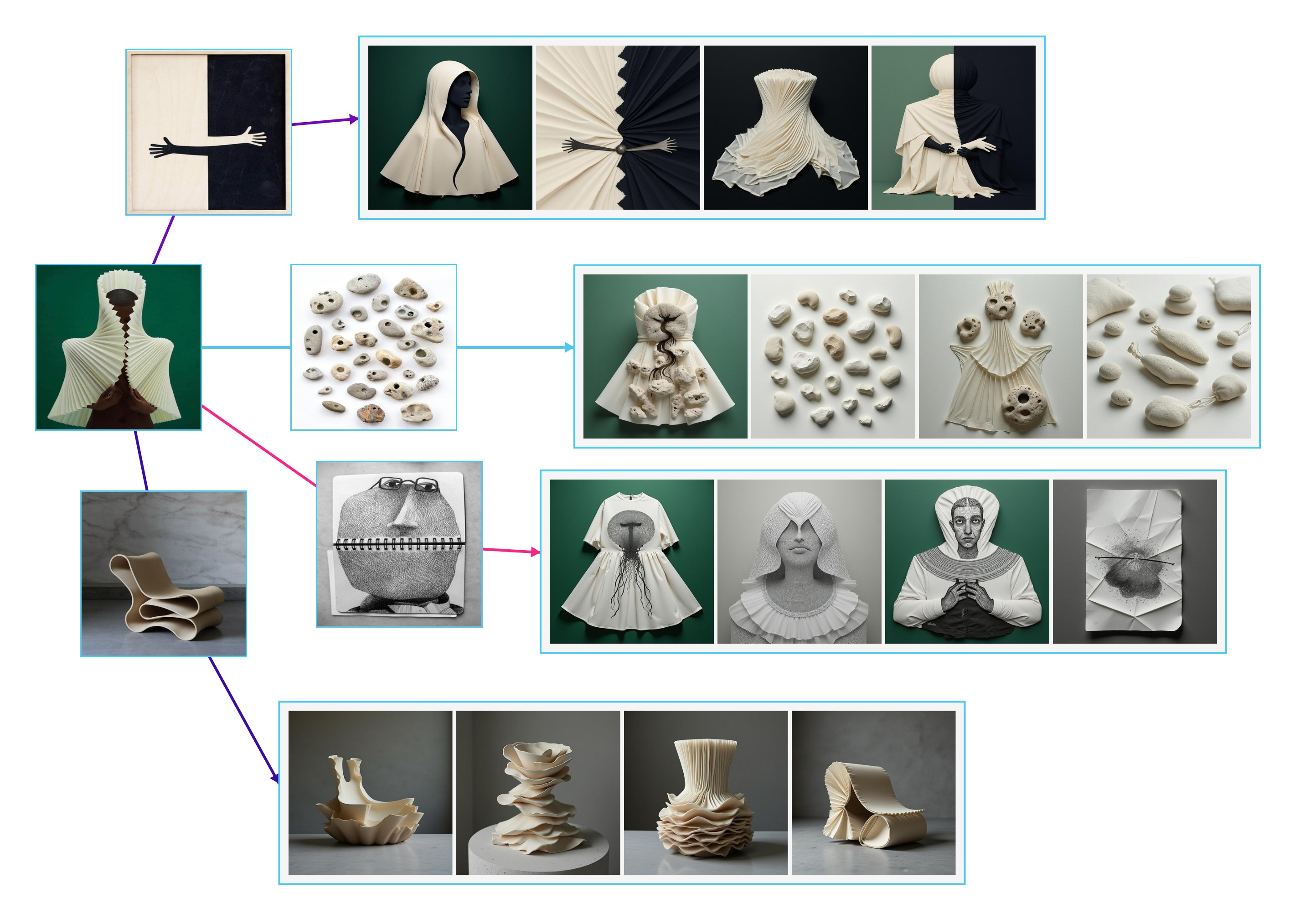}
    \caption{Exploration canvas showing visual combinations generated by our method.}
    \label{fig:canvas4}
\end{figure*}

More results of our exploration canvases are provided in \Cref{fig:canvas1,fig:canvas2,fig:canvas3,fig:canvas4}.

\section{Implementation Details}

\subsection{Training Details}
Our model builds upon FLUX.1 Kontext~\cite{labs2025flux1kontextflowmatching}. We condition generation on two input images by creating a $1024 \times 1024$ white canvas and placing each input image (resized to $512 \times 512$) in the top-left and bottom-right corners. To remove textual bias during training and inference, we use a fixed prompt: ``combine the element in the top left with the element in the bottom right to create a single object inspired by both of them''.

We fine-tune using LoRA~\cite{hu2022lora} with rank 32 for linear layers and rank 16 for convolutional layers. We use AdamW with learning rate $10^{-4}$ and batch size 1 for 15,000 steps on a single NVIDIA L40s GPU (approximately 24 hours). Training is performed using the Ostris AI-Toolkit~\cite{ostrisAIToolkit}.

Given two input images, we arrange them in the $2 \times 2$ grid and generate using the fixed prompt from training. Generation takes approximately 34 seconds per image on a single NVIDIA L40s GPU.

\subsection{Decomposition Details}
\begin{figure}[t]
    \centering
    \includegraphics[width=0.9\linewidth]{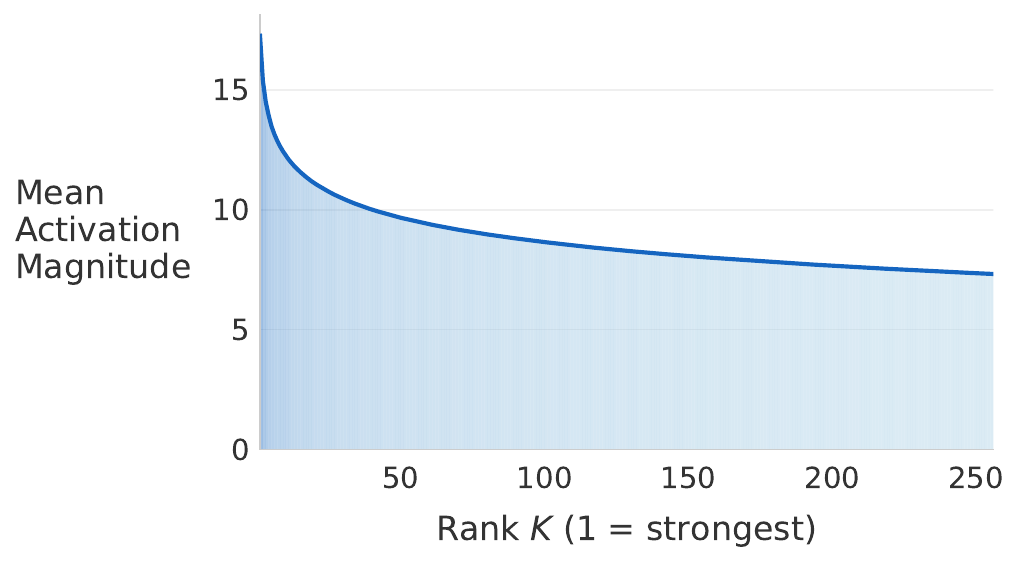}
    \caption{Histogram of SAE activation magnitudes. We use the top $32$ features to decompose images.}
    \label{fig:sae_activation_histogram}
\end{figure}

For our decomposition pipeline, we set $k=32$ for the number of features to retain in the Top-k activation selection, following the histogram of activation magnitudes shown in \Cref{fig:sae_activation_histogram}.
To improve cluster coherence, we retain the $50\%$ of vectors in each cluster closest to the centroid.

\section{Dataset Construction}
To create diverse source images for decomposition, we use several text-to-image generation models: Flux.1 Dev~\cite{blackforestlabs2024flux}, Fibo~\cite{gutflaish2025generatingimage1000words}, Reve~\cite{reve2024}, and Seedream4~\cite{seedream2025seedream40nextgenerationmultimodal}. We design two prompting strategies that serve complementary purposes, described in details next.

\subsection{Templated Prompts}
In order to create images that bundle multiple distinct visual properties and are reliably decomposable, we generate prompts using structured templates and Flux.1 Dev~\cite{blackforestlabs2024flux} as the text-to-image model.
We use two different types of templated prompts, aiming to create different levels of complexity and decomposability.
Examples of images from this set are shown in \Cref{fig:data_gen_templated}.

The first type is ontology-based prompts. We define a structured ontology over eight object categories (garments, furniture, architecture, vehicles, kitchenware, tech, food, everyday items) and six motif families (animal, material, geometry, botanical, art, texture). Each motif is paired with a canonical source description (e.g., ``a leopard, close-up fur'' for leopard spots). A prompt is assembled by sampling an object, a motif, and optional modifiers for base material (from 15 options), color (18 options), style (12 options, e.g., brutalist, art deco, mid-century modern), and environment (9 options). We sample from these sets to produce the final prompts.
These fragments are concatenated with a quality suffix (e.g., ``high detail, photorealistic.'') to form the final prompt. Below are examples of prompts obtained from this process: 
\begin{promptbox}
    "a mustard airplane, with dalmatian spots, in a mid-century modern style, on concrete floor, studio photography, crisp edges, natural shadows"
    \\
    \\
    "a teal leather courtyard, with chevron zigzags, 8k, high fidelity textures, subtle imperfections"
\end{promptbox}

The second type shifts to a product-design vocabulary. It defines 16 specific product subjects (e.g., pour-over coffee brewer, pendant light, ergonomic task chair, mechanical wristwatch), each annotated with material, palette, form, and purpose phrases. Separately, 18 inspiration entries capture natural and architectural phenomena — tidal channels, basalt ridges, bioluminescent organisms, parametric facades, shibori dye, and others. A composite prompt is built by sampling one subject and two inspirations from distinct categories, then interleaving their phrases: subject identity, form, purpose, palette, two inspiration design phrases, and optional color and material descriptors. Below are examples of the resulting prompts: 
\begin{promptbox}
    "a matte earthenware ceramic vase, with elongated shoulders tapering to a narrow mouth, crafted for contemplative floral arrangements, finished in warm sand and terra hues, shadowed by billowing overhangs that twist with atmospheric energy, with charcoal, slate, and electric blue edges, interwoven with braided flow paths that split and rejoin fluidly, with soft turquoise and mineral clay tones, set against a soft gradient backdrop with gentle, shadowless lighting, 8k detail, physically based rendering, balanced highlights"
    \\
    \\
    "a ceramic bezel mechanical wristwatch, with a domed sapphire crystal and faceted indices, crafted for collectors, with warm sepia and champagne palette, threaded with translucent layers that emit a subtle inner glow, detailed with repeating scales that tighten toward the center, with sage and pale jade accents, photographed on a seamless warm grey backdrop under neutral studio lighting, high detail, controlled reflections, refined product lighting"
\end{promptbox}

\subsection{LLM-Expanded Prompts}
To encourage more creative visual concepts that are less trivially decomposed, we use LLMs \cite{gemini25_2025} to generate long, more ``creative'' prompts and then generate images from them using several different text-to-image models~\cite{gutflaish2025generatingimage1000words,reve2024,seedream2025seedream40nextgenerationmultimodal}.
The combination of multiple models helps reduce the bias of generated images with respect to any single model.
We begin by prompting Gemini 2.5 Flash to generate short, vague prompts (e.g., ``silence practicing resonance'', ``a place that never was'').
These prompts give us unique general ideas of interesting scenes and objects, however, these prompts as is are not sufficient as inputs to text-to-image models to create high quality, diverse images.
Therefore, we use Gemini in a second stage, where we expand these prompts into longer, more specific prompts to serve as input to text-to-image models, where we task Gemini with providing three different concrete stylistic interpretations of the given vague concepts.

We split the full set of expanded prompts between Reve, FIBO, and Seedream4, and generate images from them.
Examples of generated images form this set are shown in \Cref{fig:data_gen_vague}.

\newcommand{\dataimgwidth}{0.3\linewidth}

\begin{figure}[t]
    \centering
    \setlength{\tabcolsep}{3pt}
    \begin{tabular}{ccc}
        \includegraphics[width=\dataimgwidth]{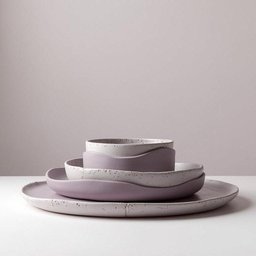} &
        \includegraphics[width=\dataimgwidth]{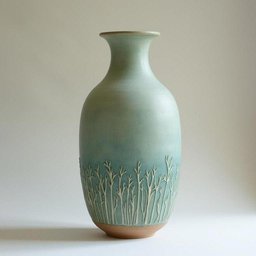} &
        \includegraphics[width=\dataimgwidth]{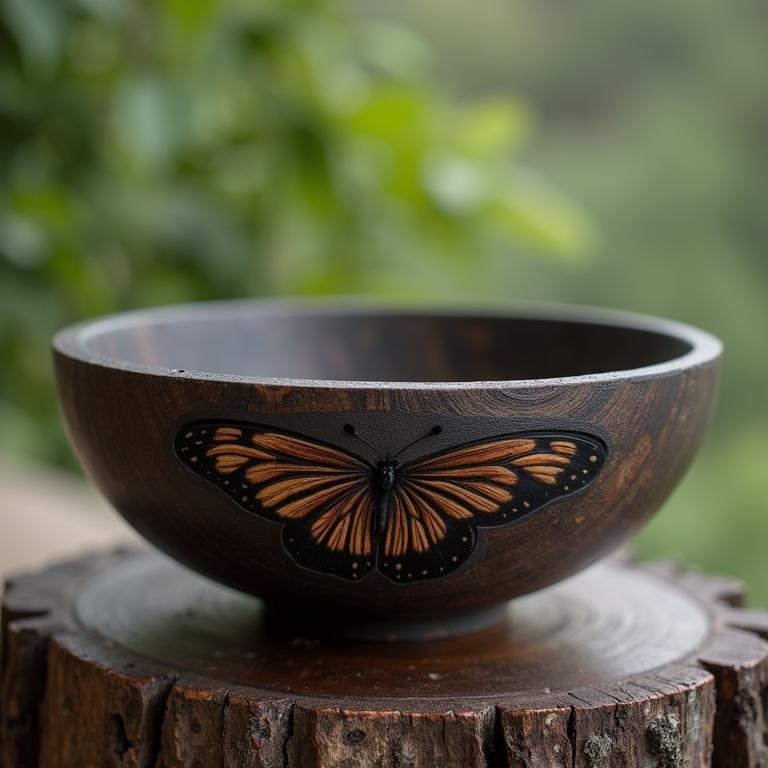} \\[2pt]
        \includegraphics[width=\dataimgwidth]{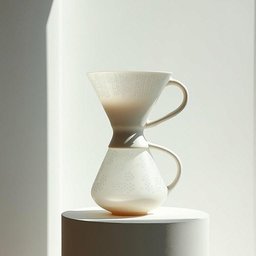} &
        \includegraphics[width=\dataimgwidth]{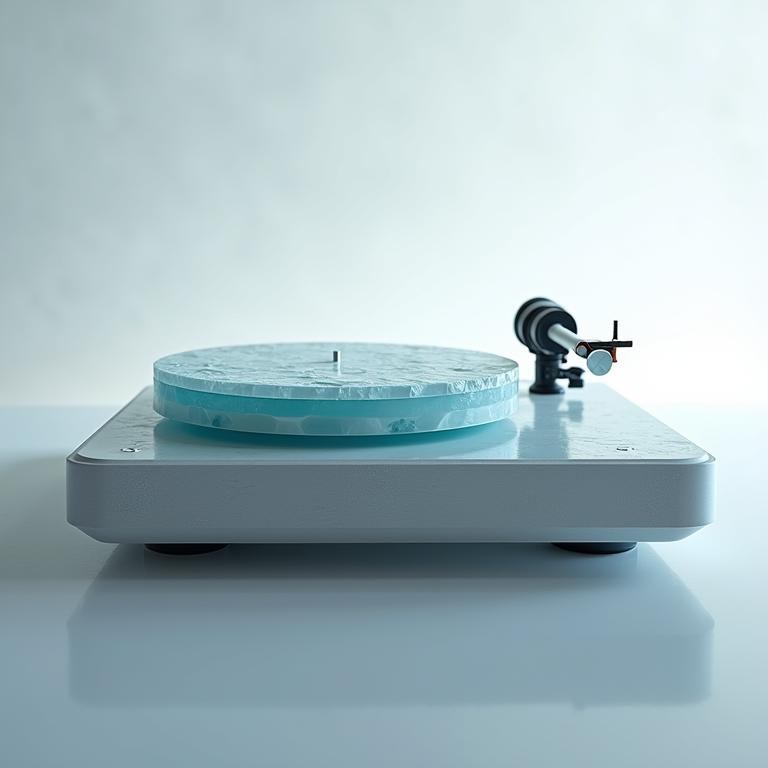} &
        \includegraphics[width=\dataimgwidth]{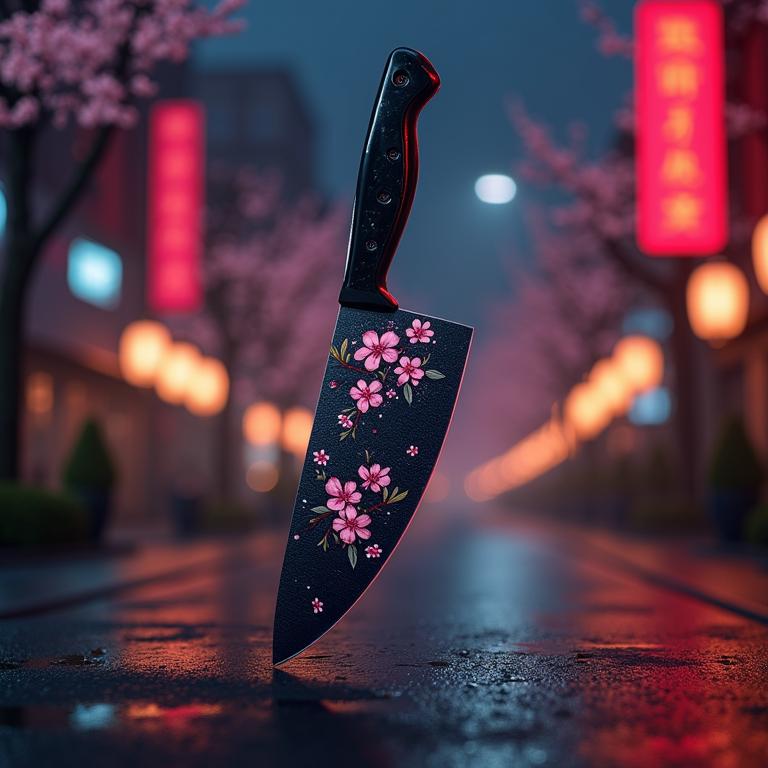} \\[2pt]
        \includegraphics[width=\dataimgwidth]{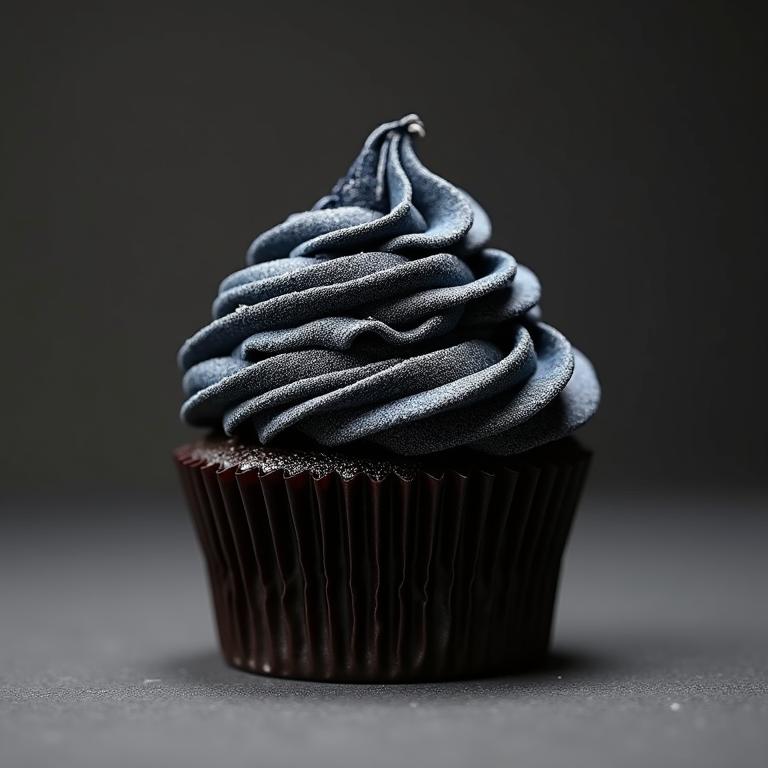} &
        \includegraphics[width=\dataimgwidth]{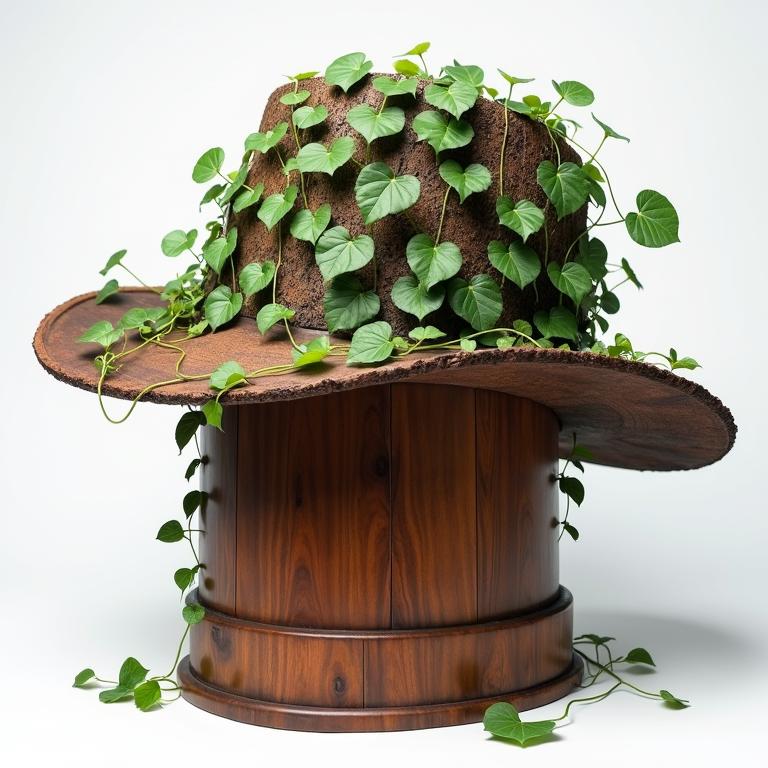} &
        \includegraphics[width=\dataimgwidth]{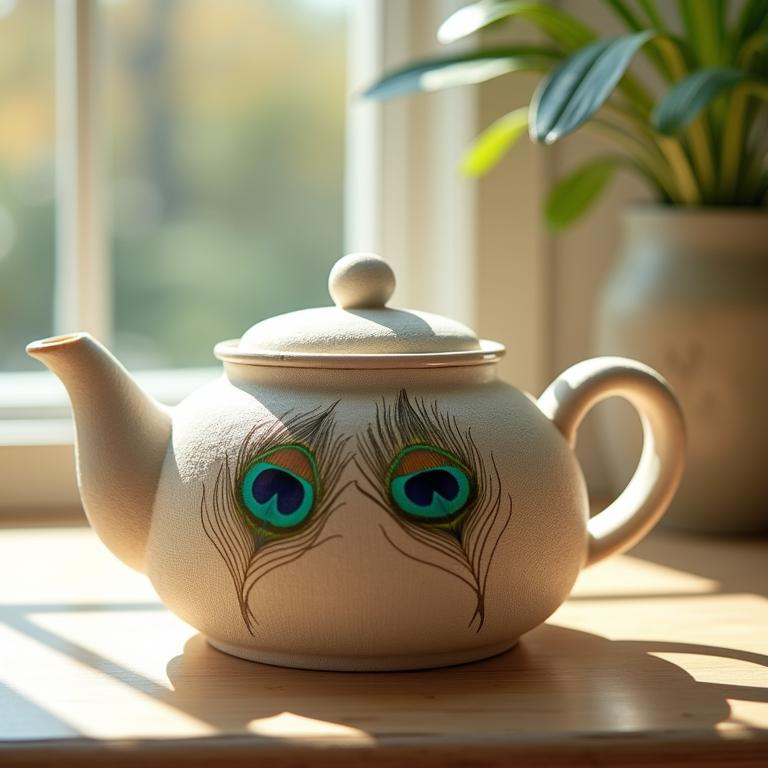} \\
    \end{tabular}
    \caption{Examples of images in our data pool (templated). These images are designed to contain multiple distinct visual aspects and will be decomposed by our SAE-based decomposition technique.}
    \label{fig:data_gen_templated}
\end{figure}

\begin{figure}[t]
    \centering
    \setlength{\tabcolsep}{3pt}
    \begin{tabular}{ccc}
        \includegraphics[width=\dataimgwidth]{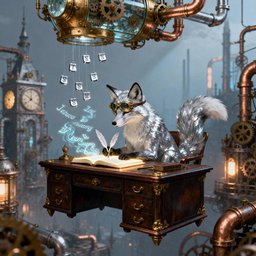} &
        \includegraphics[width=\dataimgwidth]{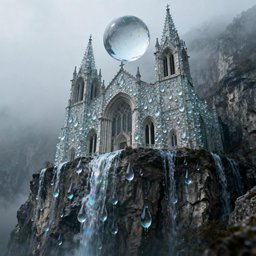} &
        \includegraphics[width=\dataimgwidth]{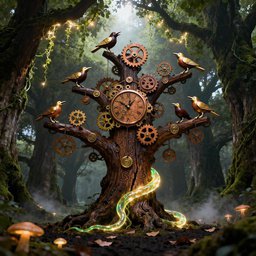} \\[2pt]
        \includegraphics[width=\dataimgwidth]{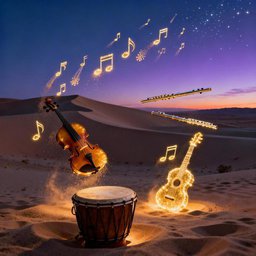} &
        \includegraphics[width=\dataimgwidth]{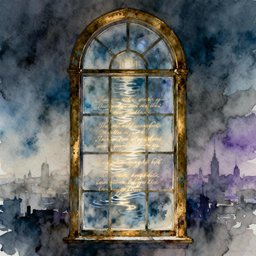} &
        \includegraphics[width=\dataimgwidth]{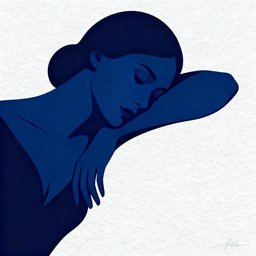} \\[2pt]
        \includegraphics[width=\dataimgwidth]{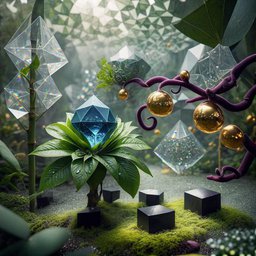} &
        \includegraphics[width=\dataimgwidth]{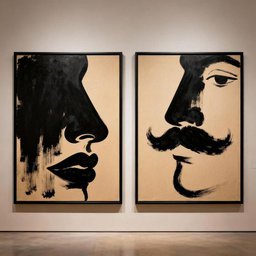} &
        \includegraphics[width=\dataimgwidth]{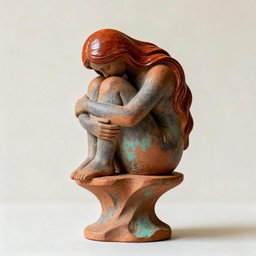} \\
    \end{tabular}
    \caption{Examples of images in our data pool (vague prompts). These images are designed to contain multiple distinct visual aspects and will be decomposed by our SAE-based decomposition technique.}
    \label{fig:data_gen_vague}
\end{figure}

\section{Additional Results}

\subsection{Comparison with Baselines}
\begin{figure*}[p]
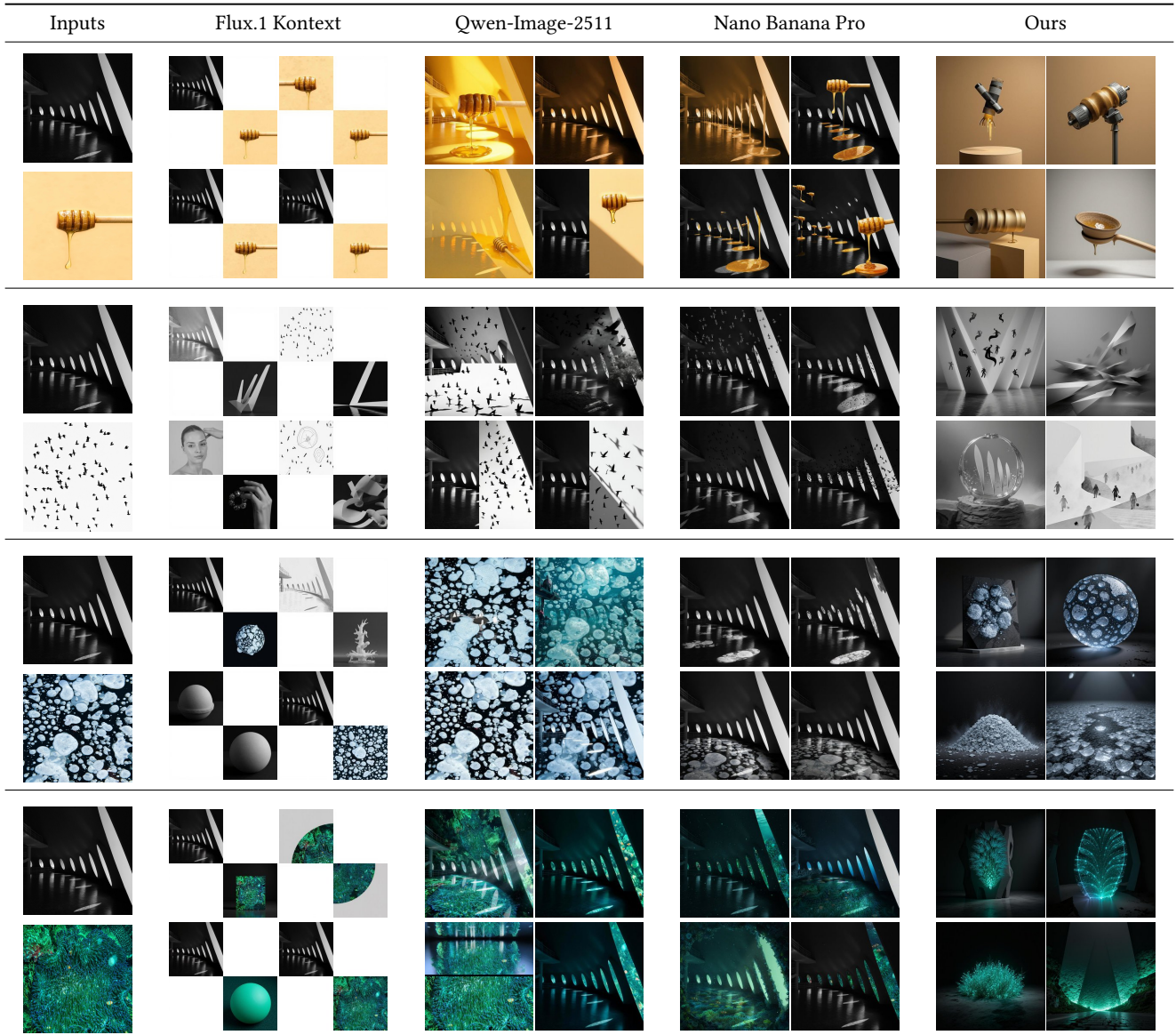

    \centering
    \setlength{\tabcolsep}{8pt}
    \begin{tabular}{ccccc}
    \toprule
        Inputs & Flux.1 Kontext & Qwen-Image-2511 & Nano Banana Pro & Ours \\[2pt]
        \hline \\[-6pt]
        \comprow{arch2__food2}{arch2}{food2} \rowsep
        \comprow{arch2__nature3}{arch2}{nature3} \rowsep
        \comprow{arch2__nature5}{arch2}{nature5} \rowsep
        \comprow{arch2__sea9}{arch2}{sea9} \\
    \end{tabular}
    \caption{Generation results comparison (1/5). Each row shows two input images (left) and outputs from four methods, each displaying results from 4 different seeds in a 2$\times$2 grid.}
    \label{fig:generation_results_comparison}
\end{figure*}

In the main paper, we show one output per method for visualization clarity. Here we provide the complete comparison results, showing all four seeds generated per method for each input pair, along with many more examples in \Cref{fig:generation_results_comparison,fig:generation_results_comparison_2,fig:generation_results_comparison_3,fig:generation_results_comparison_4,fig:generation_results_comparison_5}.

\subsection{Decomposition Results}
\begin{figure*}[p]
    \centering
    \setlength{\tabcolsep}{1pt}
    \newcommand{\imgwd}{0.070\linewidth}
    \newcommand{\methodcolm}[1]{\rotatebox{90}{\tiny\textbf{#1}}}
    \tiny
    \begin{tabular}{c@{\hspace{1pt}}ccc@{\hspace{4pt}}c@{\hspace{1pt}}ccc@{\hspace{4pt}}c@{\hspace{1pt}}ccc@{\hspace{4pt}}c@{\hspace{1pt}}ccc}
        & \textbf{Input} & \textbf{C1} & \textbf{C2} & & \textbf{Input} & \textbf{C1} & \textbf{C2} & & \textbf{Input} & \textbf{C1} & \textbf{C2} & & \textbf{Input} & \textbf{C1} & \textbf{C2} \\[1pt]
        \hline
        \\[-3pt]

        \methodcolm{Ours} &
        \raisebox{-0.5\height}{\includegraphics[width=\imgwd]{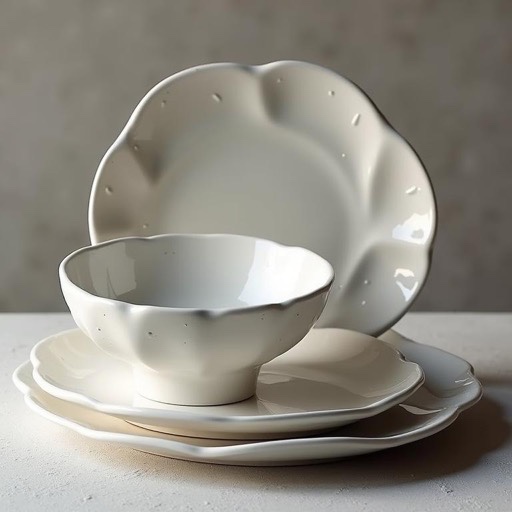}} &
        \raisebox{-0.5\height}{\includegraphics[width=\imgwd]{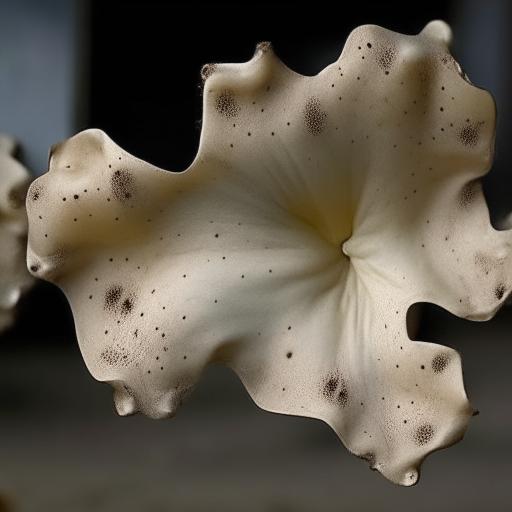}} &
        \raisebox{-0.5\height}{\includegraphics[width=\imgwd]{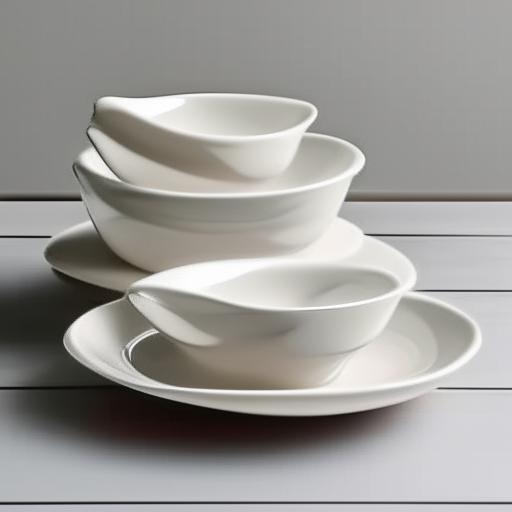}} &
        \methodcolm{Ours} &
        \raisebox{-0.5\height}{\includegraphics[width=\imgwd]{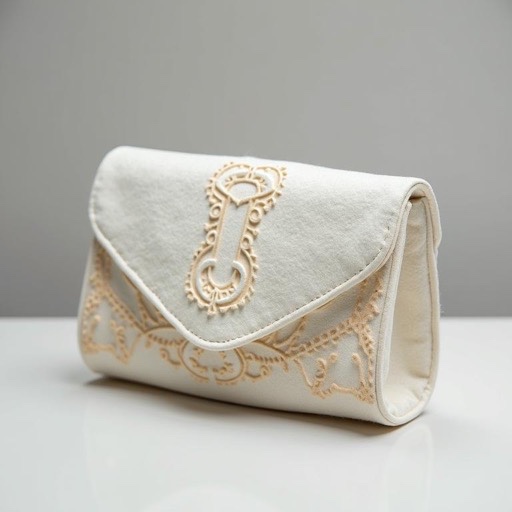}} &
        \raisebox{-0.5\height}{\includegraphics[width=\imgwd]{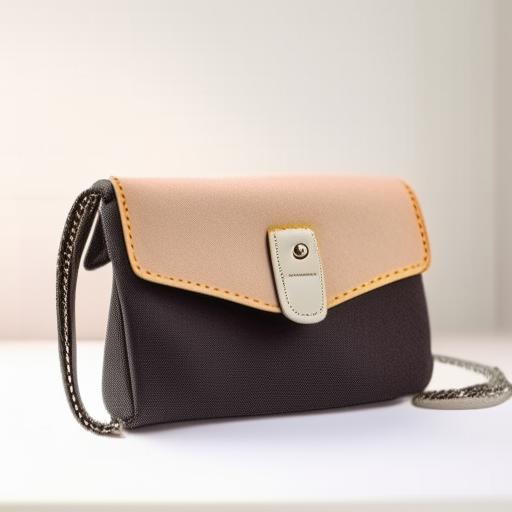}} &
        \raisebox{-0.5\height}{\includegraphics[width=\imgwd]{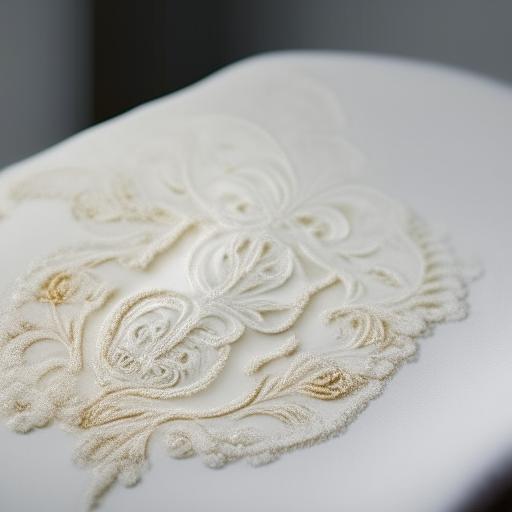}} &
        \methodcolm{Ours} &
        \raisebox{-0.5\height}{\includegraphics[width=\imgwd]{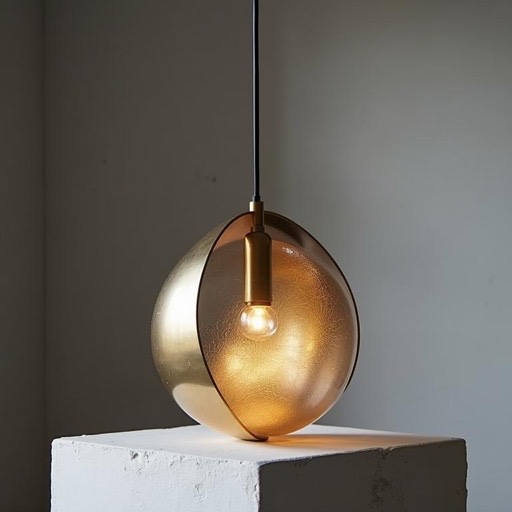}} &
        \raisebox{-0.5\height}{\includegraphics[width=\imgwd]{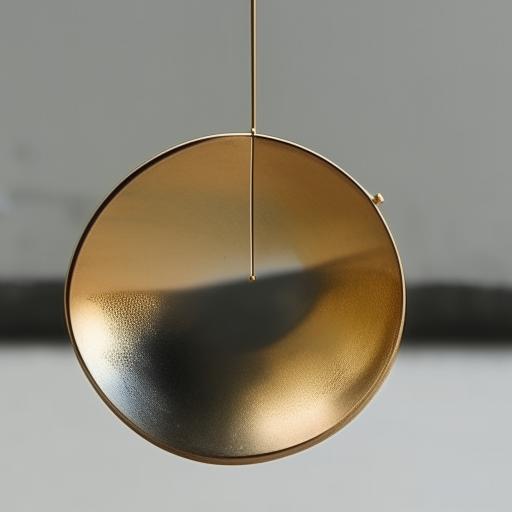}} &
        \raisebox{-0.5\height}{\includegraphics[width=\imgwd]{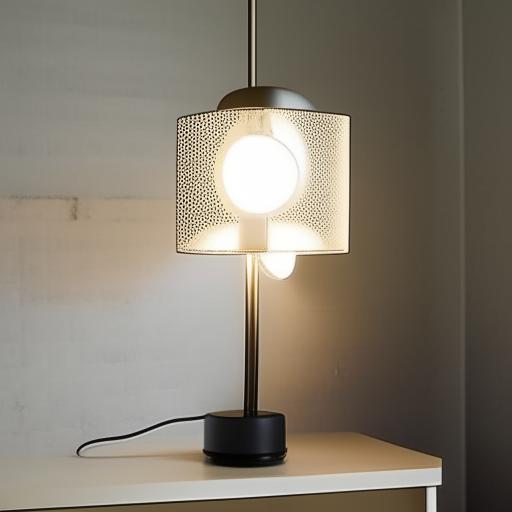}} &
        \methodcolm{Ours} &
        \raisebox{-0.5\height}{\includegraphics[width=\imgwd]{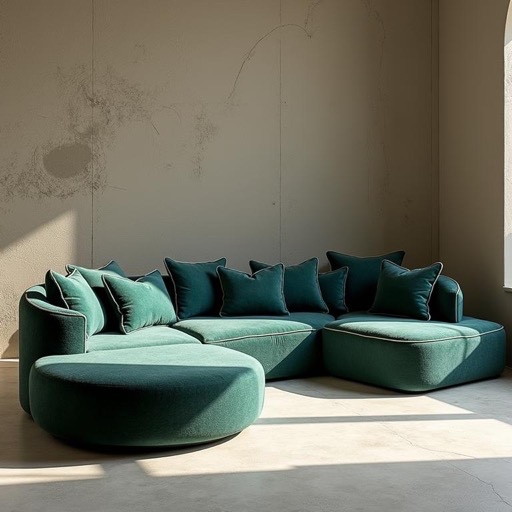}} &
        \raisebox{-0.5\height}{\includegraphics[width=\imgwd]{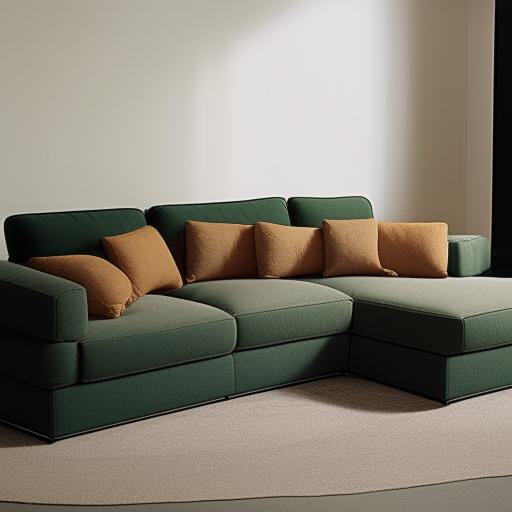}} &
        \raisebox{-0.5\height}{\includegraphics[width=\imgwd]{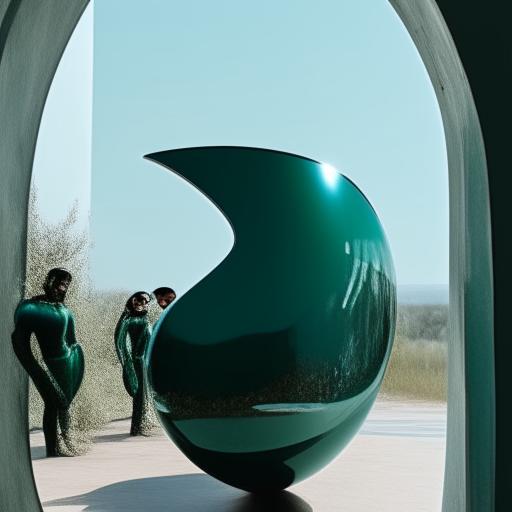}} \\[0pt]
        \methodcolm{T2I} &
        &
        \raisebox{-0.5\height}{\includegraphics[width=\imgwd]{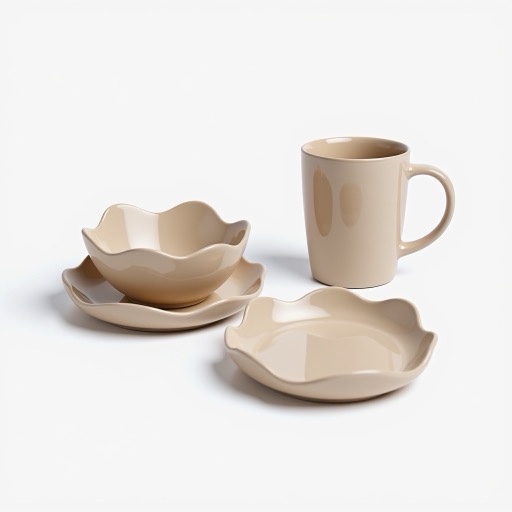}} &
        \raisebox{-0.5\height}{\includegraphics[width=\imgwd]{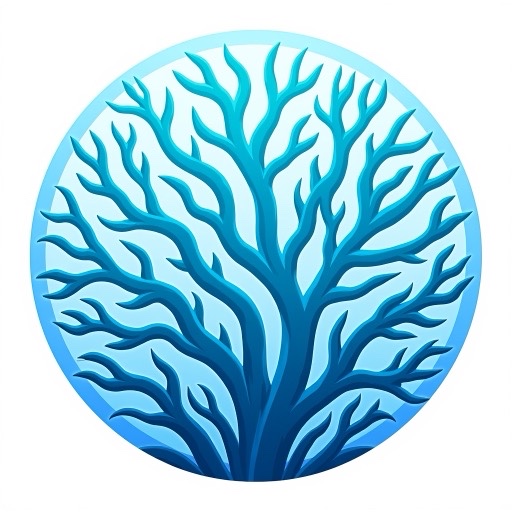}} &
        \methodcolm{T2I} &
        &
        \raisebox{-0.5\height}{\includegraphics[width=\imgwd]{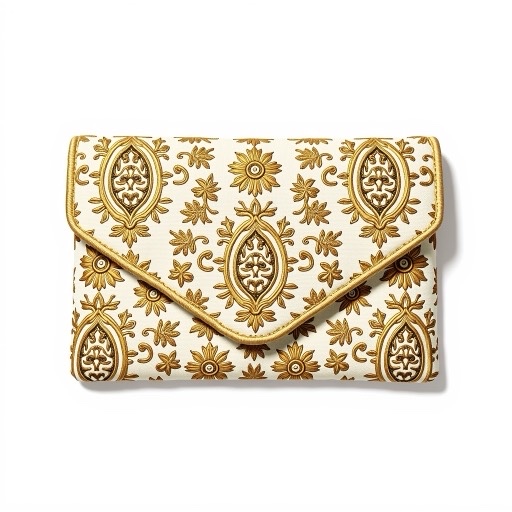}} &
        \raisebox{-0.5\height}{\includegraphics[width=\imgwd]{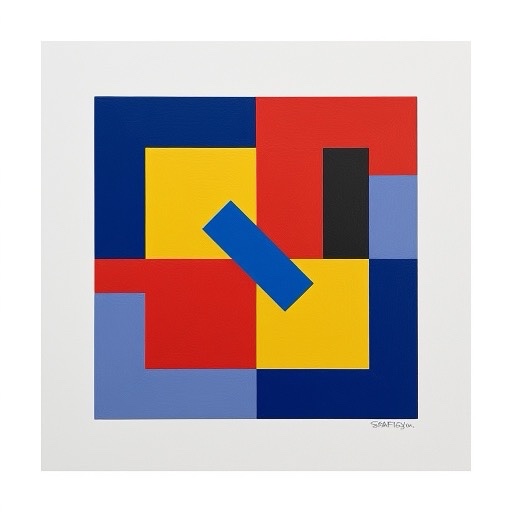}} &
        \methodcolm{T2I} &
        &
        \raisebox{-0.5\height}{\includegraphics[width=\imgwd]{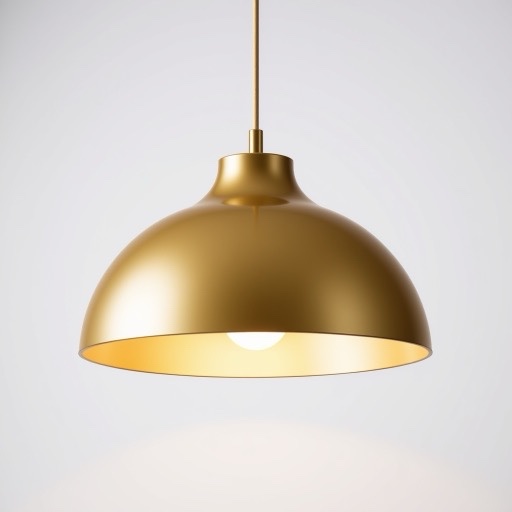}} &
        \raisebox{-0.5\height}{\includegraphics[width=\imgwd]{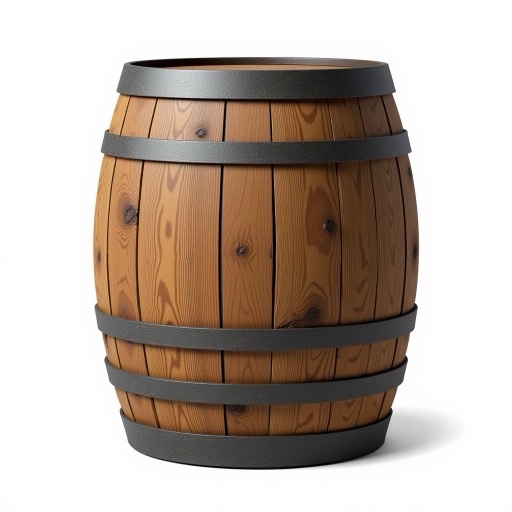}} &
        \methodcolm{T2I} &
        &
        \raisebox{-0.5\height}{\includegraphics[width=\imgwd]{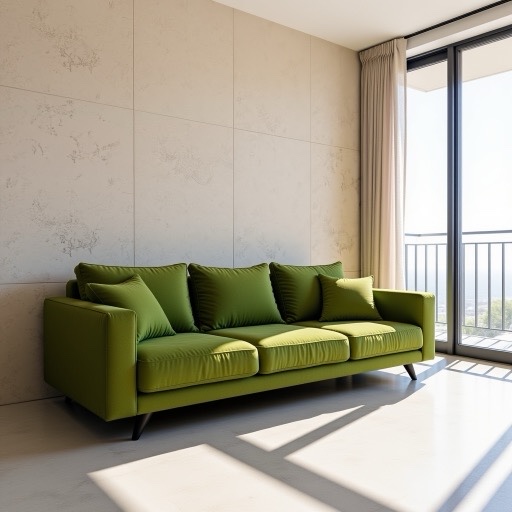}} &
        \raisebox{-0.5\height}{\includegraphics[width=\imgwd]{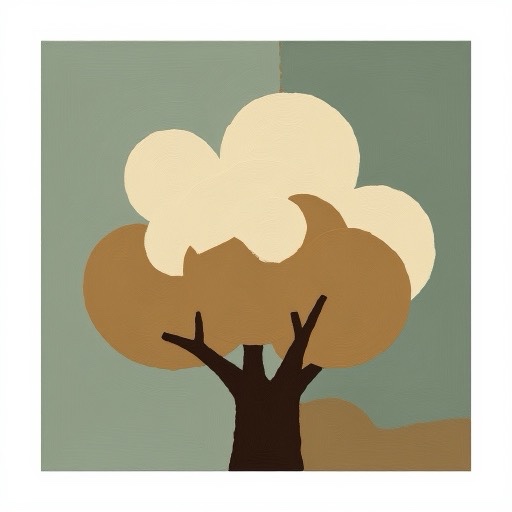}} \\[0pt]
        \methodcolm{I2I} &
        &
        \raisebox{-0.5\height}{\includegraphics[width=\imgwd]{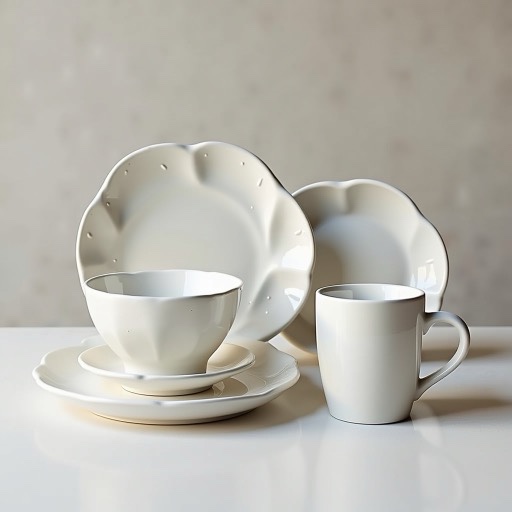}} &
        \raisebox{-0.5\height}{\includegraphics[width=\imgwd]{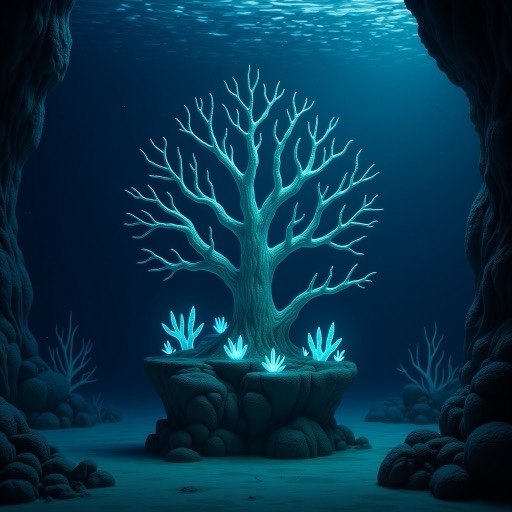}} &
        \methodcolm{I2I} &
        &
        \raisebox{-0.5\height}{\includegraphics[width=\imgwd]{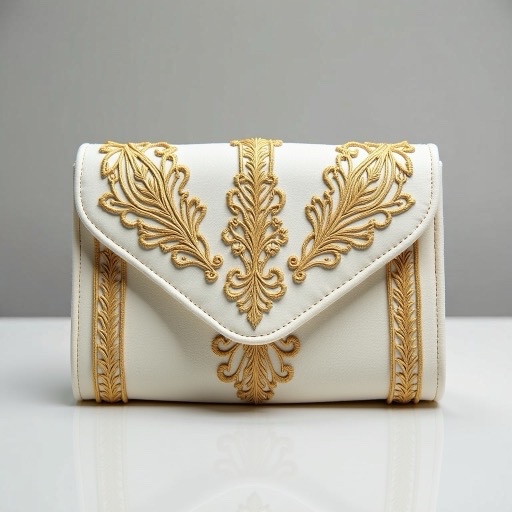}} &
        \raisebox{-0.5\height}{\includegraphics[width=\imgwd]{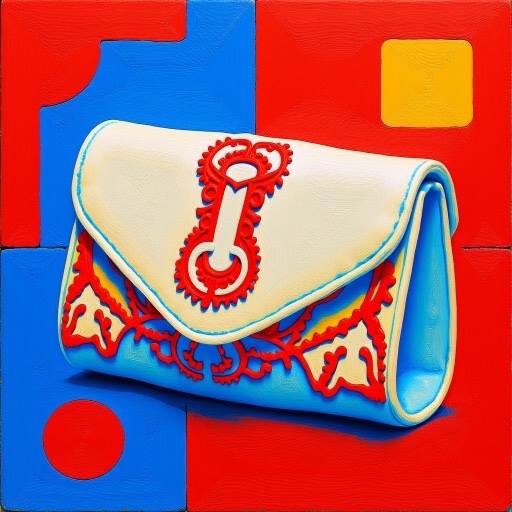}} &
        \methodcolm{I2I} &
        &
        \raisebox{-0.5\height}{\includegraphics[width=\imgwd]{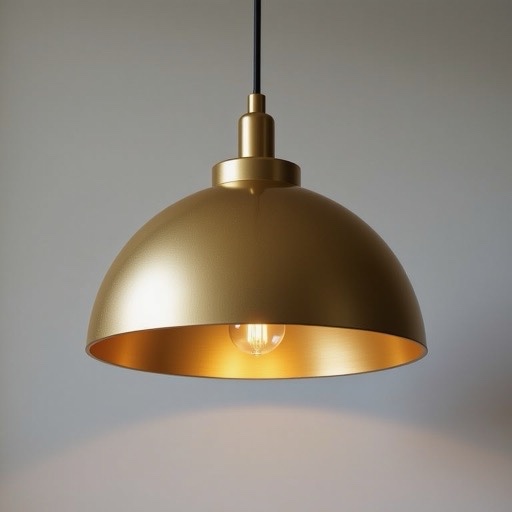}} &
        \raisebox{-0.5\height}{\includegraphics[width=\imgwd]{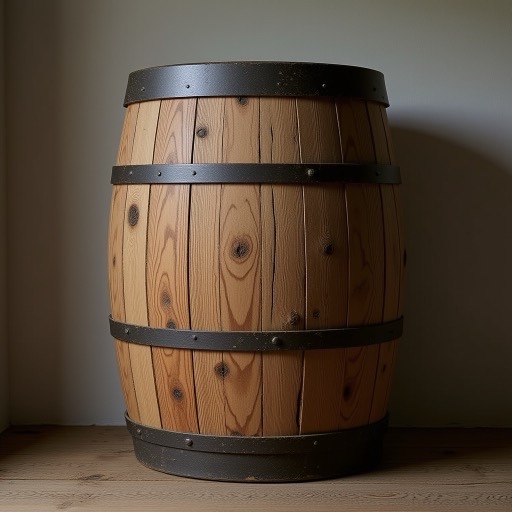}} &
        \methodcolm{I2I} &
        &
        \raisebox{-0.5\height}{\includegraphics[width=\imgwd]{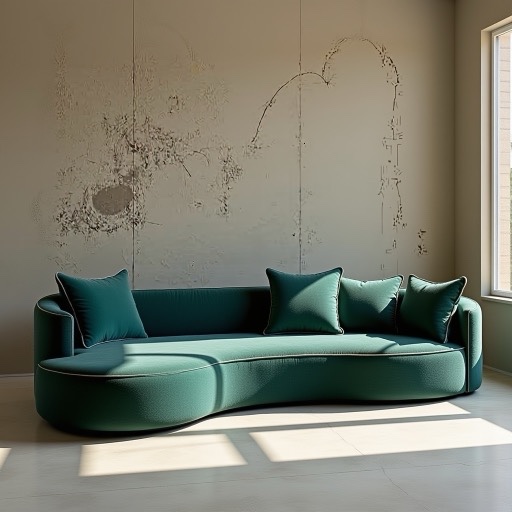}} &
        \raisebox{-0.5\height}{\includegraphics[width=\imgwd]{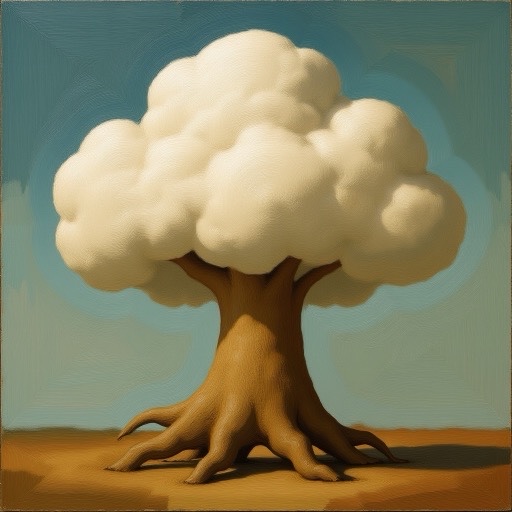}} \\[2pt]
        \hline
        \\[-3pt]

        \methodcolm{Ours} &
        \raisebox{-0.5\height}{\includegraphics[width=\imgwd]{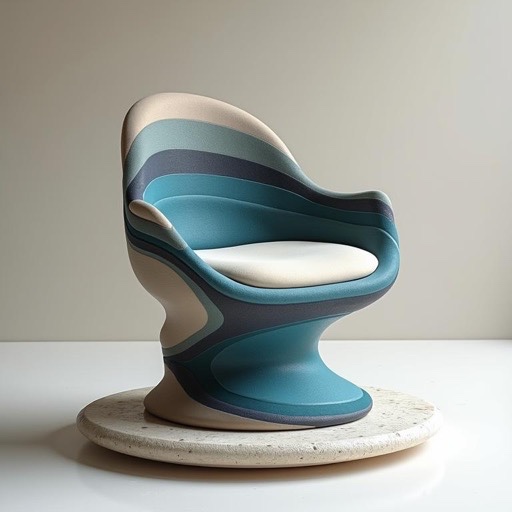}} &
        \raisebox{-0.5\height}{\includegraphics[width=\imgwd]{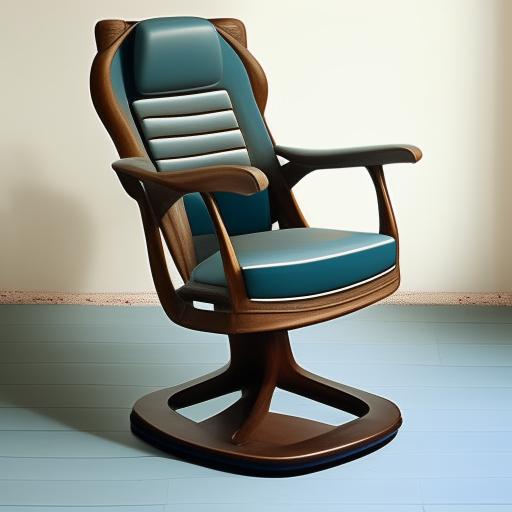}} &
        \raisebox{-0.5\height}{\includegraphics[width=\imgwd]{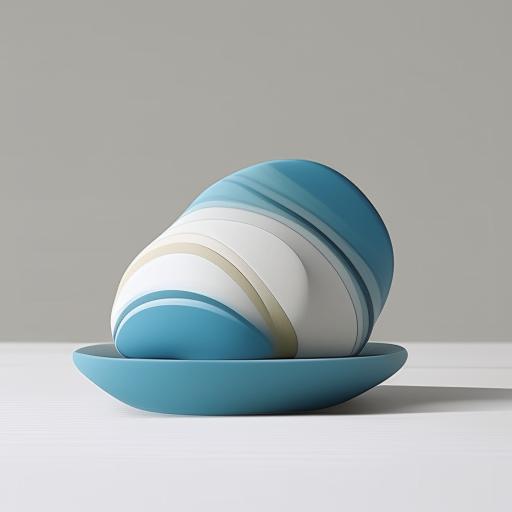}} &
        \methodcolm{Ours} &
        \raisebox{-0.5\height}{\includegraphics[width=\imgwd]{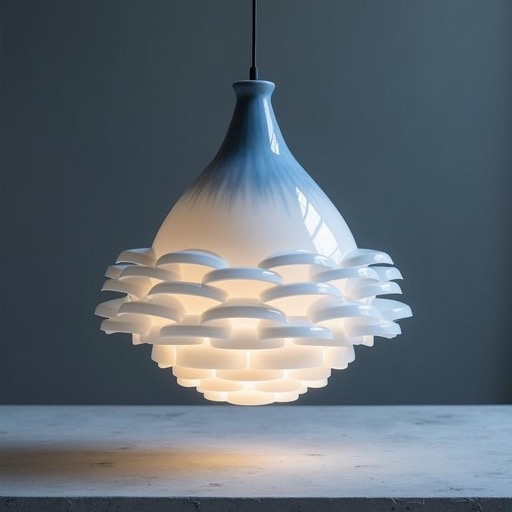}} &
        \raisebox{-0.5\height}{\includegraphics[width=\imgwd]{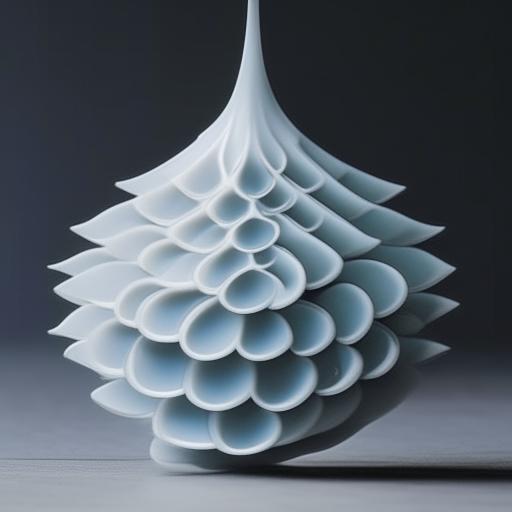}} &
        \raisebox{-0.5\height}{\includegraphics[width=\imgwd]{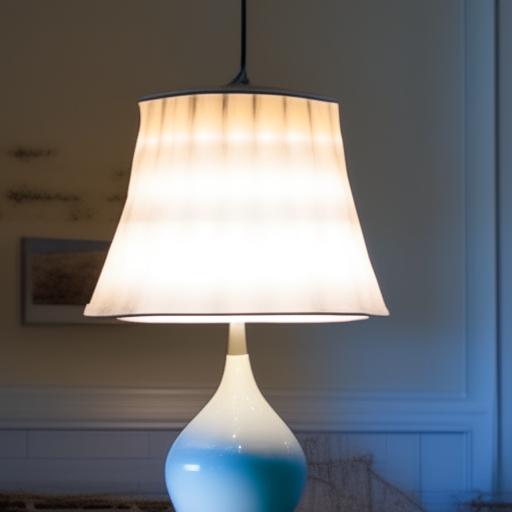}} &
        \methodcolm{Ours} &
        \raisebox{-0.5\height}{\includegraphics[width=\imgwd]{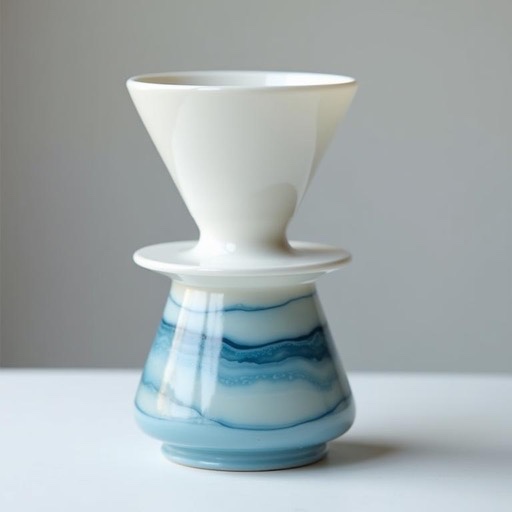}} &
        \raisebox{-0.5\height}{\includegraphics[width=\imgwd]{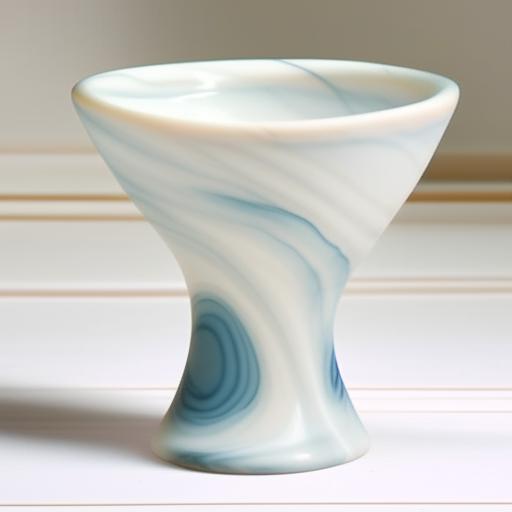}} &
        \raisebox{-0.5\height}{\includegraphics[width=\imgwd]{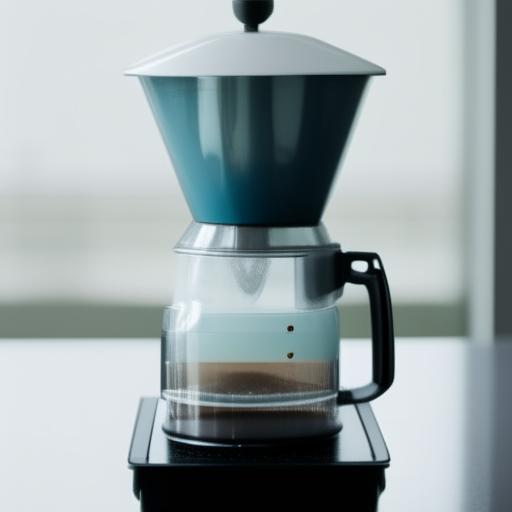}} &
        \methodcolm{Ours} &
        \raisebox{-0.5\height}{\includegraphics[width=\imgwd]{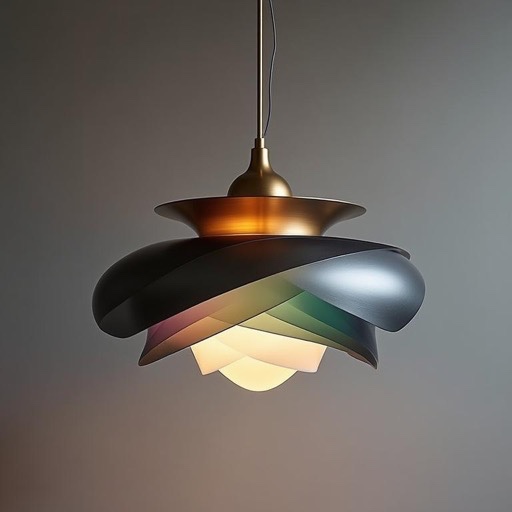}} &
        \raisebox{-0.5\height}{\includegraphics[width=\imgwd]{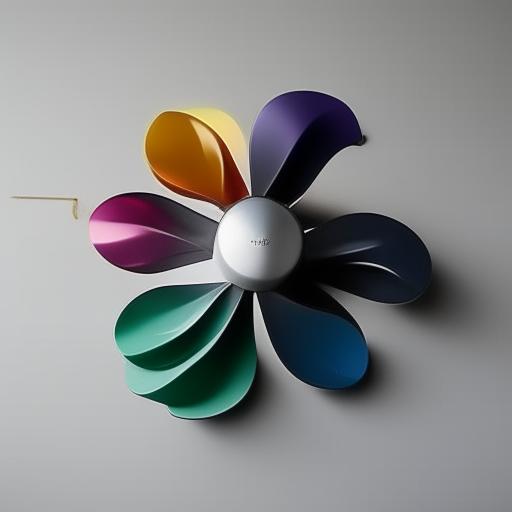}} &
        \raisebox{-0.5\height}{\includegraphics[width=\imgwd]{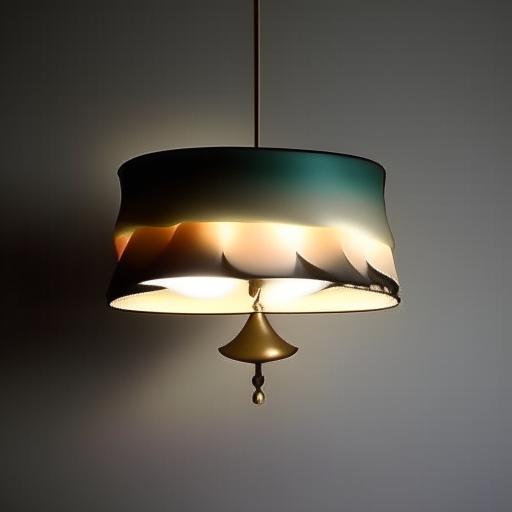}} \\[0pt]
        \methodcolm{T2I} &
        &
        \raisebox{-0.5\height}{\includegraphics[width=\imgwd]{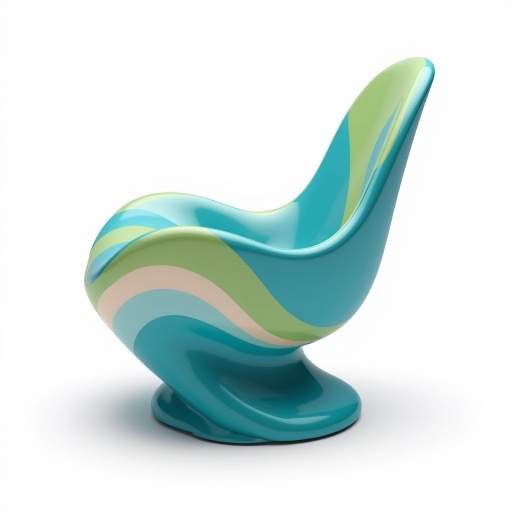}} &
        \raisebox{-0.5\height}{\includegraphics[width=\imgwd]{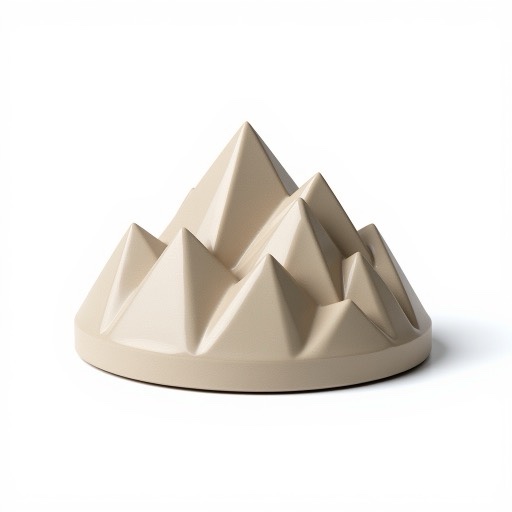}} &
        \methodcolm{T2I} &
        &
        \raisebox{-0.5\height}{\includegraphics[width=\imgwd]{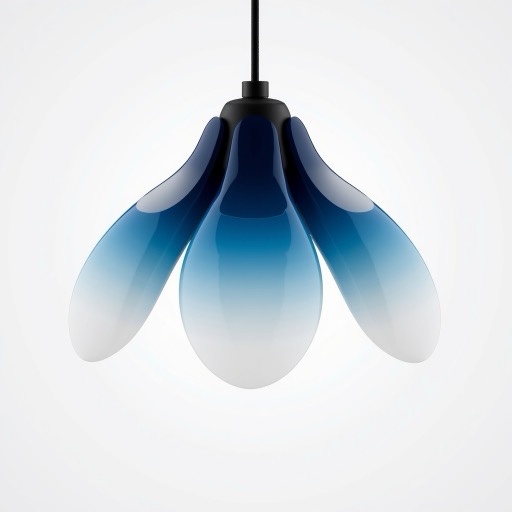}} &
        \raisebox{-0.5\height}{\includegraphics[width=\imgwd]{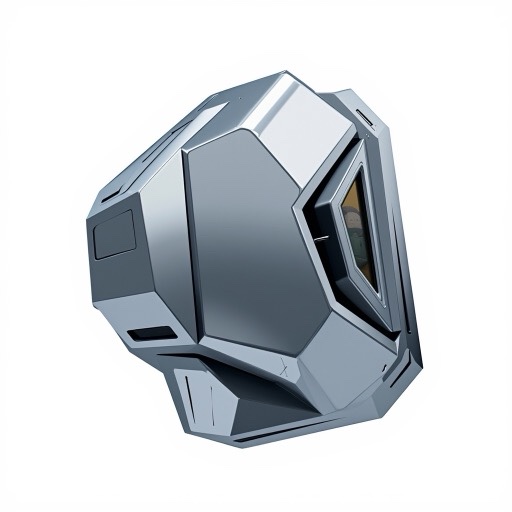}} &
        \methodcolm{T2I} &
        &
        \raisebox{-0.5\height}{\includegraphics[width=\imgwd]{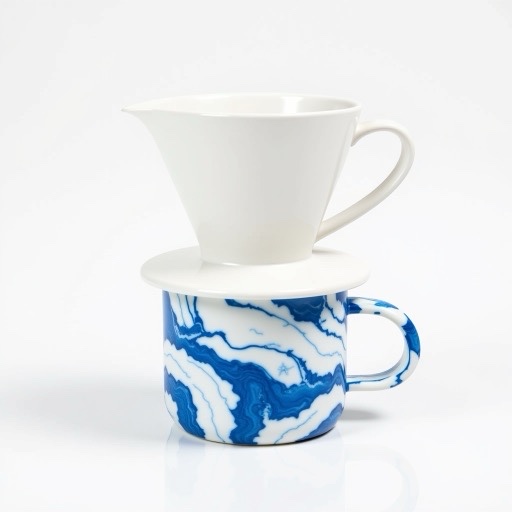}} &
        \raisebox{-0.5\height}{\includegraphics[width=\imgwd]{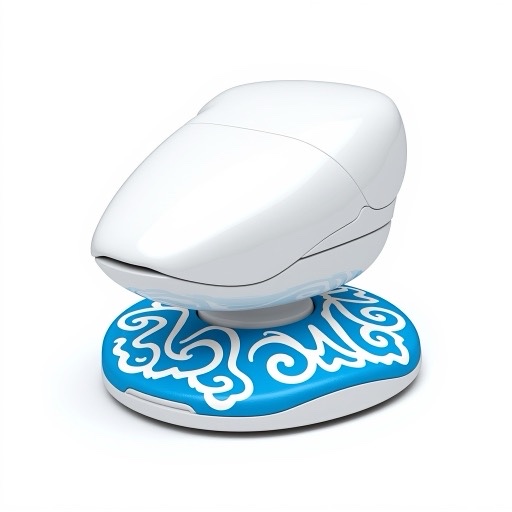}} &
        \methodcolm{T2I} &
        &
        \raisebox{-0.5\height}{\includegraphics[width=\imgwd]{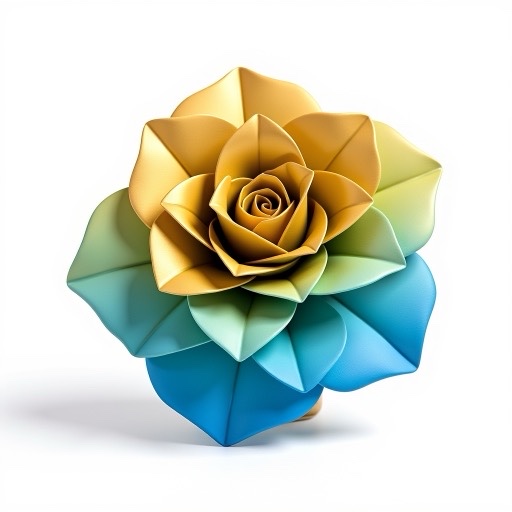}} &
        \raisebox{-0.5\height}{\includegraphics[width=\imgwd]{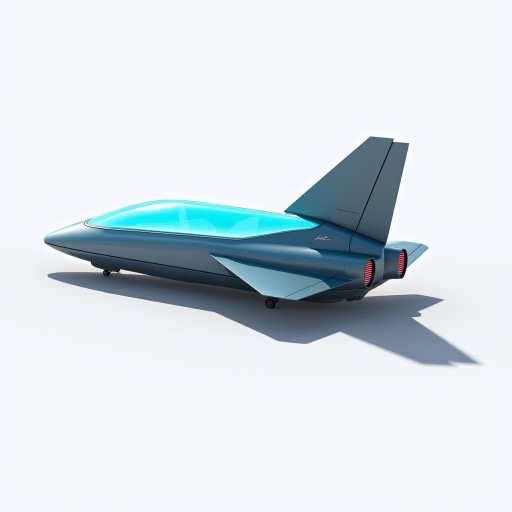}} \\[0pt]
        \methodcolm{I2I} &
        &
        \raisebox{-0.5\height}{\includegraphics[width=\imgwd]{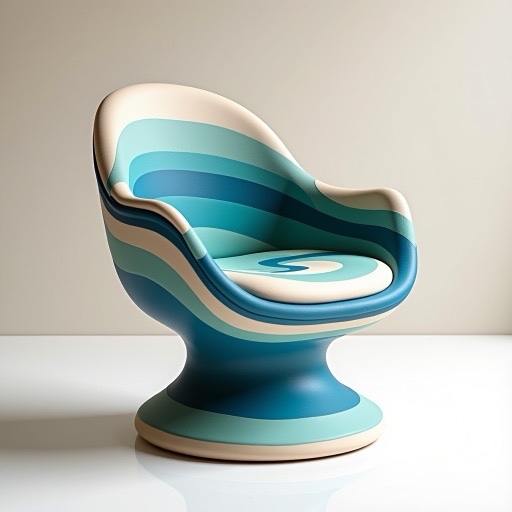}} &
        \raisebox{-0.5\height}{\includegraphics[width=\imgwd]{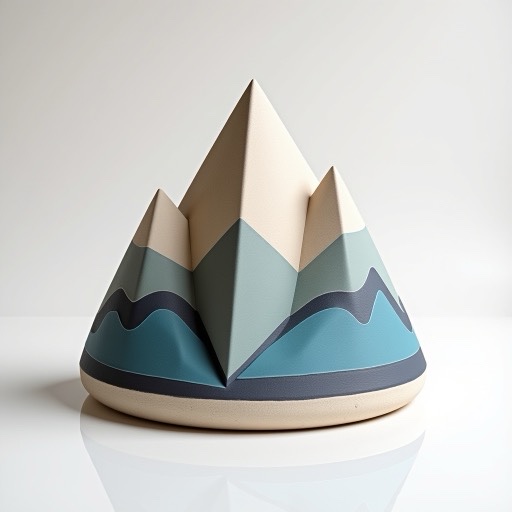}} &
        \methodcolm{I2I} &
        &
        \raisebox{-0.5\height}{\includegraphics[width=\imgwd]{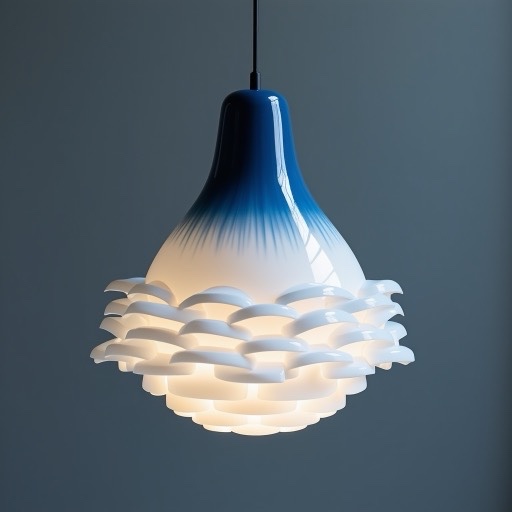}} &
        \raisebox{-0.5\height}{\includegraphics[width=\imgwd]{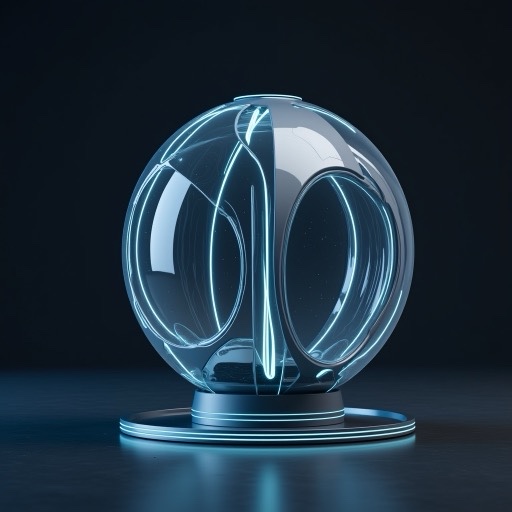}} &
        \methodcolm{I2I} &
        &
        \raisebox{-0.5\height}{\includegraphics[width=\imgwd]{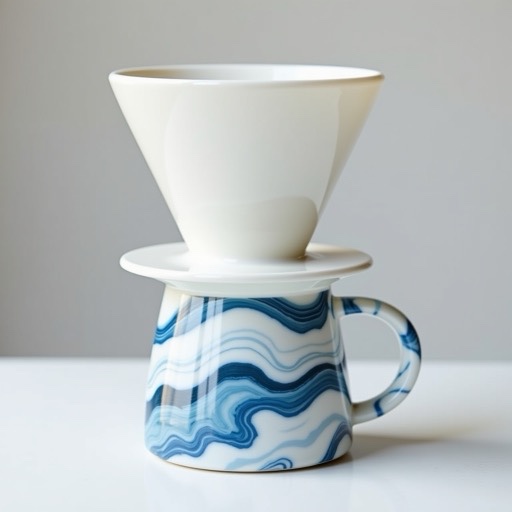}} &
        \raisebox{-0.5\height}{\includegraphics[width=\imgwd]{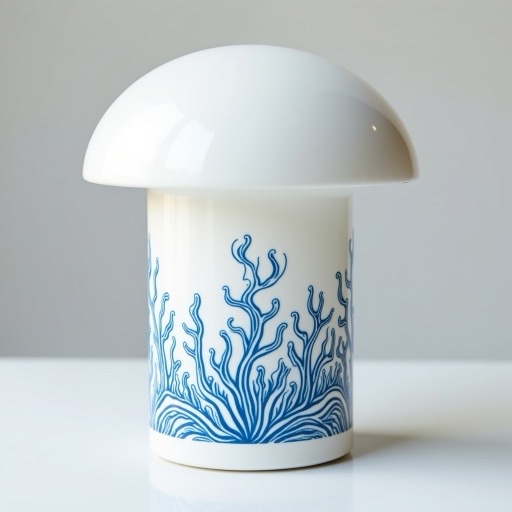}} &
        \methodcolm{I2I} &
        &
        \raisebox{-0.5\height}{\includegraphics[width=\imgwd]{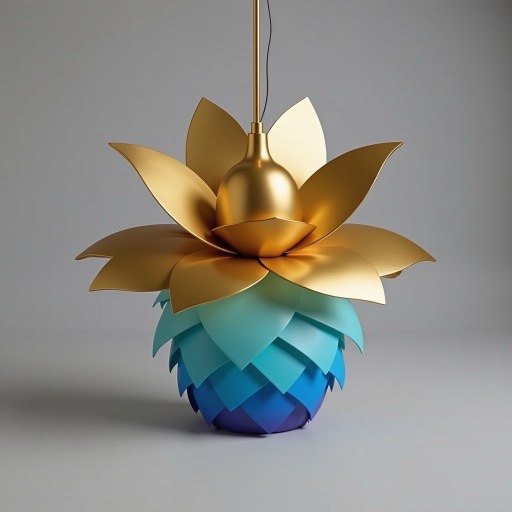}} &
        \raisebox{-0.5\height}{\includegraphics[width=\imgwd]{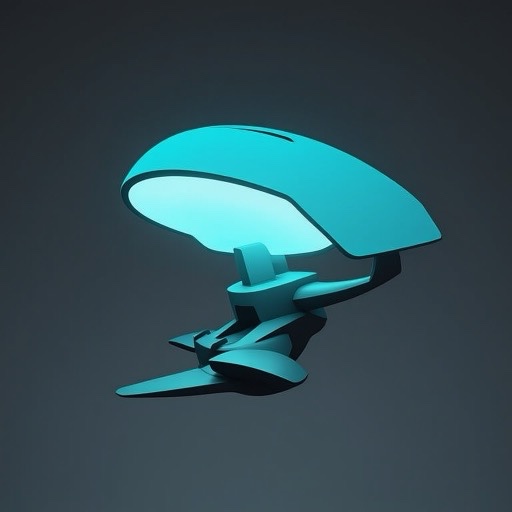}} \\[2pt]
        \hline
        \\[-3pt]

        \methodcolm{Ours} &
        \raisebox{-0.5\height}{\includegraphics[width=\imgwd]{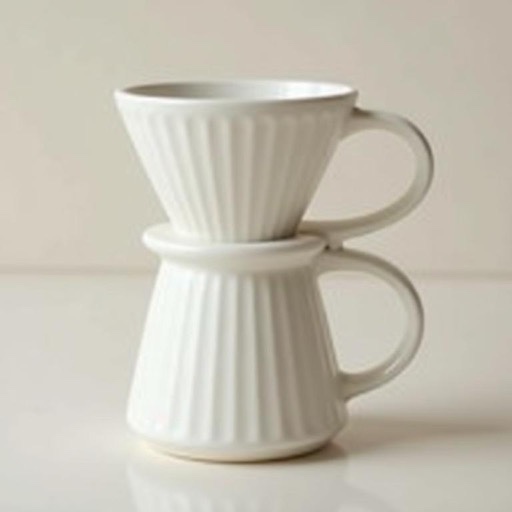}} &
        \raisebox{-0.5\height}{\includegraphics[width=\imgwd]{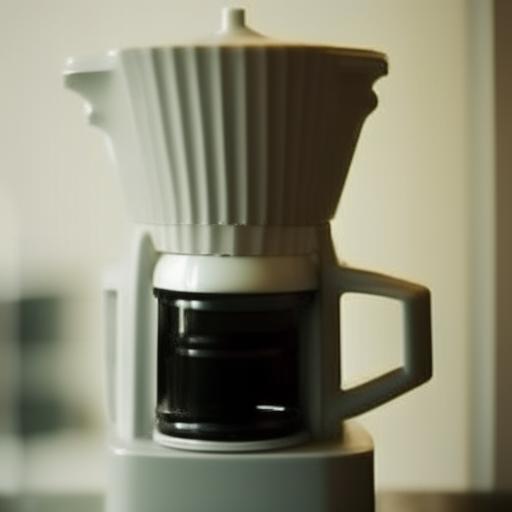}} &
        \raisebox{-0.5\height}{\includegraphics[width=\imgwd]{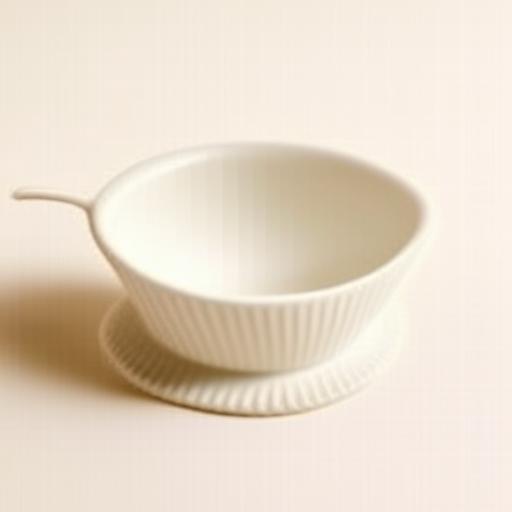}} &
        \methodcolm{Ours} &
        \raisebox{-0.5\height}{\includegraphics[width=\imgwd]{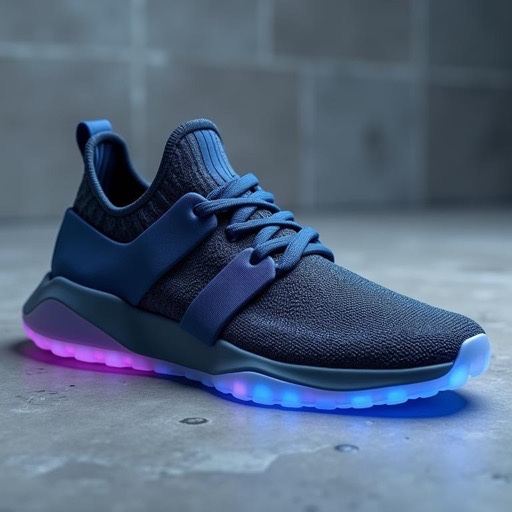}} &
        \raisebox{-0.5\height}{\includegraphics[width=\imgwd]{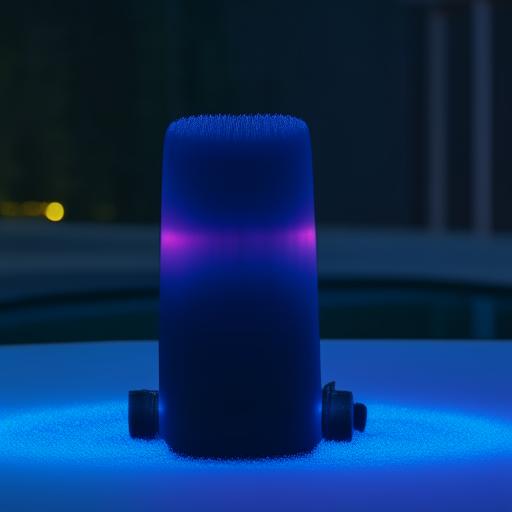}} &
        \raisebox{-0.5\height}{\includegraphics[width=\imgwd]{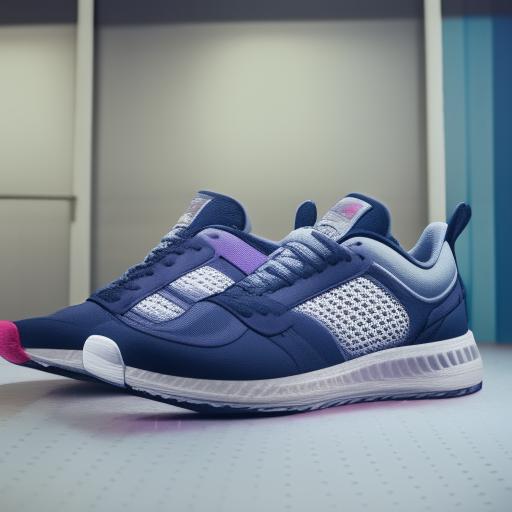}} &
        \methodcolm{Ours} &
        \raisebox{-0.5\height}{\includegraphics[width=\imgwd]{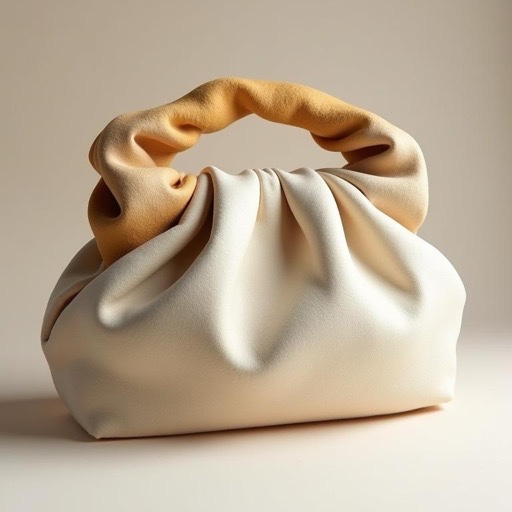}} &
        \raisebox{-0.5\height}{\includegraphics[width=\imgwd]{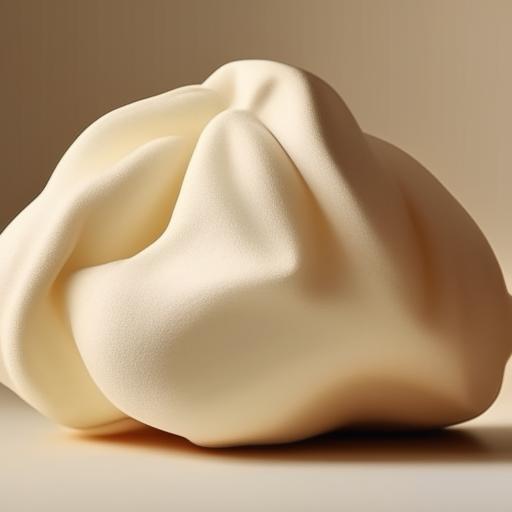}} &
        \raisebox{-0.5\height}{\includegraphics[width=\imgwd]{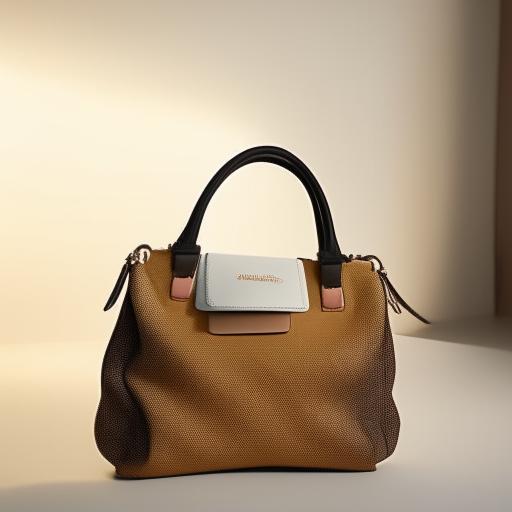}} &
        \methodcolm{Ours} &
        \raisebox{-0.5\height}{\includegraphics[width=\imgwd]{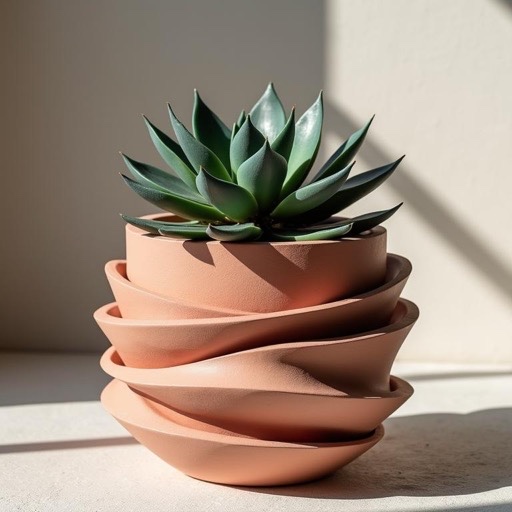}} &
        \raisebox{-0.5\height}{\includegraphics[width=\imgwd]{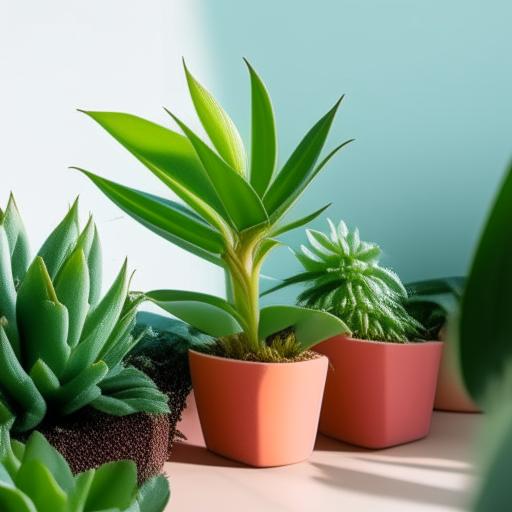}} &
        \raisebox{-0.5\height}{\includegraphics[width=\imgwd]{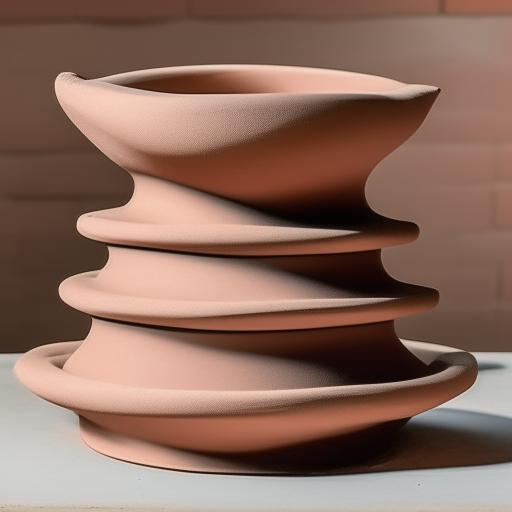}} \\[0pt]
        \methodcolm{T2I} &
        &
        \raisebox{-0.5\height}{\includegraphics[width=\imgwd]{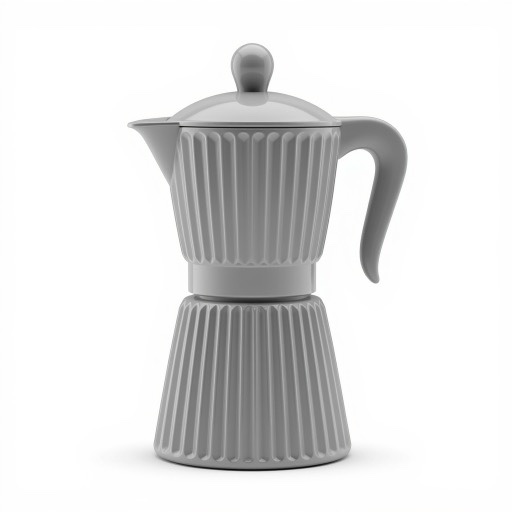}} &
        \raisebox{-0.5\height}{\includegraphics[width=\imgwd]{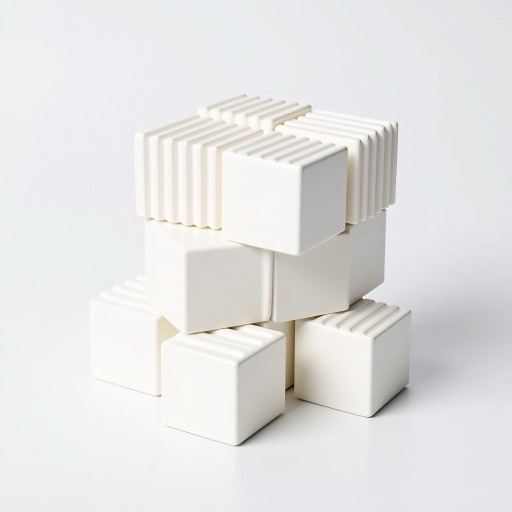}} &
        \methodcolm{T2I} &
        &
        \raisebox{-0.5\height}{\includegraphics[width=\imgwd]{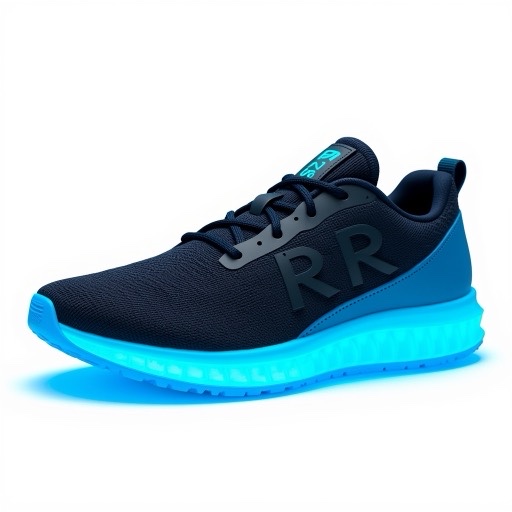}} &
        \raisebox{-0.5\height}{\includegraphics[width=\imgwd]{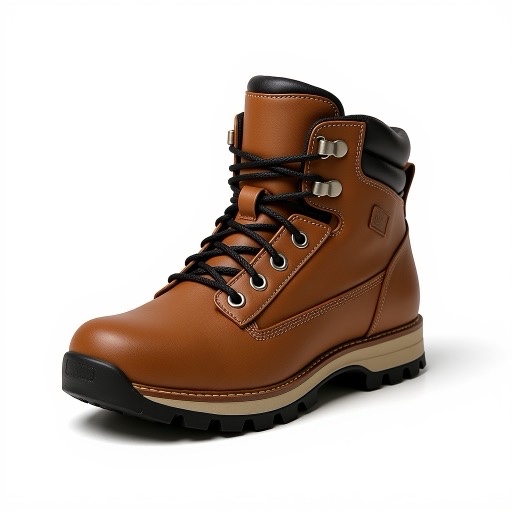}} &
        \methodcolm{T2I} &
        &
        \raisebox{-0.5\height}{\includegraphics[width=\imgwd]{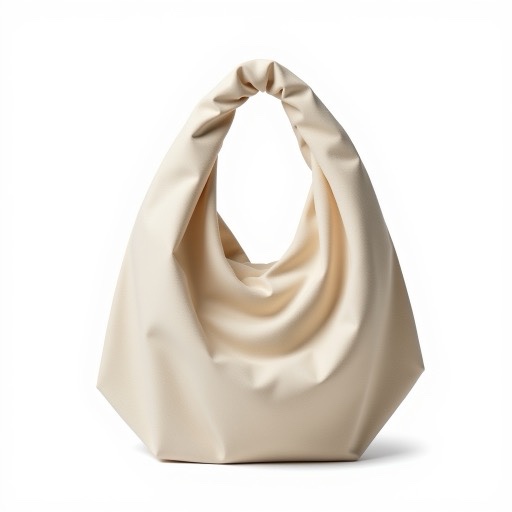}} &
        \raisebox{-0.5\height}{\includegraphics[width=\imgwd]{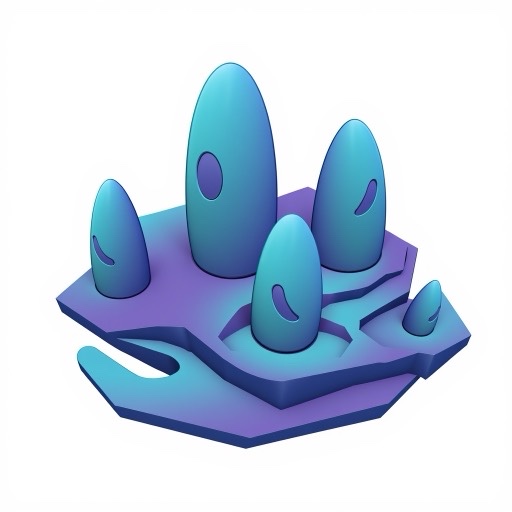}} &
        \methodcolm{T2I} &
        &
        \raisebox{-0.5\height}{\includegraphics[width=\imgwd]{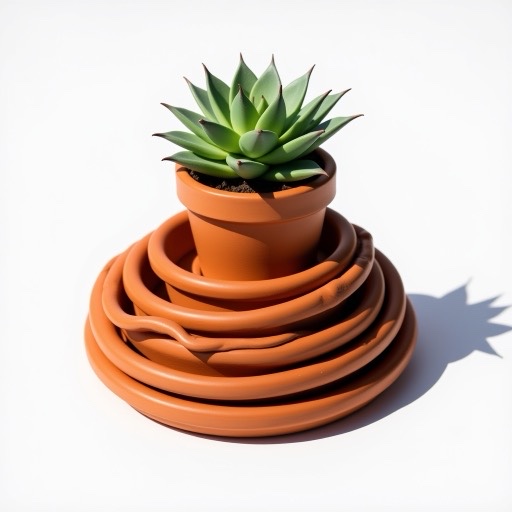}} &
        \raisebox{-0.5\height}{\includegraphics[width=\imgwd]{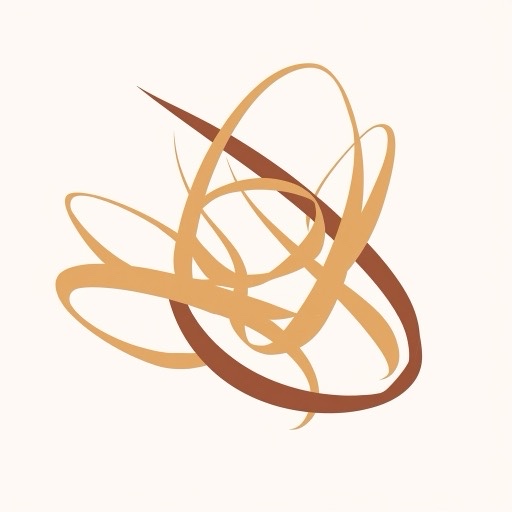}} \\[0pt]
        \methodcolm{I2I} &
        &
        \raisebox{-0.5\height}{\includegraphics[width=\imgwd]{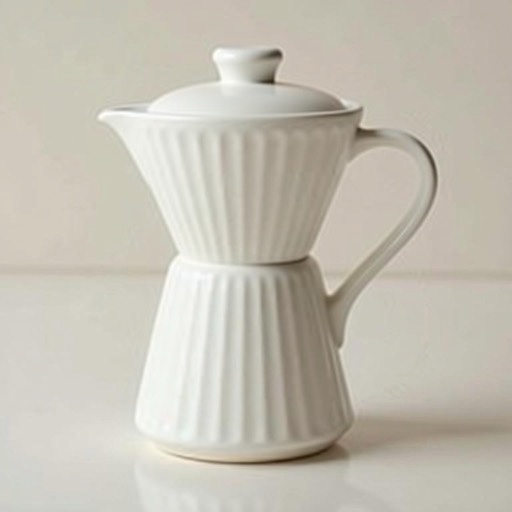}} &
        \raisebox{-0.5\height}{\includegraphics[width=\imgwd]{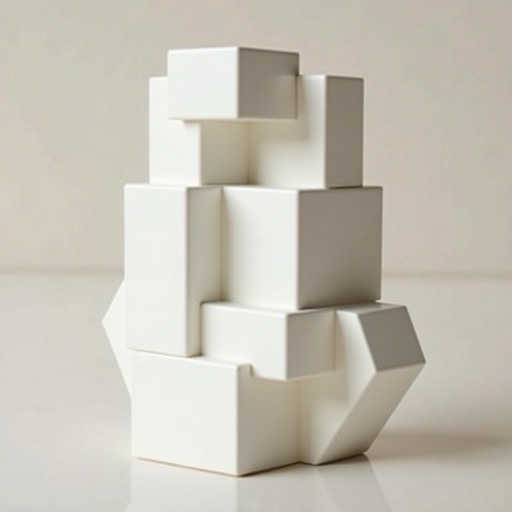}} &
        \methodcolm{I2I} &
        &
        \raisebox{-0.5\height}{\includegraphics[width=\imgwd]{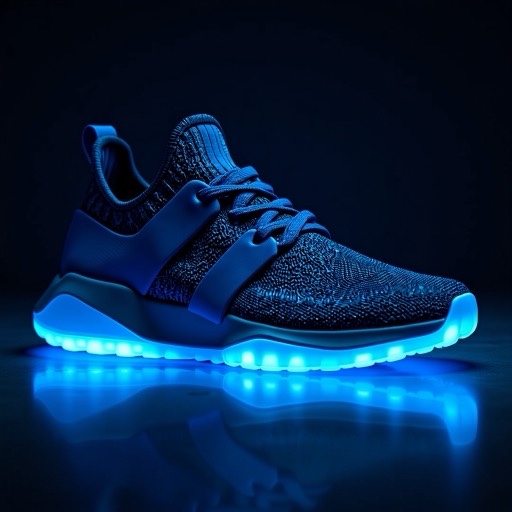}} &
        \raisebox{-0.5\height}{\includegraphics[width=\imgwd]{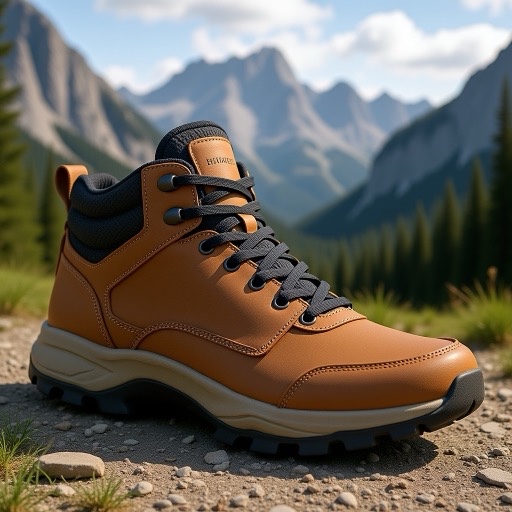}} &
        \methodcolm{I2I} &
        &
        \raisebox{-0.5\height}{\includegraphics[width=\imgwd]{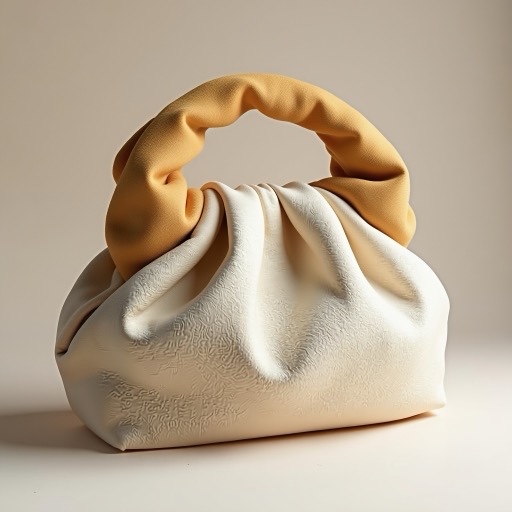}} &
        \raisebox{-0.5\height}{\includegraphics[width=\imgwd]{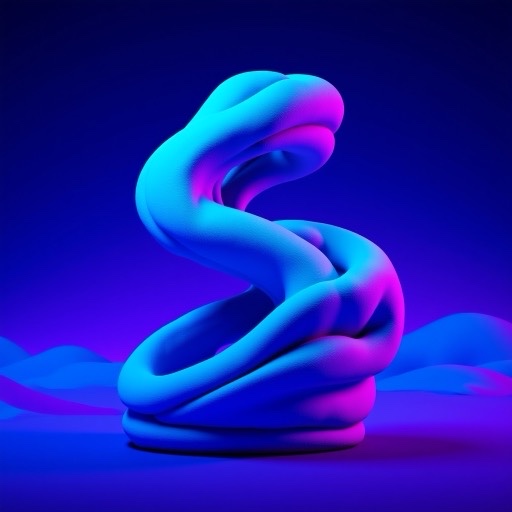}} &
        \methodcolm{I2I} &
        &
        \raisebox{-0.5\height}{\includegraphics[width=\imgwd]{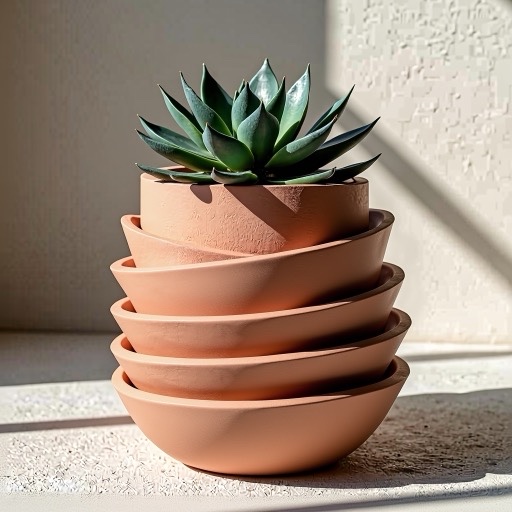}} &
        \raisebox{-0.5\height}{\includegraphics[width=\imgwd]{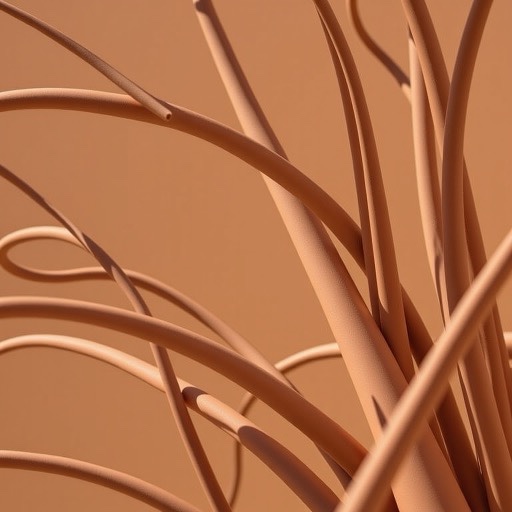}} \\
    \end{tabular}
    \caption{Decomposition results comparison. Given an input image, we decompose it into two components (C1, C2) using three methods. Our SAE-based approach produces components that capture distinct visual aspects while maintaining semantic relevance. The T2I baseline generates from VLM-provided text prompts only, while I2I uses the input image with text prompts.}
    \label{fig:decomposition_results_comparison_supp}
\end{figure*}

We provide additional decomposition examples comparing our SAE-based approach to the T2I and I2I baselines described in the main paper in \Cref{fig:decomposition_results_comparison_supp}.

\section{Comparison with CLIP Space Interpolation}
We compare our composition method against a naive baseline of interpolating in CLIP space.
Given two input images, the baseline embeds each into CLIP space, computes the mean of their embeddings, and uses Kandinsky~\cite{Razzhigaev2023KandinskyAI} to generate an image from this averaged representation.

\begin{figure*}[t]
    \centering
    \setlength{\tabcolsep}{4pt}
    \begin{tabular}{ccc|ccc}
    \toprule
        Inputs & Ours & CLIP Interp. & Inputs & Ours & CLIP Interp. \\[2pt]
        \hline \\[-6pt]

        \inputstack{arch2__food2}{arch2}{food2} &
        \methodgrid{arch2__food2}{ours} &
        \begin{tabular}{@{}c@{\hspace{1pt}}c@{}}
            \includegraphics[width=\gridimg]{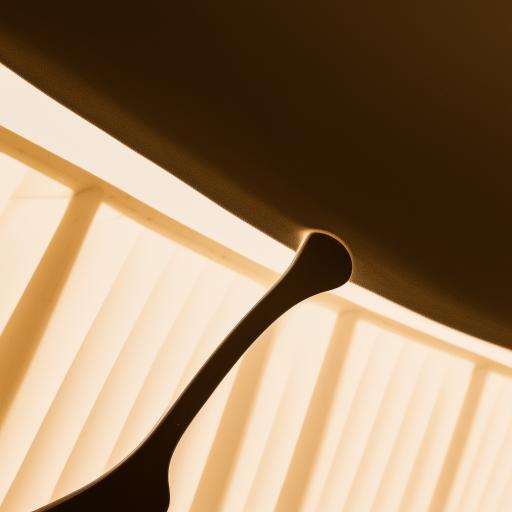} &
            \includegraphics[width=\gridimg]{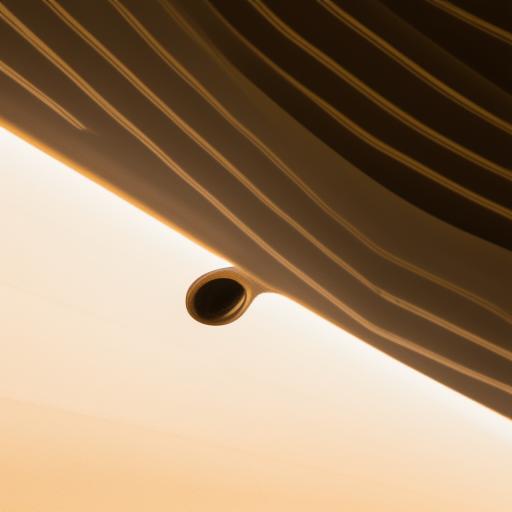} \\[-1pt]
            \includegraphics[width=\gridimg]{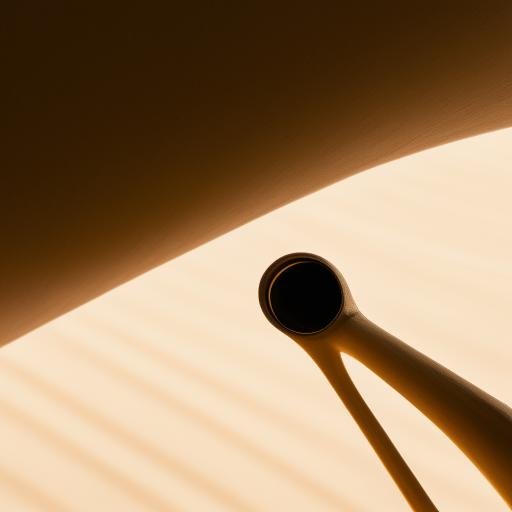} &
            \includegraphics[width=\gridimg]{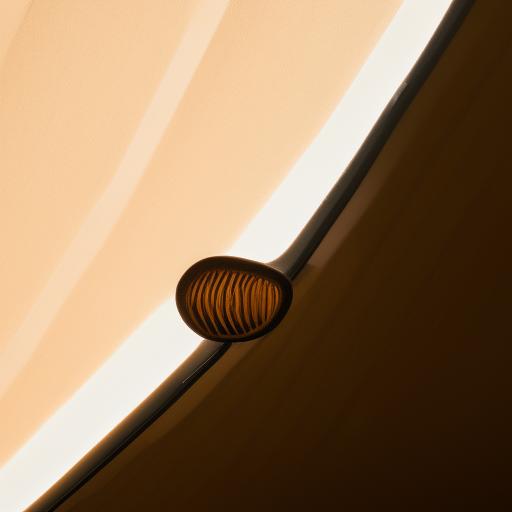}
        \end{tabular} &
        \inputstack{arch7__sea8}{arch7}{sea8} &
        \methodgrid{arch7__sea8}{ours} &
        \begin{tabular}{@{}c@{\hspace{1pt}}c@{}}
            \includegraphics[width=\gridimg]{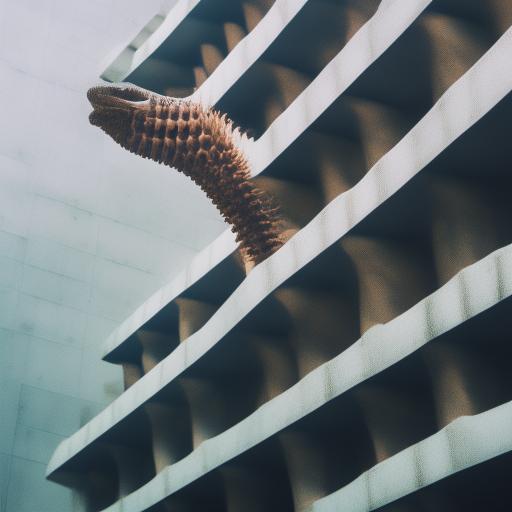} &
            \includegraphics[width=\gridimg]{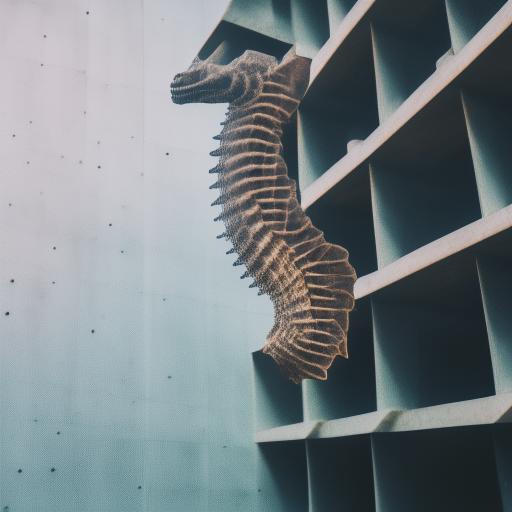} \\[-1pt]
            \includegraphics[width=\gridimg]{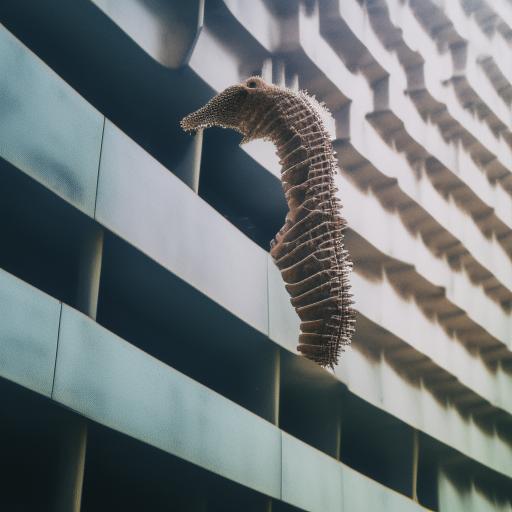} &
            \includegraphics[width=\gridimg]{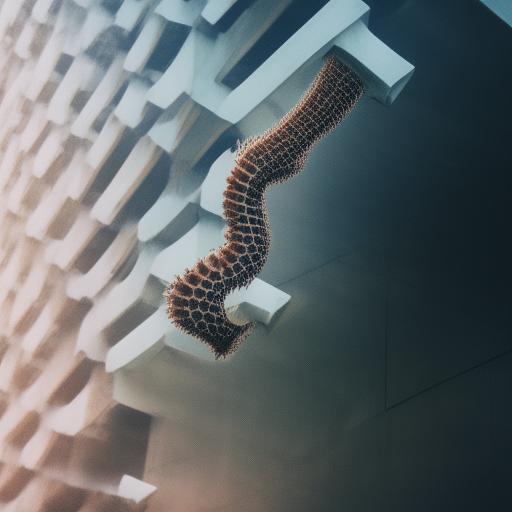}
        \end{tabular}
        \rowsep

        \inputstack{fashion3__nature6}{fashion3}{nature6} &
        \methodgrid{fashion3__nature6}{ours} &
        \begin{tabular}{@{}c@{\hspace{1pt}}c@{}}
            \includegraphics[width=\gridimg]{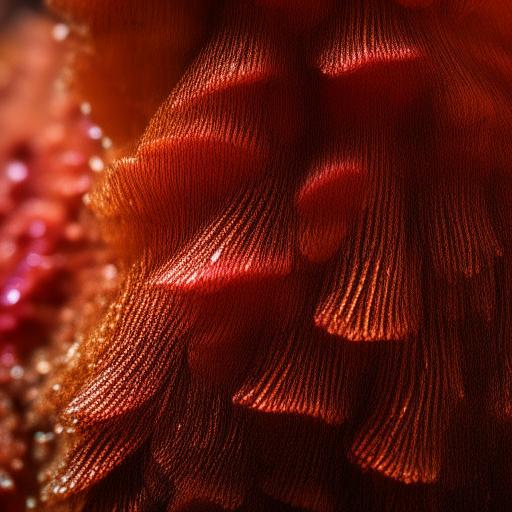} &
            \includegraphics[width=\gridimg]{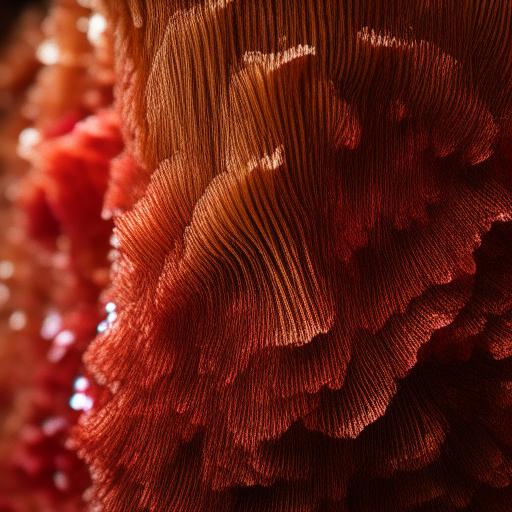} \\[-1pt]
            \includegraphics[width=\gridimg]{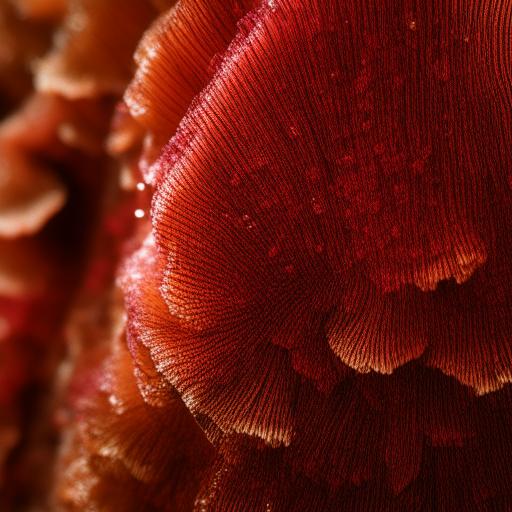} &
            \includegraphics[width=\gridimg]{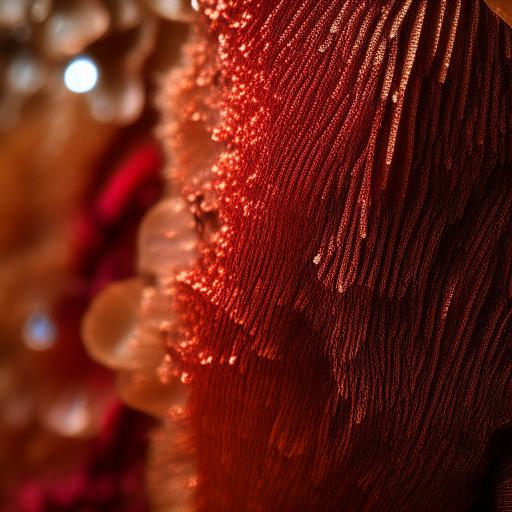}
        \end{tabular} &
        \inputstack{fashion4__arch4}{fashion4}{arch4} &
        \methodgrid{fashion4__arch4}{ours} &
        \begin{tabular}{@{}c@{\hspace{1pt}}c@{}}
            \includegraphics[width=\gridimg]{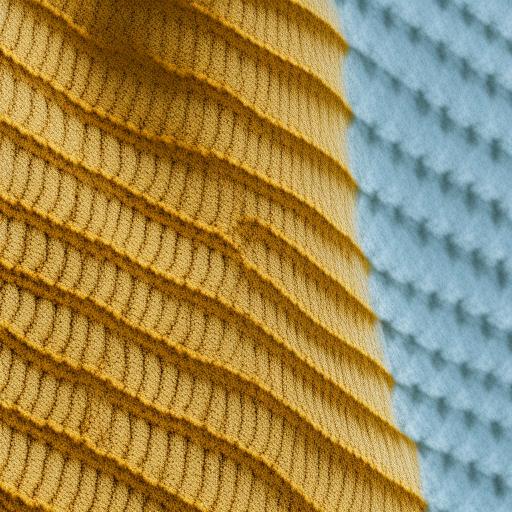} &
            \includegraphics[width=\gridimg]{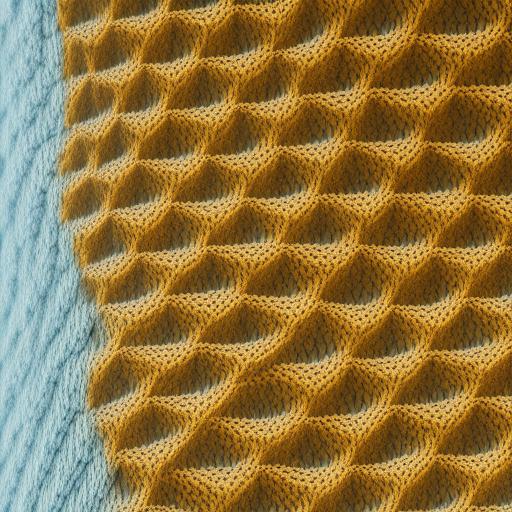} \\[-1pt]
            \includegraphics[width=\gridimg]{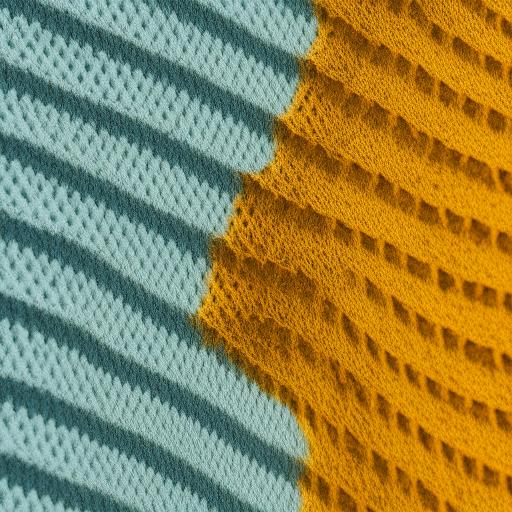} &
            \includegraphics[width=\gridimg]{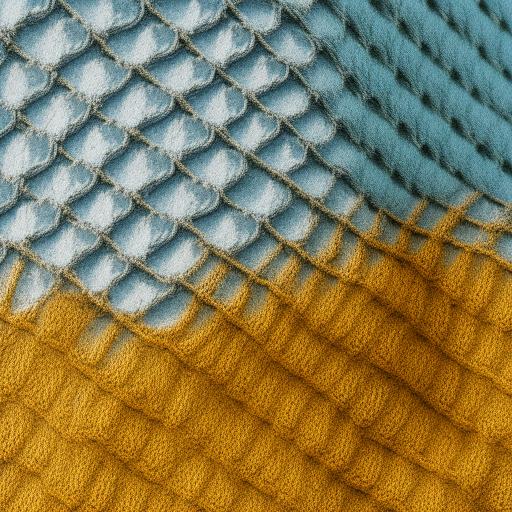}
        \end{tabular}
        \rowsep

        \inputstack{fashion5__food1}{fashion5}{food1} &
        \methodgrid{fashion5__food1}{ours} &
        \begin{tabular}{@{}c@{\hspace{1pt}}c@{}}
            \includegraphics[width=\gridimg]{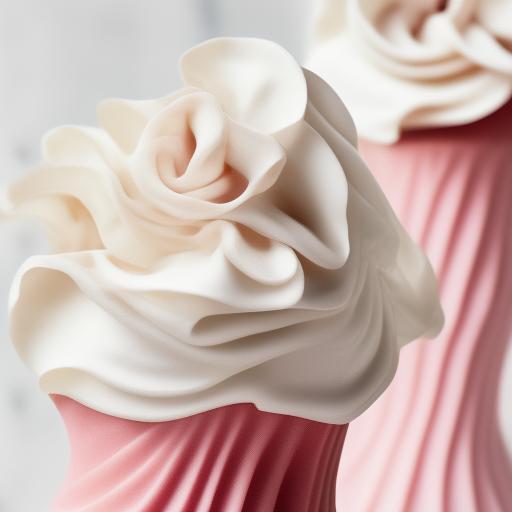} &
            \includegraphics[width=\gridimg]{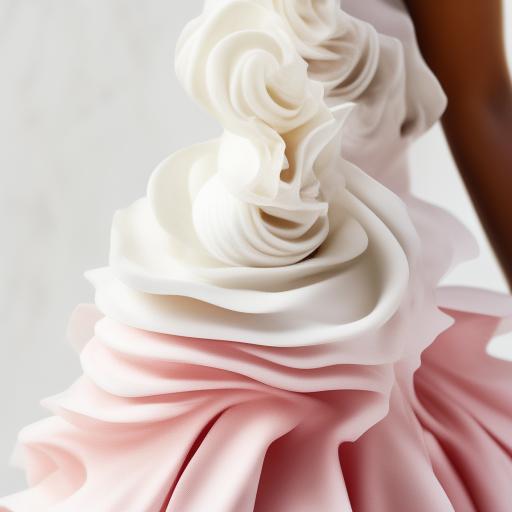} \\[-1pt]
            \includegraphics[width=\gridimg]{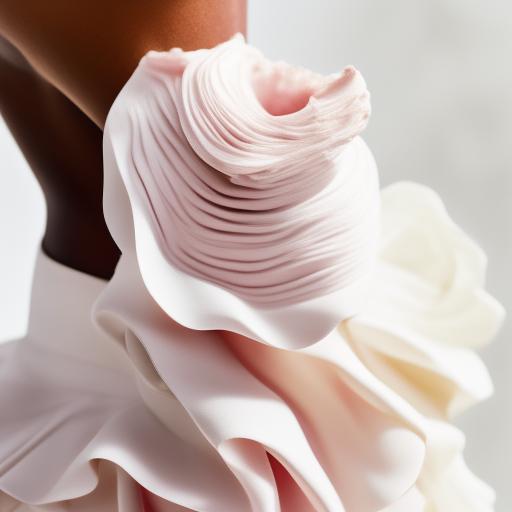} &
            \includegraphics[width=\gridimg]{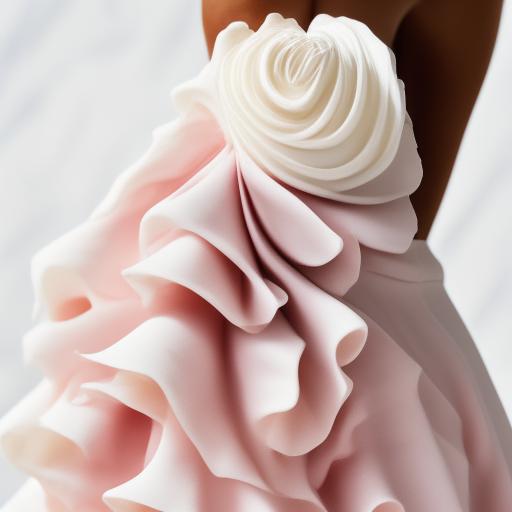}
        \end{tabular} &
        \inputstack{food1__nature2}{food1}{nature2} &
        \methodgrid{food1__nature2}{ours} &
        \begin{tabular}{@{}c@{\hspace{1pt}}c@{}}
            \includegraphics[width=\gridimg]{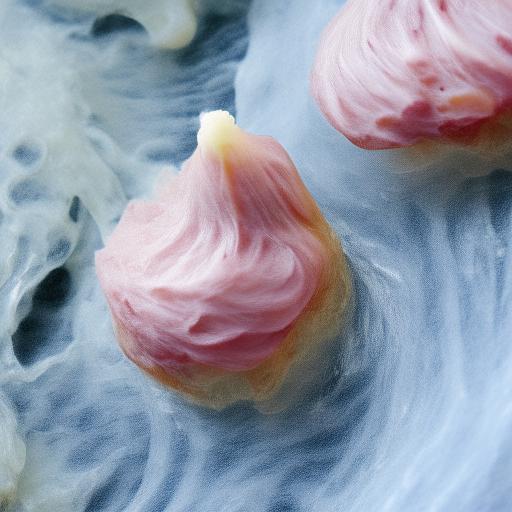} &
            \includegraphics[width=\gridimg]{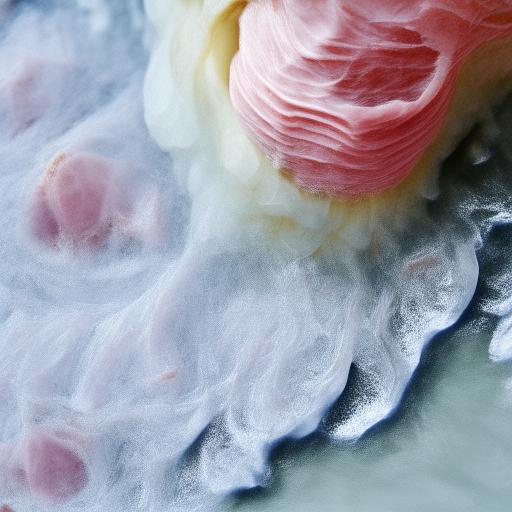} \\[-1pt]
            \includegraphics[width=\gridimg]{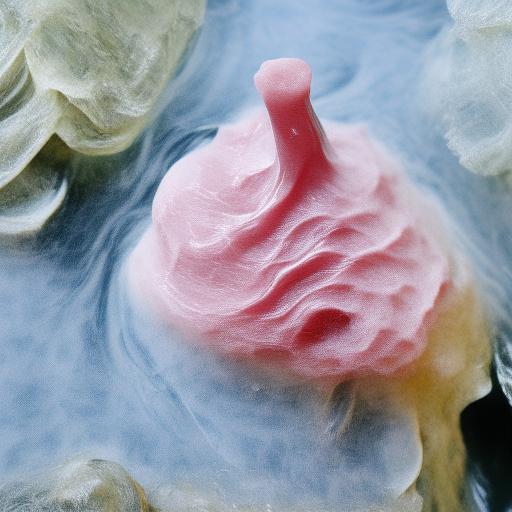} &
            \includegraphics[width=\gridimg]{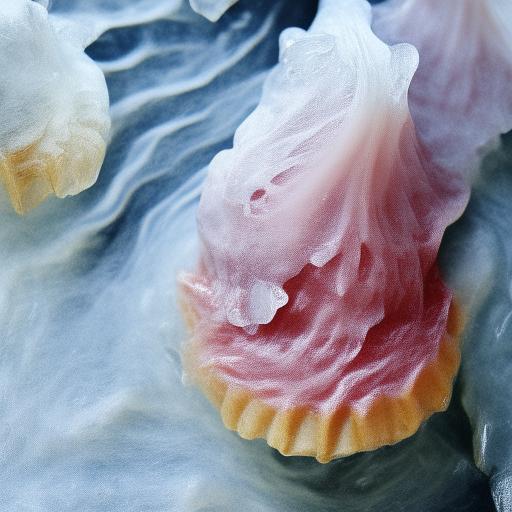}
        \end{tabular}
        \rowsep

        \inputstack{food4__sea9}{food4}{sea9} &
        \methodgrid{food4__sea9}{ours} &
        \begin{tabular}{@{}c@{\hspace{1pt}}c@{}}
            \includegraphics[width=\gridimg]{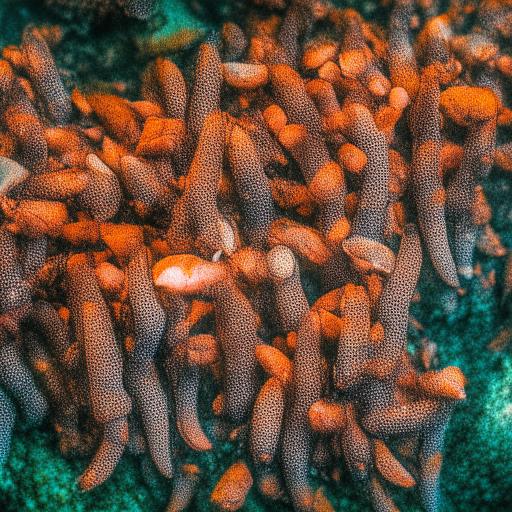} &
            \includegraphics[width=\gridimg]{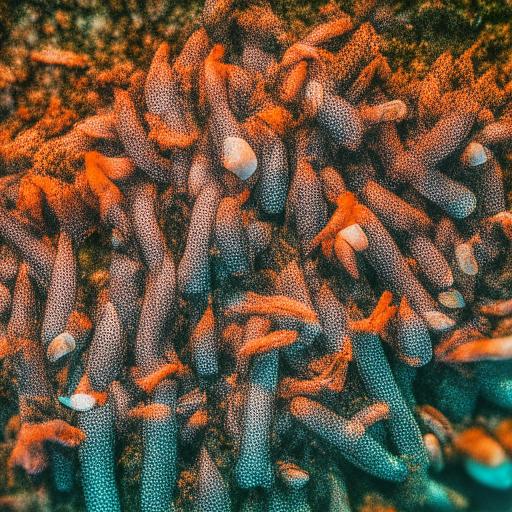} \\[-1pt]
            \includegraphics[width=\gridimg]{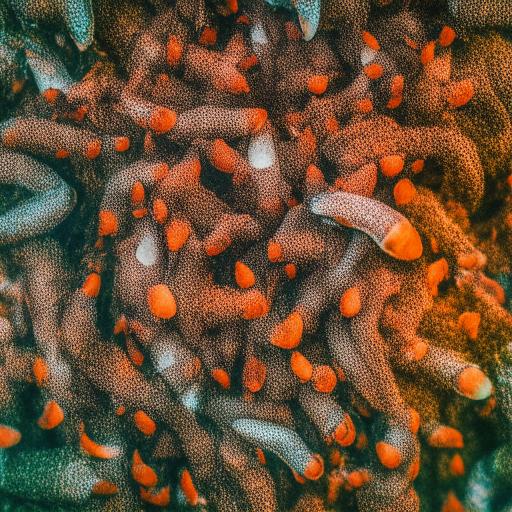} &
            \includegraphics[width=\gridimg]{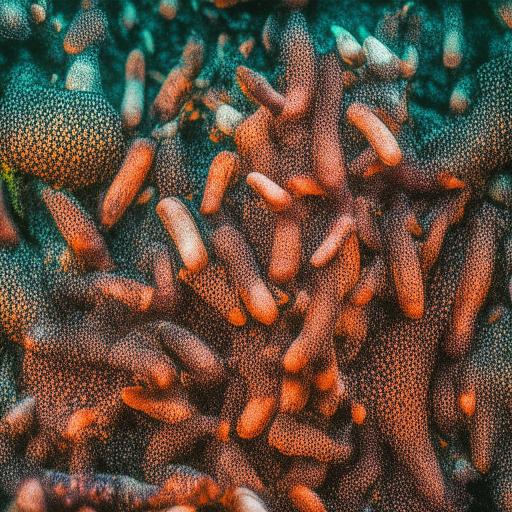}
        \end{tabular} &
        \inputstack{food6__nature8}{food6}{nature8} &
        \methodgrid{food6__nature8}{ours} &
        \begin{tabular}{@{}c@{\hspace{1pt}}c@{}}
            \includegraphics[width=\gridimg]{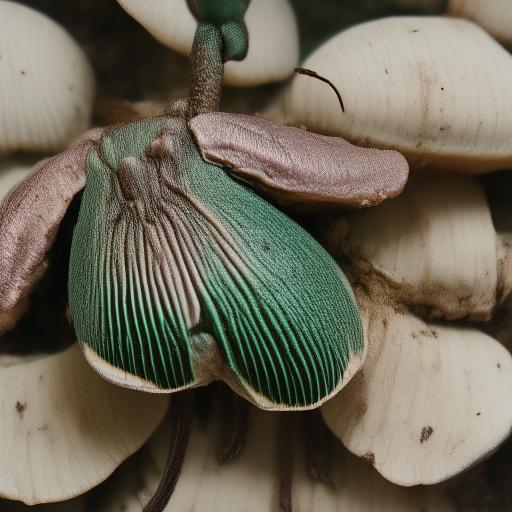} &
            \includegraphics[width=\gridimg]{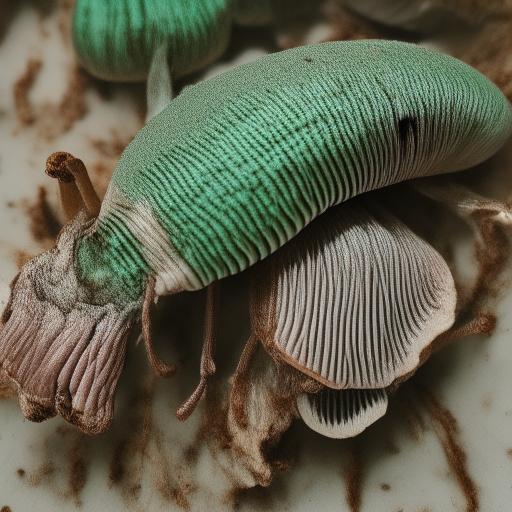} \\[-1pt]
            \includegraphics[width=\gridimg]{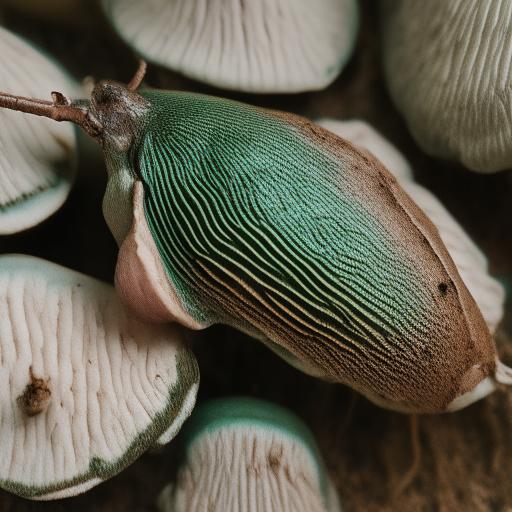} &
            \includegraphics[width=\gridimg]{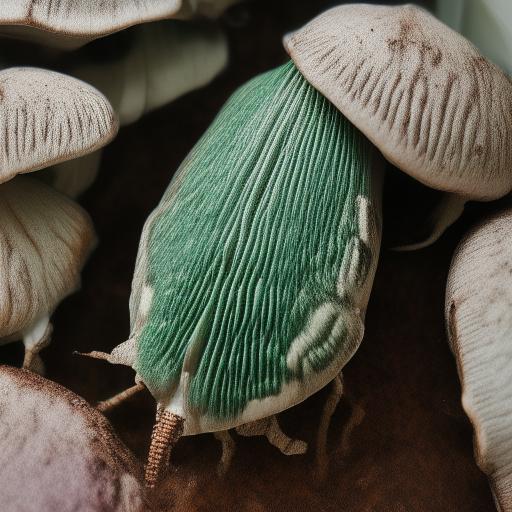}
        \end{tabular}
        \rowsep

        \inputstack{nature3__sea7}{nature3}{sea7} &
        \methodgrid{nature3__sea7}{ours} &
        \begin{tabular}{@{}c@{\hspace{1pt}}c@{}}
            \includegraphics[width=\gridimg]{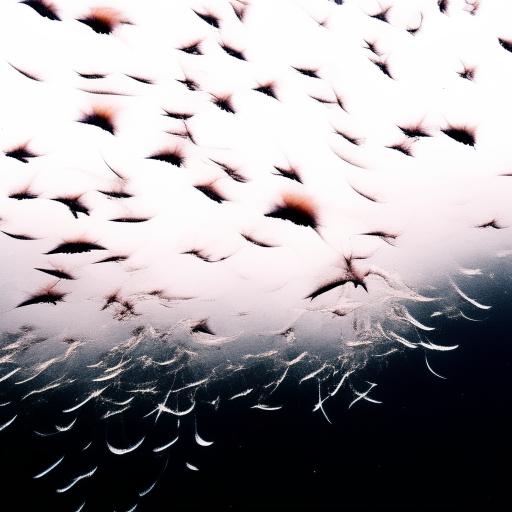} &
            \includegraphics[width=\gridimg]{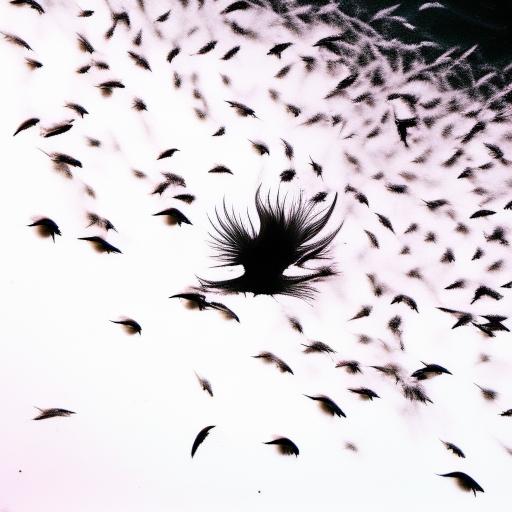} \\[-1pt]
            \includegraphics[width=\gridimg]{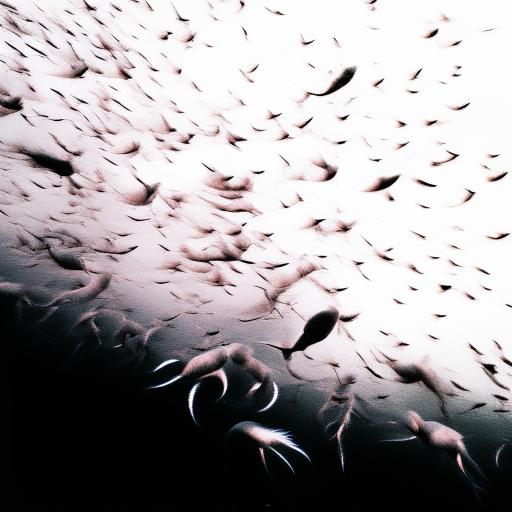} &
            \includegraphics[width=\gridimg]{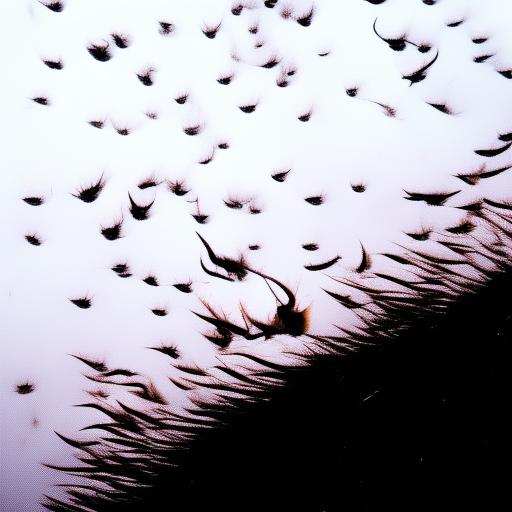}
        \end{tabular} &
        \inputstack{food4__other2}{food4}{other2} &
        \methodgrid{food4__other2}{ours} &
        \begin{tabular}{@{}c@{\hspace{1pt}}c@{}}
            \includegraphics[width=\gridimg]{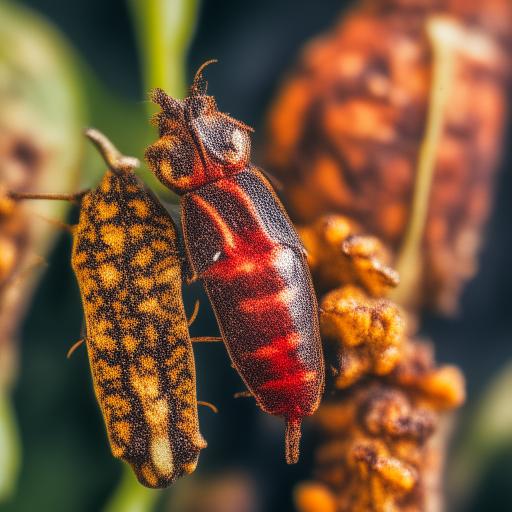} &
            \includegraphics[width=\gridimg]{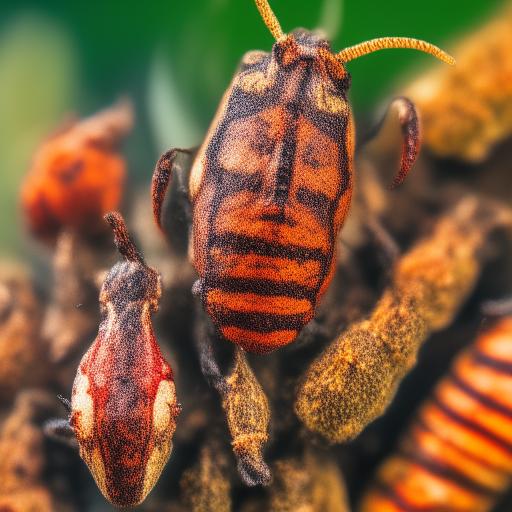} \\[-1pt]
            \includegraphics[width=\gridimg]{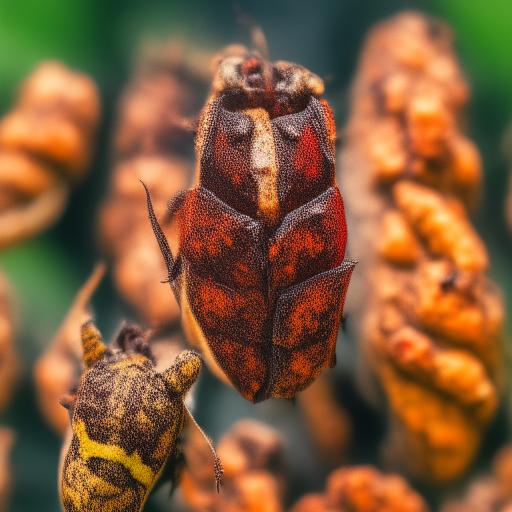} &
            \includegraphics[width=\gridimg]{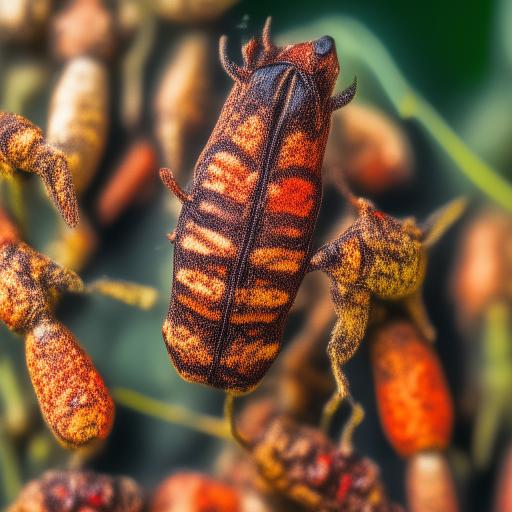}
        \end{tabular} \\

    \end{tabular}
    \caption{Comparison with CLIP space interpolation baseline. Each row shows two input pairs, with our method's outputs (middle) and the CLIP interpolation baseline (right) for each pair. Results are shown from 4 different seeds in a 2$\times$2 grid.}
    \label{fig:clip_interpolation_ablation}
\end{figure*}

As shown in \Cref{fig:clip_interpolation_ablation}, the CLIP interpolation baseline tends to produce blended or averaged results that are not always visually coherent.

\section{Description Complexity Evaluation}
As described in the main paper, we use description complexity as a proxy for measuring the non-triviality of visual combinations. We prompt Gemini 2.5 Flash~\cite{gemini25_2025} to describe how each output image could be reconstructed from its two source images, then measure the word count of the response.
We use the following prompt for evaluating our method, Nano Banana and Qwen-Image-2511:

\begin{quote}
\texttt{The first two images inspired the third. Describe briefly how you would recreate the output using only the two inputs.}

\texttt{Use short bullet points, not paragraphs. Maximum 5 bullets total, but you do not have to use them all.}

\texttt{Notes:}\\
\texttt{* Use ``*'' to denote bullets. Your answer should include only bullet points, no free text.}\\
\texttt{* Be concise when possible.}\\
\texttt{* If the output image is very similar to one of the inputs you can just say ``copy <image1>/<image2>'' accordingly.}\\
\texttt{* Examples of instructions you can use: ``place object from <image1> in the scene from <image2>'', ``copy <image1>'', ``copy <image2>'', ``use the object from <image1> and the texture from <image2>''. These are just examples, you can write your own instructions.}
\end{quote}

For Flux.1 Kontext, we pass the two inputs as a single image with the same grid structure used for inference, and slightly modify the prompt to describe this structure:
\begin{quote}
\texttt{The first image is 2x2 grid with two images in the top-left and bottom-right quadrants, which inspired the second image. Describe briefly how you would recreate the output using only the two images in the grid.}

\texttt{Use short bullet points, not paragraphs. Maximum 5 bullets total, but you do not have to use them all.}

\texttt{Notes:}\\
\texttt{* Use ``*'' to denote bullets. Your answer should include only bullet points, no free text.}\\
\texttt{* Be concise when possible.}\\
\texttt{* If the output image is very similar to one of the inputs you can just say ``copy <image1>/<image2>'' accordingly.}\\
\texttt{* Examples of instructions you can use: ``place object from <image1> in the scene from <image2>'', ``copy <image1>'', ``copy <image2>'', ``copy entire grid'', ``use the object from <image1> and the texture from <image2>''. These are just examples, you can write your own instructions.}
\end{quote}

\begin{figure*}[t]
\centering
\setlength{\tabcolsep}{1pt}
\renewcommand{\arraystretch}{1.0}

\begin{tabular}{c c c c c c c c c}
\toprule
Inputs &
\multicolumn{2}{c}{Flux.1 Kontext} &
\multicolumn{2}{c}{Qwen-Image-2511} &
\multicolumn{2}{c}{Nano Banana} &
\multicolumn{2}{c}{Ours} \\
\midrule \noalign{\vskip-8pt}

\raisebox{-\height}{\begin{tabular}[t]{@{}c@{}}
\includegraphics[width=0.065\linewidth,height=0.065\linewidth]{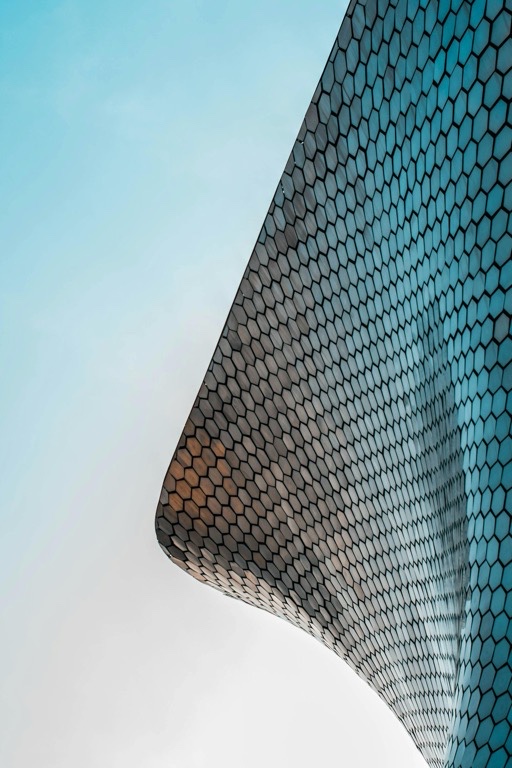} \\
\includegraphics[width=0.065\linewidth,height=0.065\linewidth]{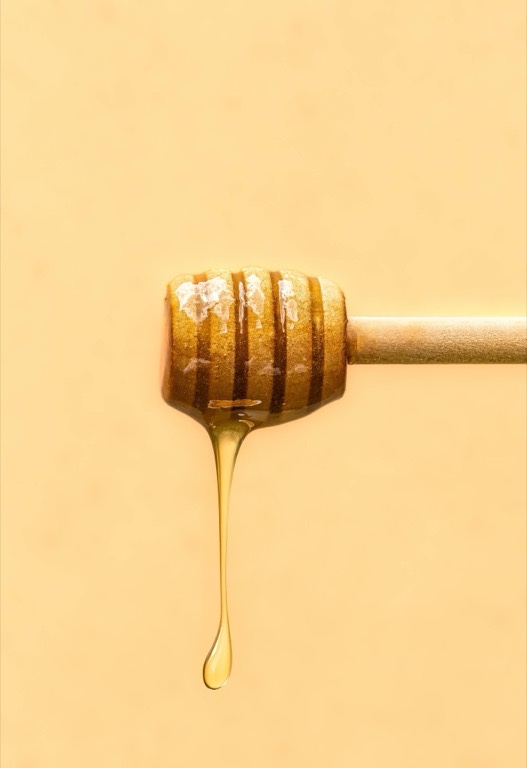}
\end{tabular}}
&
\raisebox{-\height}{\begin{tabular}[t]{@{}c@{}}
\includegraphics[width=0.1\linewidth,height=0.1\linewidth]{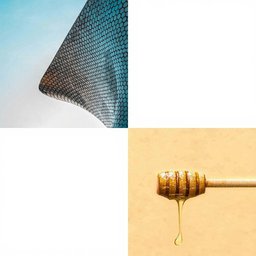} \\[-4pt]
\scriptsize(3 words)
\end{tabular}}
&
\raisebox{-\height}{\parbox[t]{0.1\linewidth}{\raggedright\scriptsize
\textbullet~copy entire grid
}}
&
\raisebox{-\height}{\begin{tabular}[t]{@{}c@{}}
\includegraphics[width=0.1\linewidth,height=0.1\linewidth]{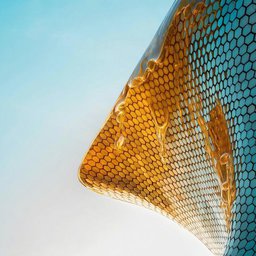} \\[-4pt]
\scriptsize(44 words)
\end{tabular}}
&
\raisebox{-\height}{\parbox[t]{0.1\linewidth}{\raggedright\scriptsize
\textbullet~Use the building from image 1 as the base.
\textbullet~Extract the dripping honey from image 2.
\textbullet~Apply the extracted honey to the hexagonal surface of the building in image 1.
\textbullet~Adjust the honey to flow and coat the building's texture.
}}
&
\raisebox{-\height}{\begin{tabular}[t]{@{}c@{}}
\includegraphics[width=0.1\linewidth,height=0.1\linewidth]{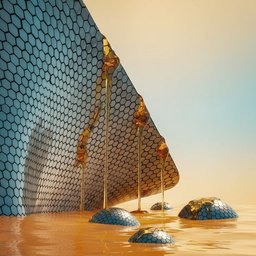} \\[-4pt]
\scriptsize(73 words)
\end{tabular}}
&
\raisebox{-\height}{\parbox[t]{0.1\linewidth}{\raggedright\scriptsize
\textbullet~Use the architectural structure and hexagonal pattern from image 1.
\textbullet~Extract the honey color and dripping texture from image 2.
\textbullet~Submerge parts of the hexagonal structure from image 1 into a body of the extracted honey liquid.
\textbullet~Create new elements mimicking the honey drips from image 2 emerging from the liquid.
}}
&
\raisebox{-\height}{\begin{tabular}[t]{@{}c@{}}
\includegraphics[width=0.1\linewidth,height=0.1\linewidth]{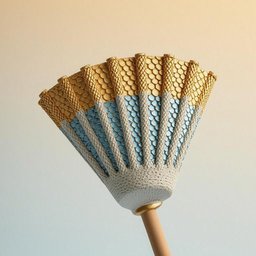} \\[-4pt]
\scriptsize(76 words)
\end{tabular}}
&
\raisebox{-\height}{\parbox[t]{0.23\linewidth}{\raggedright\scriptsize
\textbullet~Form a fan-like object with radiating segments.
\textbullet~Apply the hexagonal pattern from image 1, coloring sections with blue from image 1 and golden from image 2.
\textbullet~For other segments, apply a ribbed texture inspired by the honey dipper's grooves (image 2) and colored light grey from image 1.
\textbullet~Attach the wooden handle from image 2 to the base of the object.
}}
\\[4pt]
\midrule \noalign{\vskip-8pt}

\raisebox{-\height}{\begin{tabular}[t]{@{}c@{}}
\includegraphics[width=0.065\linewidth,height=0.065\linewidth]{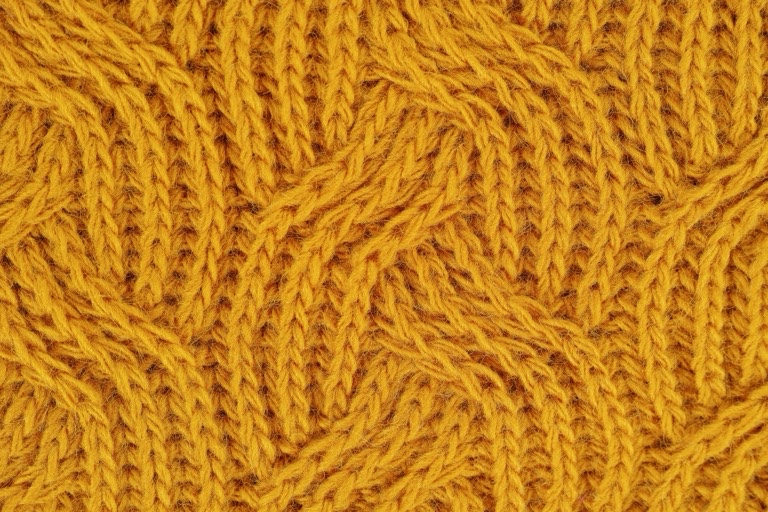} \\
\includegraphics[width=0.065\linewidth,height=0.065\linewidth]{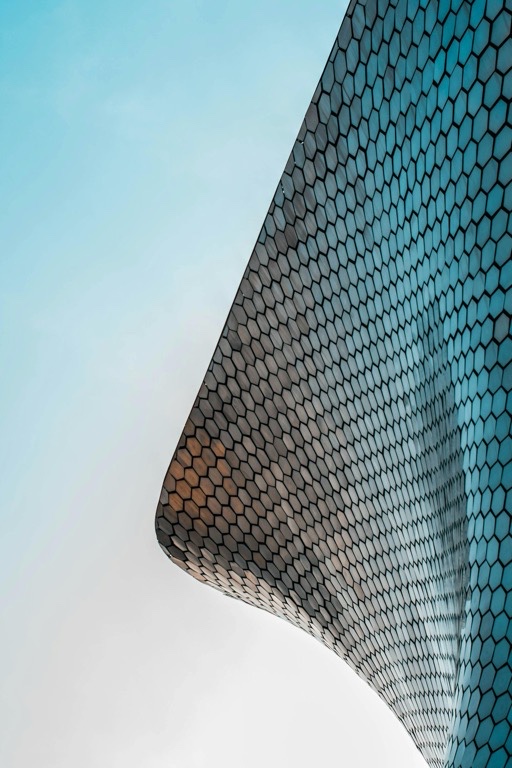}
\end{tabular}}
&
\raisebox{-\height}{\begin{tabular}[t]{@{}c@{}}
\includegraphics[width=0.1\linewidth,height=0.1\linewidth]{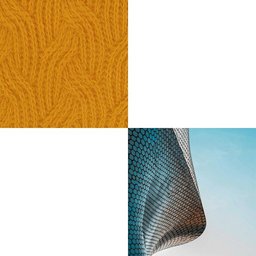} \\[-4pt]
\scriptsize(29 words)
\end{tabular}}
&
\raisebox{-\height}{\parbox[t]{0.1\linewidth}{\raggedright\scriptsize
\textbullet~Soften the knitted texture of the top-left image to a fuzzy appearance.
\textbullet~Emphasize the hexagonal pattern of the bottom-right image as a stark black overlay on the building's form.
}}
&
\raisebox{-\height}{\begin{tabular}[t]{@{}c@{}}
\includegraphics[width=0.1\linewidth,height=0.1\linewidth]{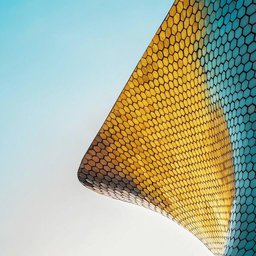} \\[-4pt]
\scriptsize(26 words)
\end{tabular}}
&
\raisebox{-\height}{\parbox[t]{0.1\linewidth}{\raggedright\scriptsize
\textbullet~Use the architectural structure and sky from image 2.
\textbullet~Recolor the golden-hued section of the building's facade with the yellow color from image 1.
}}
&
\raisebox{-\height}{\begin{tabular}[t]{@{}c@{}}
\includegraphics[width=0.1\linewidth,height=0.1\linewidth]{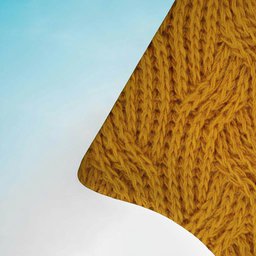} \\[-4pt]
\scriptsize(43 words)
\end{tabular}}
&
\raisebox{-\height}{\parbox[t]{0.1\linewidth}{\raggedright\scriptsize
\textbullet~Extract the background sky and gradient from image 2.
\textbullet~Extract the shape of the building from image 2.
\textbullet~Fill the extracted building shape with the texture from image 1.
\textbullet~Place the filled shape onto the background from image 2.
}}
&
\raisebox{-\height}{\begin{tabular}[t]{@{}c@{}}
\includegraphics[width=0.1\linewidth,height=0.1\linewidth]{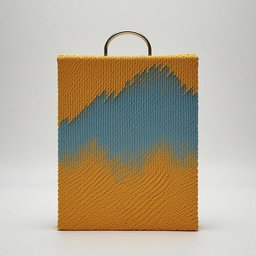} \\[-4pt]
\scriptsize(73 words)
\end{tabular}}
&
\raisebox{-\height}{\parbox[t]{0.23\linewidth}{\raggedright\scriptsize
\textbullet~Create a rectangular object with a handle.
\textbullet~Apply the knitted texture from image 1 to the entire object.
\textbullet~Color the top and bottom sections of the object using the yellow from image 1.
\textbullet~Color the middle section using the blue/grey gradient and tones from the building in image 2.
\textbullet~Form the boundary between the colored sections with an angular pattern inspired by the hexagonal structure in image 2.
}}
\\[4pt]
\midrule \noalign{\vskip-8pt}

\raisebox{-\height}{\begin{tabular}[t]{@{}c@{}}
\includegraphics[width=0.065\linewidth,height=0.065\linewidth]{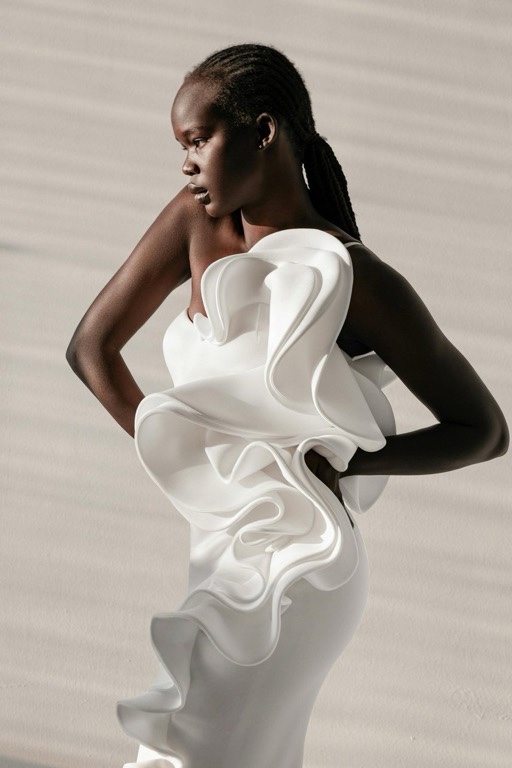} \\
\includegraphics[width=0.065\linewidth,height=0.065\linewidth]{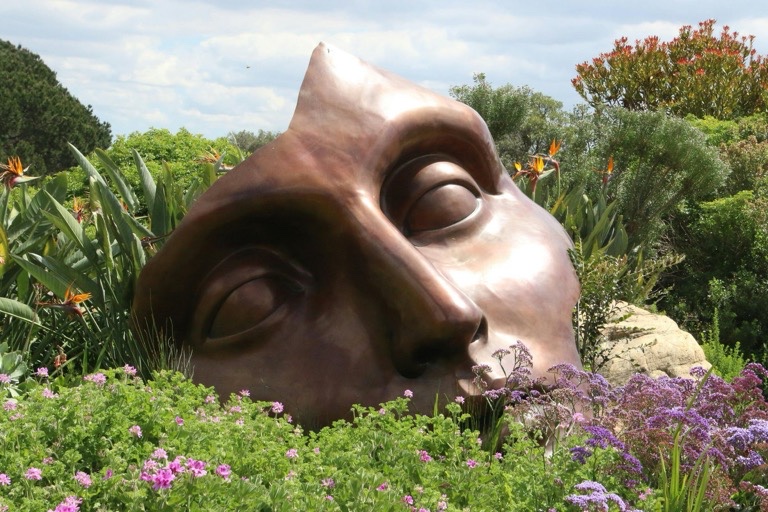}
\end{tabular}}
&
\raisebox{-\height}{\begin{tabular}[t]{@{}c@{}}
\includegraphics[width=0.1\linewidth,height=0.1\linewidth]{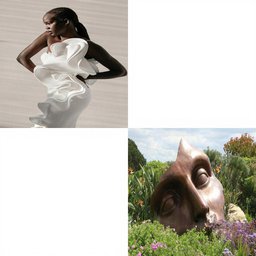} \\[-4pt]
\scriptsize(29 words)
\end{tabular}}
&
\raisebox{-\height}{\parbox[t]{0.1\linewidth}{\raggedright\scriptsize
\textbullet~Isolate the woman from the top-left image.
\textbullet~Integrate the isolated woman into the background of the bottom-right image.
\textbullet~Adjust the woman's scale and lighting to match the garden scene.
}}
&
\raisebox{-\height}{\begin{tabular}[t]{@{}c@{}}
\includegraphics[width=0.1\linewidth,height=0.1\linewidth]{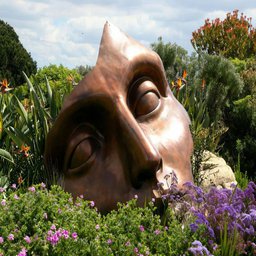} \\[-4pt]
\scriptsize(3 words)
\end{tabular}}
&
\raisebox{-\height}{\parbox[t]{0.1\linewidth}{\raggedright\scriptsize
\textbullet~copy $\langle$image2$\rangle$
}}
&
\raisebox{-\height}{\begin{tabular}[t]{@{}c@{}}
\includegraphics[width=0.1\linewidth,height=0.1\linewidth]{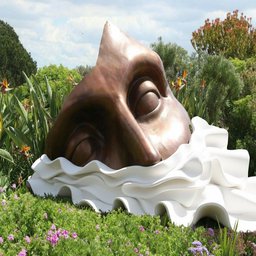} \\[-4pt]
\scriptsize(32 words)
\end{tabular}}
&
\raisebox{-\height}{\parbox[t]{0.1\linewidth}{\raggedright\scriptsize
\textbullet~Use the scene from image2.
\textbullet~Extract the white ruffled material from the dress in image1.
\textbullet~Place the extracted ruffled material around the base of the face sculpture in image2.
}}
&
\raisebox{-\height}{\begin{tabular}[t]{@{}c@{}}
\includegraphics[width=0.1\linewidth,height=0.1\linewidth]{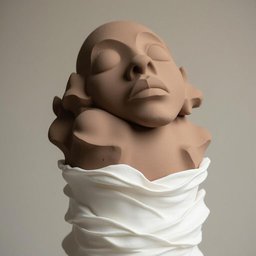} \\[-4pt]
\scriptsize(66 words)
\end{tabular}}
&
\raisebox{-\height}{\parbox[t]{0.23\linewidth}{\raggedright\scriptsize
\textbullet~Extract the sculptural head form and features from image 2.
\textbullet~Apply the matte, reddish-brown/terracotta color and texture from the sculpture in image 2 to the head and upper body.
\textbullet~Wrap the lower part of the bust with the white, ruffled, sculptural fabric texture from the dress in image 1.
\textbullet~Present the sculpture against a plain, light, minimalist background, similar to image 1.
}}
\\[4pt]
\midrule \noalign{\vskip-8pt}

\raisebox{-\height}{\begin{tabular}[t]{@{}c@{}}
\includegraphics[width=0.065\linewidth,height=0.065\linewidth]{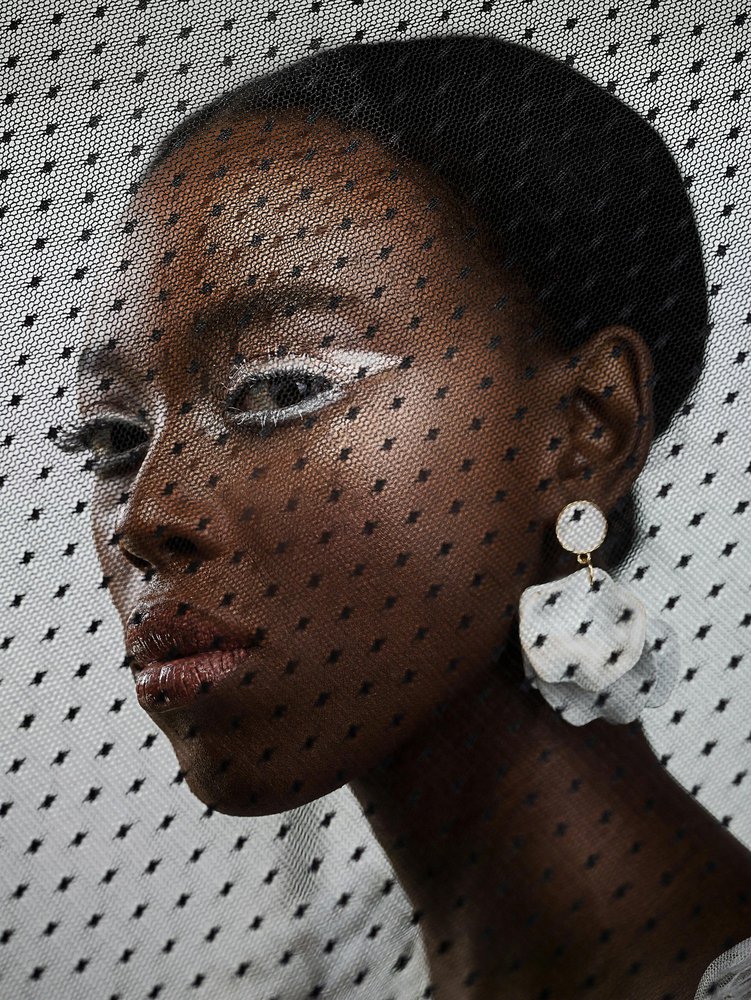} \\
\includegraphics[width=0.065\linewidth,height=0.065\linewidth]{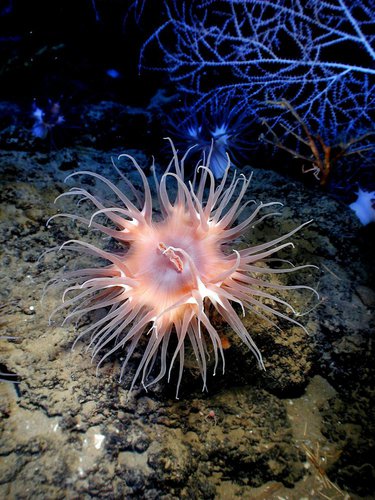}
\end{tabular}}
&
\raisebox{-\height}{\begin{tabular}[t]{@{}c@{}}
\includegraphics[width=0.1\linewidth,height=0.1\linewidth]{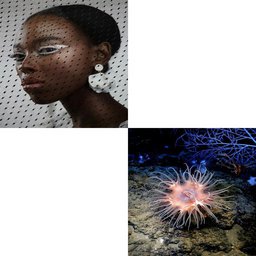} \\[-4pt]
\scriptsize(3 words)
\end{tabular}}
&
\raisebox{-\height}{\parbox[t]{0.1\linewidth}{\raggedright\scriptsize
\textbullet~copy entire grid
}}
&
\raisebox{-\height}{\begin{tabular}[t]{@{}c@{}}
\includegraphics[width=0.1\linewidth,height=0.1\linewidth]{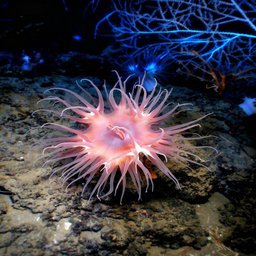} \\[-4pt]
\scriptsize(15 words)
\end{tabular}}
&
\raisebox{-\height}{\parbox[t]{0.1\linewidth}{\raggedright\scriptsize
\textbullet~Copy image 2.
\textbullet~Overlay the dotted texture from image 1 onto image 2.
}}
&
\raisebox{-\height}{\begin{tabular}[t]{@{}c@{}}
\includegraphics[width=0.1\linewidth,height=0.1\linewidth]{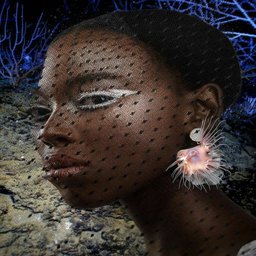} \\[-4pt]
\scriptsize(47 words)
\end{tabular}}
&
\raisebox{-\height}{\parbox[t]{0.1\linewidth}{\raggedright\scriptsize
\textbullet~Extract the woman's face and veil from image 1.
\textbullet~Replace the background with the underwater scene from image 2.
\textbullet~Replace the earring from image 1 with the sea anemone from image 2.
\textbullet~Position the sea anemone as an earring on the woman's ear.
}}
&
\raisebox{-\height}{\begin{tabular}[t]{@{}c@{}}
\includegraphics[width=0.1\linewidth,height=0.1\linewidth]{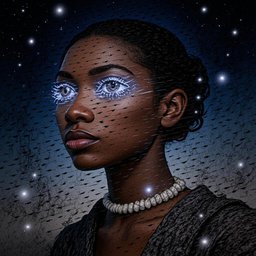} \\[-4pt]
\scriptsize(86 words)
\end{tabular}}
&
\raisebox{-\height}{\parbox[t]{0.23\linewidth}{\raggedright\scriptsize
\textbullet~Take the woman's face from image1.
\textbullet~Extract the dotted pattern from the veil in image1, stylize these dots into arrow shapes, and overlay them on the face and in the background.
\textbullet~Enhance the white eyeliner from image1 to create a glowing effect for the eyes.
\textbullet~Integrate the translucent, tentacle-like structures and luminosity from the sea anemone in image2 into the glowing details of the eyes.
\textbullet~Adopt the dark, atmospheric color palette and subtle light sources from image2 for the overall background.
}}
\\

\bottomrule
\end{tabular}

\caption{Description complexity comparison (1/2). Each row shows two input images and outputs from four methods. Below each output, we show the VLM-generated instructions describing how to recreate it, along with the word count. Our method produces outputs that require significantly more words to describe, indicating more complex and non-trivial combinations.}
\label{fig:caption_length_comparison}
\end{figure*}

\begin{figure*}[t]
\centering
\setlength{\tabcolsep}{1pt}
\renewcommand{\arraystretch}{1.0}

\begin{tabular}{c c c c c c c c c}
\toprule
Inputs &
\multicolumn{2}{c}{Flux.1 Kontext} &
\multicolumn{2}{c}{Qwen-Image-2511} &
\multicolumn{2}{c}{Nano Banana} &
\multicolumn{2}{c}{Ours} \\
\midrule \noalign{\vskip-8pt}

\raisebox{-\height}{\begin{tabular}[t]{@{}c@{}}
\includegraphics[width=0.065\linewidth,height=0.065\linewidth]{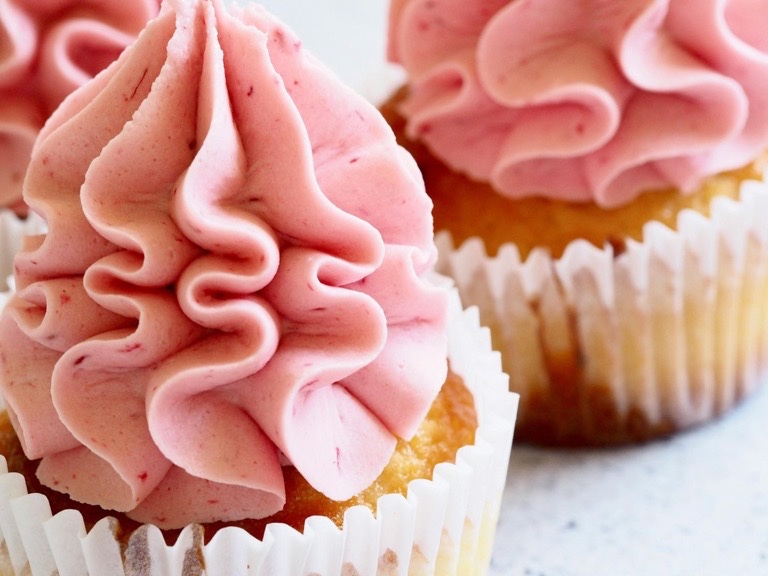} \\
\includegraphics[width=0.065\linewidth,height=0.065\linewidth]{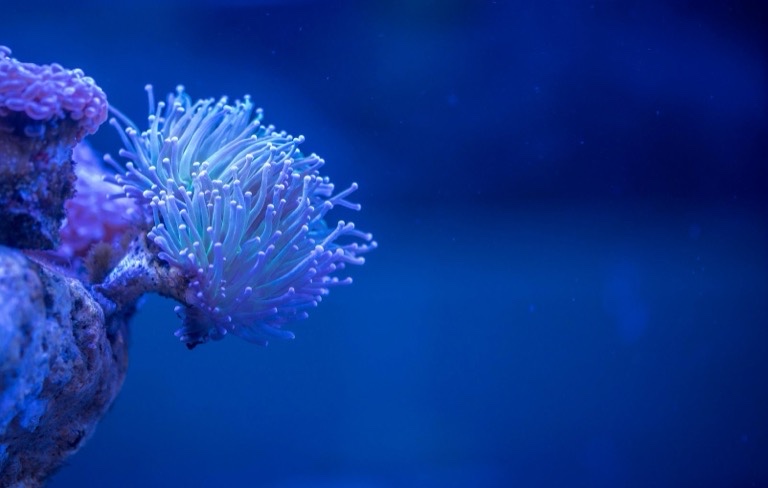}
\end{tabular}}
&
\raisebox{-\height}{\begin{tabular}[t]{@{}c@{}}
\includegraphics[width=0.1\linewidth,height=0.1\linewidth]{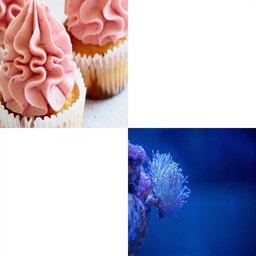} \\[-4pt]
\scriptsize(3 words)
\end{tabular}}
&
\raisebox{-\height}{\parbox[t]{0.1\linewidth}{\raggedright\scriptsize
\textbullet~copy entire grid
}}
&
\raisebox{-\height}{\begin{tabular}[t]{@{}c@{}}
\includegraphics[width=0.1\linewidth,height=0.1\linewidth]{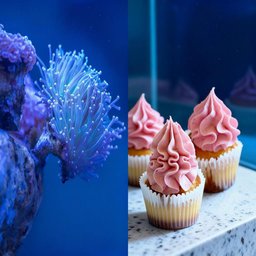} \\[-4pt]
\scriptsize(28 words)
\end{tabular}}
&
\raisebox{-\height}{\parbox[t]{0.1\linewidth}{\raggedright\scriptsize
\textbullet~Create a vertically split image.
\textbullet~Place image2 on the left side.
\textbullet~Place scaled-down cupcakes from image1 onto a white speckled counter on the right side.
}}
&
\raisebox{-\height}{\begin{tabular}[t]{@{}c@{}}
\includegraphics[width=0.1\linewidth,height=0.1\linewidth]{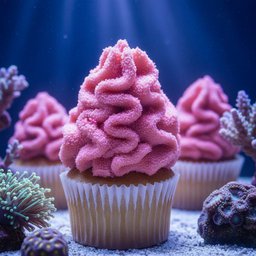} \\[-4pt]
\scriptsize(46 words)
\end{tabular}}
&
\raisebox{-\height}{\parbox[t]{0.1\linewidth}{\raggedright\scriptsize
\textbullet~Place the cupcakes from image1 into the underwater scene from image2.
\textbullet~Apply the intricate texture and shape of the coral/anemone from image2 to the pink frosting of the cupcakes.
\textbullet~Incorporate other coral elements from image2 into the foreground and background of the scene.
}}
&
\raisebox{-\height}{\begin{tabular}[t]{@{}c@{}}
\includegraphics[width=0.1\linewidth,height=0.1\linewidth]{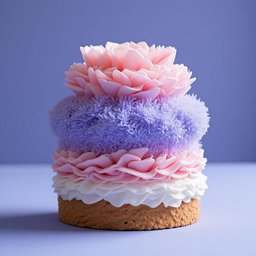} \\[-4pt]
\scriptsize(83 words)
\end{tabular}}
&
\raisebox{-\height}{\parbox[t]{0.23\linewidth}{\raggedright\scriptsize
\textbullet~Use the texture and color of the frosting from $\langle$image1$\rangle$ to create pink flower petals and ruffled layers.
\textbullet~Extract the tentacle texture and overall form of the coral from $\langle$image2$\rangle$, recoloring it purple, for the fluffy middle layer.
\textbullet~Apply the texture of the frosting from $\langle$image1$\rangle$ to create a white ruffled layer.
\textbullet~Extract the color and textured surface from the cupcake base of $\langle$image1$\rangle$ for the bottom layer.
\textbullet~Use the gradient background style from $\langle$image2$\rangle$, recolored to purple.
}}
\\[4pt]
\midrule \noalign{\vskip-8pt}

\raisebox{-\height}{\begin{tabular}[t]{@{}c@{}}
\includegraphics[width=0.065\linewidth,height=0.065\linewidth]{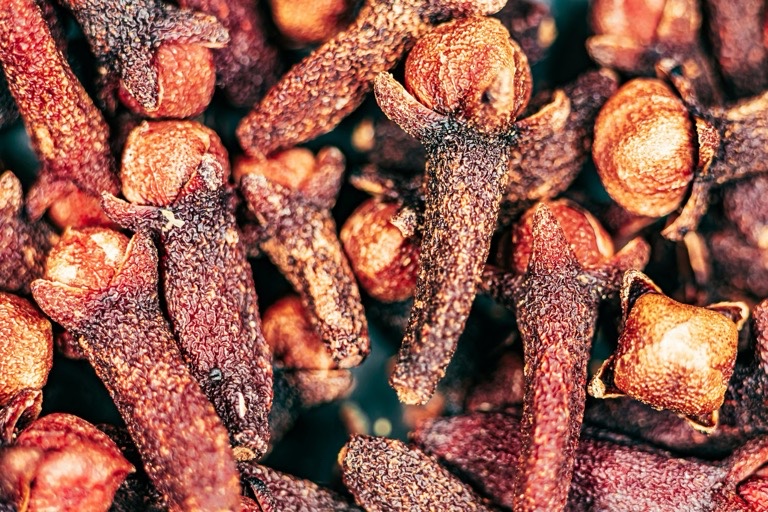} \\
\includegraphics[width=0.065\linewidth,height=0.065\linewidth]{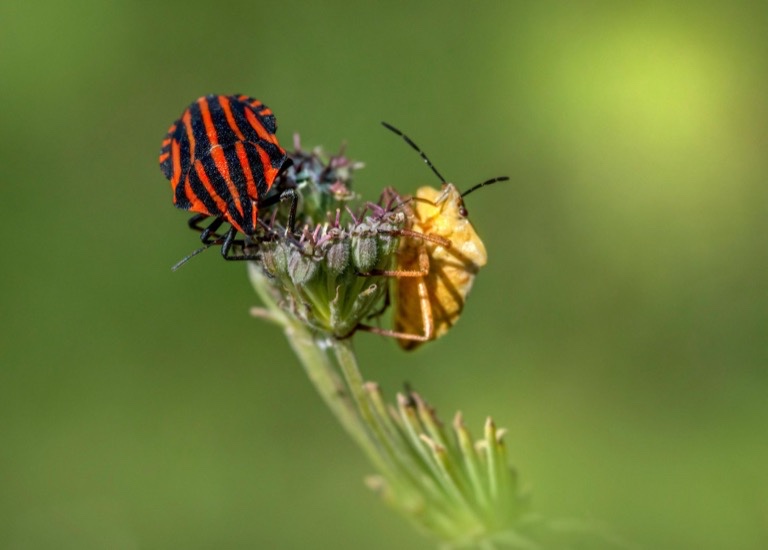}
\end{tabular}}
&
\raisebox{-\height}{\begin{tabular}[t]{@{}c@{}}
\includegraphics[width=0.1\linewidth,height=0.1\linewidth]{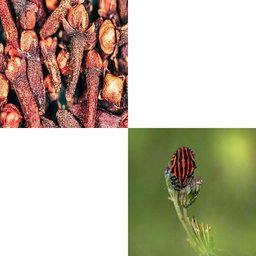} \\[-4pt]
\scriptsize(3 words)
\end{tabular}}
&
\raisebox{-\height}{\parbox[t]{0.1\linewidth}{\raggedright\scriptsize
\textbullet~copy bottom-right image
}}
&
\raisebox{-\height}{\begin{tabular}[t]{@{}c@{}}
\includegraphics[width=0.1\linewidth,height=0.1\linewidth]{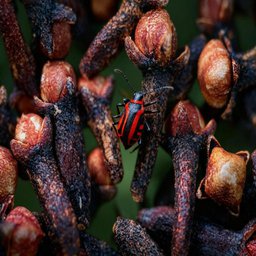} \\[-4pt]
\scriptsize(30 words)
\end{tabular}}
&
\raisebox{-\height}{\parbox[t]{0.1\linewidth}{\raggedright\scriptsize
\textbullet~Place the striped bug from image 2 onto the cloves from image 1.
\textbullet~Replace the background of the clove composition with a blurred green, inspired by image 2.
}}
&
\raisebox{-\height}{\begin{tabular}[t]{@{}c@{}}
\includegraphics[width=0.1\linewidth,height=0.1\linewidth]{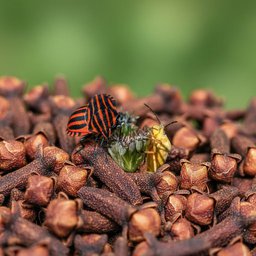} \\[-4pt]
\scriptsize(41 words)
\end{tabular}}
&
\raisebox{-\height}{\parbox[t]{0.1\linewidth}{\raggedright\scriptsize
\textbullet~Isolate the bugs and the plant stem from image 2.
\textbullet~Use the cloves from image 1 as the new foreground/surface.
\textbullet~Place the isolated bugs and plant stem onto the cloves, retaining the blurred green background from image 2.
}}
&
\raisebox{-\height}{\begin{tabular}[t]{@{}c@{}}
\includegraphics[width=0.1\linewidth,height=0.1\linewidth]{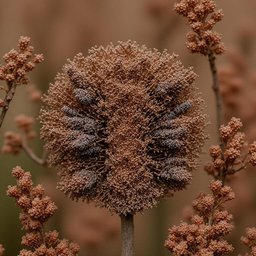} \\[-4pt]
\scriptsize(83 words)
\end{tabular}}
&
\raisebox{-\height}{\parbox[t]{0.23\linewidth}{\raggedright\scriptsize
\textbullet~Extract and multiply individual clove buds and stalks from Image 1.
\textbullet~Arrange these elements to form a dense spherical structure for the main object and smaller clusters for background details.
\textbullet~Apply a strong desaturation and color shift to achieve a muted brown/tan palette.
\textbullet~Integrate darker, hue-shifted clove elements from Image 1 into the main sphere for visual accents.
\textbullet~Apply a shallow depth of field blur to the background and surrounding elements, similar to the effect in Image 2.
}}
\\[4pt]
\midrule \noalign{\vskip-8pt}

\raisebox{-\height}{\begin{tabular}[t]{@{}c@{}}
\includegraphics[width=0.065\linewidth,height=0.065\linewidth]{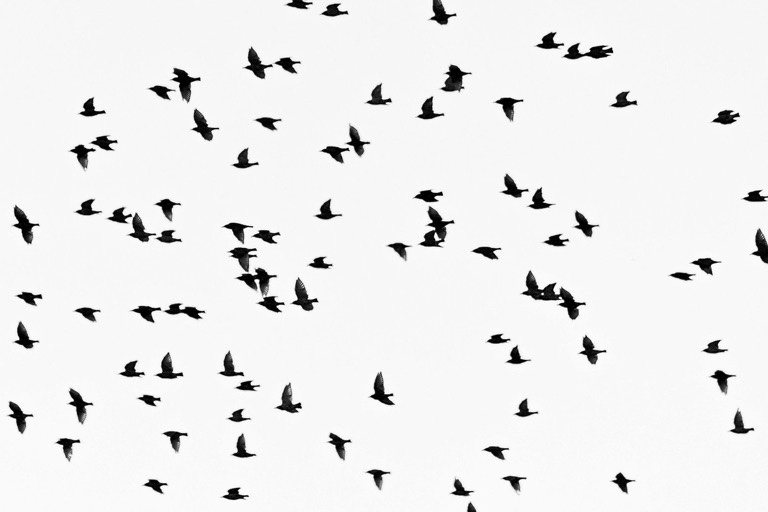} \\
\includegraphics[width=0.065\linewidth,height=0.065\linewidth]{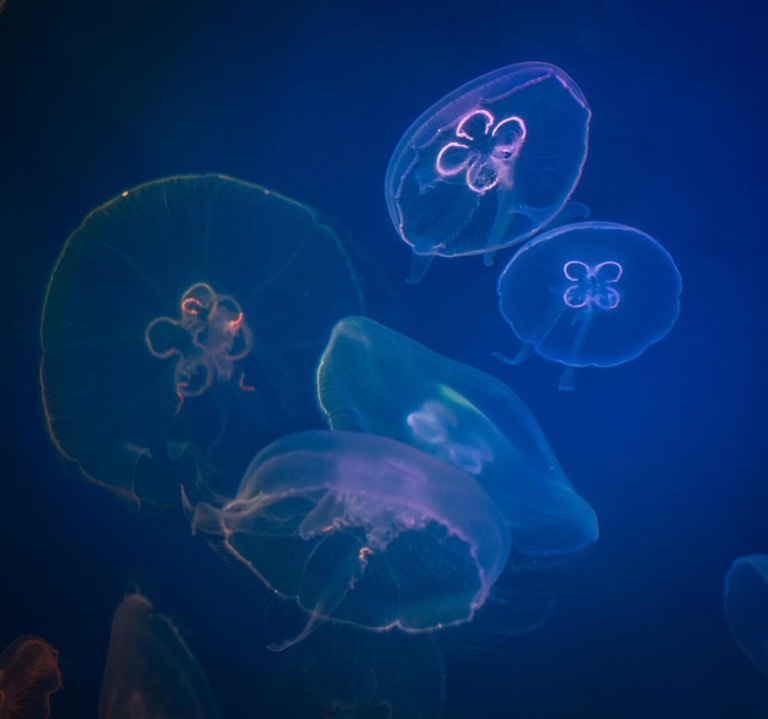}
\end{tabular}}
&
\raisebox{-\height}{\begin{tabular}[t]{@{}c@{}}
\includegraphics[width=0.1\linewidth,height=0.1\linewidth]{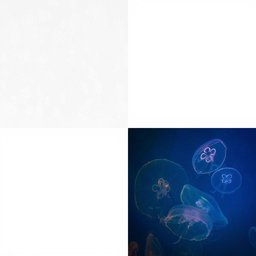} \\[-4pt]
\scriptsize(10 words)
\end{tabular}}
&
\raisebox{-\height}{\parbox[t]{0.1\linewidth}{\raggedright\scriptsize
\textbullet~Replace the birds image in the top-left quadrant with white.
}}
&
\raisebox{-\height}{\begin{tabular}[t]{@{}c@{}}
\includegraphics[width=0.1\linewidth,height=0.1\linewidth]{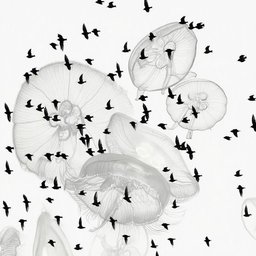} \\[-4pt]
\scriptsize(36 words)
\end{tabular}}
&
\raisebox{-\height}{\parbox[t]{0.1\linewidth}{\raggedright\scriptsize
\textbullet~Extract jellyfish from image2.
\textbullet~Convert extracted jellyfish to grayscale outlines/sketches.
\textbullet~Place the outlined jellyfish on a white background.
\textbullet~Extract birds from image1.
\textbullet~Overlay birds from image1 onto and around the jellyfish.
}}
&
\raisebox{-\height}{\begin{tabular}[t]{@{}c@{}}
\includegraphics[width=0.1\linewidth,height=0.1\linewidth]{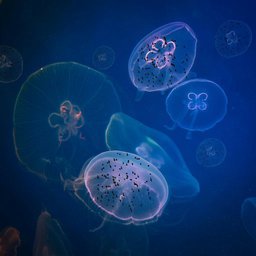} \\[-4pt]
\scriptsize(26 words)
\end{tabular}}
&
\raisebox{-\height}{\parbox[t]{0.1\linewidth}{\raggedright\scriptsize
\textbullet~Copy image 2.
\textbullet~Extract birds from image 1.
\textbullet~Place and scale down multiple groups of extracted birds onto the jellyfish in image 2.
}}
&
\raisebox{-\height}{\begin{tabular}[t]{@{}c@{}}
\includegraphics[width=0.1\linewidth,height=0.1\linewidth]{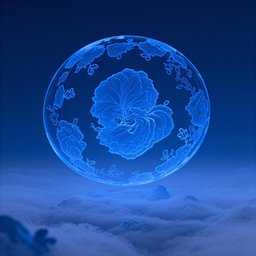} \\[-4pt]
\scriptsize(88 words)
\end{tabular}}
&
\raisebox{-\height}{\parbox[t]{0.23\linewidth}{\raggedright\scriptsize
\textbullet~Establish a deep blue background and luminous glow, inspired by the environment of Image 2.
\textbullet~Create a translucent spherical object as the central focal point.
\textbullet~Generate intricate floral patterns by abstracting and stylizing the internal structures of the jellyfish from Image 2.
\textbullet~Apply the sharp definition and silhouette-like quality from Image 1 to these patterns, rendering them with the glowing blue color and translucency of Image 2.
\textbullet~Add blue, glowing cloud formations at the base, consistent with the ethereal atmosphere of Image 2.
}}
\\[4pt]
\midrule \noalign{\vskip-8pt}

\raisebox{-\height}{\begin{tabular}[t]{@{}c@{}}
\includegraphics[width=0.065\linewidth,height=0.065\linewidth]{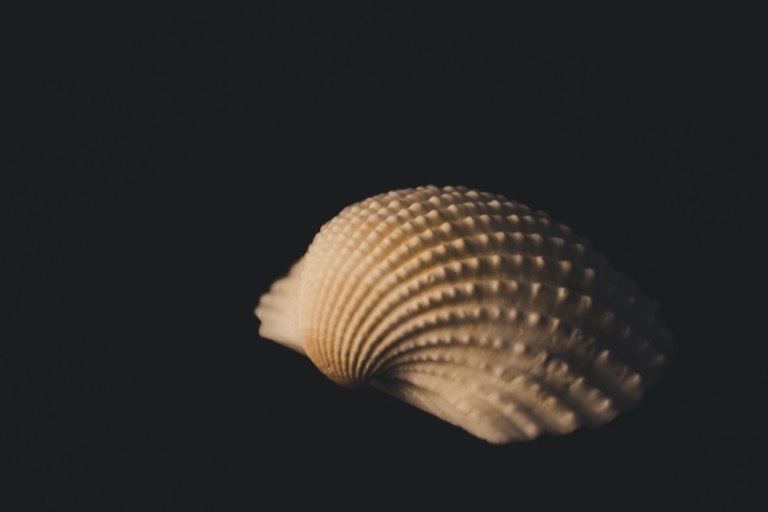} \\
\includegraphics[width=0.065\linewidth,height=0.065\linewidth]{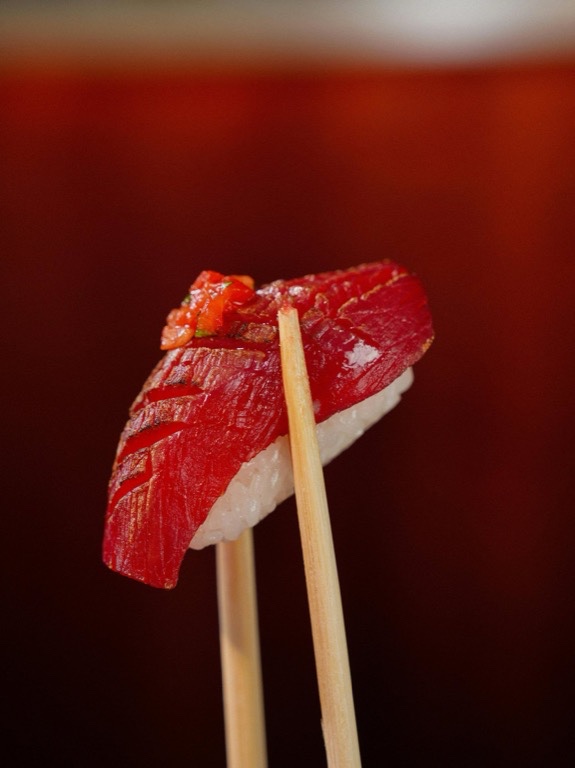}
\end{tabular}}
&
\raisebox{-\height}{\begin{tabular}[t]{@{}c@{}}
\includegraphics[width=0.1\linewidth,height=0.1\linewidth]{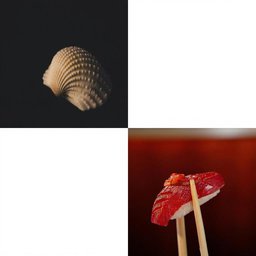} \\[-4pt]
\scriptsize(3 words)
\end{tabular}}
&
\raisebox{-\height}{\parbox[t]{0.1\linewidth}{\raggedright\scriptsize
\textbullet~copy entire grid
}}
&
\raisebox{-\height}{\begin{tabular}[t]{@{}c@{}}
\includegraphics[width=0.1\linewidth,height=0.1\linewidth]{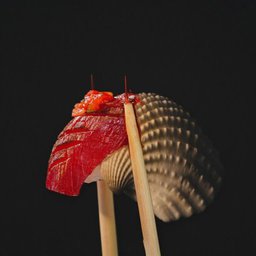} \\[-4pt]
\scriptsize(36 words)
\end{tabular}}
&
\raisebox{-\height}{\parbox[t]{0.1\linewidth}{\raggedright\scriptsize
\textbullet~Extract the sushi and chopsticks from $\langle$image2$\rangle$.
\textbullet~Extract the seashell from $\langle$image1$\rangle$.
\textbullet~Combine these two elements, placing the sushi and chopsticks in front of the seashell.
\textbullet~Use the dark background from $\langle$image1$\rangle$.
}}
&
\raisebox{-\height}{\begin{tabular}[t]{@{}c@{}}
\includegraphics[width=0.1\linewidth,height=0.1\linewidth]{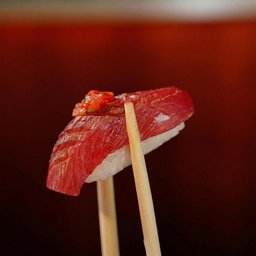} \\[-4pt]
\scriptsize(3 words)
\end{tabular}}
&
\raisebox{-\height}{\parbox[t]{0.1\linewidth}{\raggedright\scriptsize
\textbullet~copy $\langle$image2$\rangle$
}}
&
\raisebox{-\height}{\begin{tabular}[t]{@{}c@{}}
\includegraphics[width=0.1\linewidth,height=0.1\linewidth]{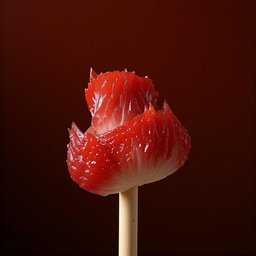} \\[-4pt]
\scriptsize(57 words)
\end{tabular}}
&
\raisebox{-\height}{\parbox[t]{0.23\linewidth}{\raggedright\scriptsize
\textbullet~Use the seashell object and its intricate texture from image 1.
\textbullet~Apply the red color and glossy, translucent material from the tuna in image 2 to the seashell shape.
\textbullet~Extract one chopstick from image 2 and position it to hold the modified seashell.
\textbullet~Place the resulting object in the background from image 2.
}}
\\

\bottomrule
\end{tabular}

\caption{Description complexity comparison (2/2). Continued from previous figure. Note how Kontext consistently produces grid layouts, while Qwen and Nano Banana sometimes simply copy one input. Our method consistently generates non-trivial combinations requiring detailed descriptions.}
\label{fig:caption_length_comparison_2}
\end{figure*}

We show examples of the resulting descriptions for the outputs of different methods in \Cref{fig:caption_length_comparison,fig:caption_length_comparison_2}.

\begin{figure}
    \centering
    \includegraphics[width=1\linewidth]{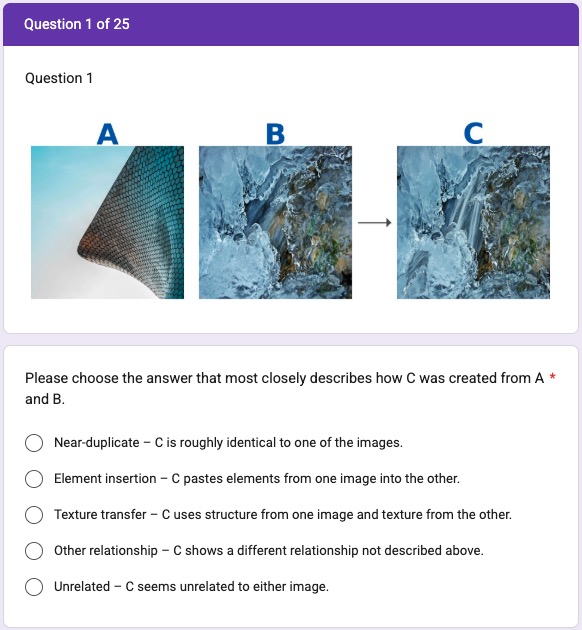}
    \caption{User study interface. Participants viewed an output image alongside its two inputs and classified the relationship between them.}
    \label{fig:user_study_interface}
\end{figure}

\newpage
\section{User Study Details}
We recruited 35 participants through university mailing lists and personal networks. Participants completed the study via Google Forms. For each of 25 trials, participants were shown an output image alongside its two input images and asked to classify the relationship between them. The study took approximately $12$ minutes to complete. \Cref{fig:user_study_interface} shows the study interface.

Participants selected from five options describing how the output relates to the inputs:
\begin{enumerate}
    \item \textbf{Near-duplicate} — the output is roughly identical to one of the input images.
    \item \textbf{Element insertion} — elements from one image are pasted into the other.
    \item \textbf{Texture transfer} — structure from one image combined with texture from the other.
    \item \textbf{Other relationship} — a relationship not captured by the above categories.
    \item \textbf{Unrelated} — no apparent connection to either input.
\end{enumerate}

 We sampled 25 output images stratified by description length to ensure coverage across the complexity spectrum, comprising 11 images from our method and 7 each from Nano Banana and Qwen-Image-2511.
 We omit Flux.1 Kontext from the study as it has a different input format which it tends to copy and would require different classification options, and the other methods produced superior results.

\begin{figure*}[p]
    \centering
    \setlength{\tabcolsep}{8pt}
    \begin{tabular}{ccccc}
    \toprule
        Inputs & Flux.1 Kontext & Qwen-Image-2511 & Nano Banana Pro & Ours \\[2pt]
        \hline \\[-6pt]
        \comprow{arch3__food2}{arch3}{food2} \rowsep
        \comprow{arch6__nature2}{arch6}{nature2} \rowsep
        \comprow{arch7__sea8}{arch7}{sea8} \rowsep
        \comprow{fashion3__nature6}{fashion3}{nature6} \\
    \end{tabular}
    \caption{Generation results comparison (2/5). Continued from previous figure.}
    \label{fig:generation_results_comparison_2}
\end{figure*}

\begin{figure*}[p]
    \centering
    \setlength{\tabcolsep}{8pt}
    \begin{tabular}{ccccc}
    \toprule
        Inputs & Flux.1 Kontext & Qwen-Image-2511 & Nano Banana Pro & Ours \\[2pt]
        \hline \\[-6pt]
        \comprow{fashion4__arch4}{fashion4}{arch4} \rowsep
        \comprow{fashion4__food6}{fashion4}{food6} \rowsep
        \comprow{fashion5__food1}{fashion5}{food1} \rowsep
        \comprow{fashion6__nature7}{fashion6}{nature7} \\
    \end{tabular}
    \caption{Generation results comparison (3/5). Continued from previous figure.}
    \label{fig:generation_results_comparison_3}
\end{figure*}

\begin{figure*}[p]
    \centering
    \setlength{\tabcolsep}{8pt}
    \begin{tabular}{ccccc}
    \toprule
        Inputs & Flux.1 Kontext & Qwen-Image-2511 & Nano Banana Pro & Ours \\[2pt]
        \hline \\[-6pt]
        \comprow{food1__nature2}{food1}{nature2} \rowsep
        \comprow{food1__other1}{food1}{other1} \rowsep
        \comprow{food2__nature4}{food2}{nature4} \rowsep
        \comprow{food4__other2}{food4}{other2} \\
    \end{tabular}
    \caption{Generation results comparison (4/5). Continued from previous figure.}
    \label{fig:generation_results_comparison_4}
\end{figure*}

\begin{figure*}[p]
    \centering
    \setlength{\tabcolsep}{8pt}
    \begin{tabular}{ccccc}
    \toprule
        Inputs & Flux.1 Kontext & Qwen-Image-2511 & Nano Banana Pro & Ours \\[2pt]
        \hline \\[-6pt]
        \comprow{food4__sea9}{food4}{sea9} \rowsep
        \comprow{food5__sea5}{food5}{sea5} \rowsep
        \comprow{food6__nature8}{food6}{nature8} \rowsep
        \comprow{nature3__sea7}{nature3}{sea7} \\
    \end{tabular}
    \caption{Generation results comparison (5/5). Continued from previous figure.}
    \label{fig:generation_results_comparison_5}
\end{figure*}

\end{document}